\documentclass[journal]{IEEEtran}

\ifCLASSINFOpdf
\else
\fi
\usepackage{amsmath,amssymb,amsthm}
\usepackage{authblk}
\usepackage{graphicx}
\usepackage{cite}
\usepackage{float}
\usepackage{epstopdf}
\usepackage{forest}
\usepackage{tikz-qtree}
\usetikzlibrary{arrows.meta, shapes.geometric, calc, shadows}
\usepackage{tabu}
\usepackage{multirow}
\usepackage{adjustbox}

\usepackage{hhline}

\usepackage{algorithm}
\usepackage{algorithmic}

\usepackage{url}
\usepackage{tocloft}
\usepackage{nccmath}

\usepackage{yfonts,color}
\usepackage{color}
\usepackage{xcolor}
\usepackage{fancyhdr}

\usepackage{rotating}

\usepackage{authblk}
\usepackage{hhline}
\usepackage{graphics}
\usepackage{graphicx}

\usepackage{epstopdf}
\usepackage{algorithm}
\usepackage{algorithmic}
\usepackage{amsmath}

\usepackage{array,booktabs,longtable,tabularx,tabulary}
\newcolumntype{L}{>{\raggedright\arraybackslash}X}
\usepackage{ltablex}
\usepackage{caption}%
\usepackage[skip=2pt,font=scriptsize]{caption}
\usepackage{subcaption}
\usepackage{amssymb}
\usepackage{multirow}
\usepackage{hhline}
\usepackage{url}
\usepackage{hyperref}
\usepackage{graphics}
\usepackage{graphicx}
\usepackage{epstopdf}
\usepackage{wrapfig}
\usepackage{color,soul}
\usepackage{soul}
\usepackage{amsfonts}
\usepackage{caption}
\usepackage{fancyhdr} 
\usepackage{rotating}
\makeatletter
\newcommand{\tocfill}{\cleaders\hbox{$\m@th \mkern\@dotsep mu . \mkern\@dotsep mu$}\hfill}
\makeatother

\newenvironment{abbreviations}{\begin{list}{}{
\setlength{\itemsep}{0pt}}}{\end{list}}

\usepackage{smartdiagram}
\usesmartdiagramlibrary{additions}

\begin{document}
\title{Unmanned Aerial Vehicles: A Survey on Civil Applications and Key Research Challenges}
	\author{Hazim Shakhatreh, Ahmad Sawalmeh, Ala Al-Fuqaha, Zuochao Dou, Eyad Almaita, Issa Khalil, Noor Shamsiah Othman, Abdallah Khreishah, Mohsen Guizani
	\thanks{Hazim Shakhatreh, Zuochao Dou, and Abdallah Khreishah are with the Department of
		Electrical and Computer Engineering, New Jersey Institute of Technology.
		(e-mail: \{hms35,zd36,abdallah\}@njit.edu)}
	\thanks{Ahmad Sawalmeh and Noor Shamsiah Othman are with the Department of Electronics and Communication Engineering, Universiti Tenaga Nasional. (e-mail: \{PE20656@utn.edu.my,Shamsiah@uniten.edu.my\})}
	\thanks{Ala Al-Fuqaha is with the Department of Computer Science, Western Michigan University, Kalamazoo, MI, 49008, USA. (e-mail: (ala.al-fuqaha@wmich.edu)}
	\thanks{Eyad Almaita is with the Department of Power and Mechatronics Engineering, Tafila Technical University. (e-mail: (e.maita@ttu.edu.jo) }
	\thanks{Issa Khalil is with Qatar Computing Research Institute (QCRI), HBKU, Doha, Qatar. (e-mail: ikhalil@hbku.edu.qa)}
	\thanks{Mohsen Guizani is with the University of Idaho, Moscow, ID, 83844, USA. (e-mail: mguizani@uidaho.edu)}}
	
	\maketitle
	
\section*{Abstract}
The use of unmanned aerial vehicles (UAVs) is growing rapidly across many civil application domains including real-time monitoring, providing wireless coverage, remote sensing, search and rescue, delivery of goods, security and surveillance, precision agriculture, and civil infrastructure inspection. Smart UAVs are the next big revolution in UAV technology promising to provide new opportunities in different applications, especially in civil infrastructure in terms of reduced risks and lower cost. Civil infrastructure is expected to dominate the more that $\$45$ Billion market value of UAV usage. In this survey, we present UAV civil applications and their challenges. We also discuss current research trends and provide future insights for potential UAV uses. Furthermore, we present the key challenges for UAV civil applications, including: charging challenges, collision avoidance and swarming challenges, and networking and security related challenges. Based on our review of the recent literature, we discuss open research challenges and draw high-level insights on how these challenges might be approached.

\begin{IEEEkeywords}
	UAVs, Wireless Coverage, Real-Time Monitoring, Remote Sensing, Search and Rescue, Delivery of goods, Security and Surveillance, Precision Agriculture, Civil Infrastructure Inspection.
\end{IEEEkeywords}

\section{Introduction}
UAVs can be used in many civil applications due to their ease of deployment, low maintenance cost, high-mobility and ability to hover~\cite{hayat2016survey}. Such vehicles are being utilized for real-time monitoring of road traffic, providing wireless coverage, remote sensing, search and rescue operations, delivery of goods, security and surveillance, precision agriculture, and civil infrastructure inspection. The recent research literature on UAVs focuses on vertical applications without considering the challenges facing UAVs within specific vertical domains and across application domains. Also, these studies do not discuss practical ways to overcome challenges that have the potential to contribute to multiple application domains.

The authors in~\cite{hayat2016survey} present the characteristics and requirements of UAV networks for envisioned civil applications over the period 2000\textendash2015 from a communications and networking viewpoint. They survey the quality of service requirements, network-relevant mission parameters, data requirements, and the minimum data to be transmitted over the network for civil applications. They also discuss general networking related requirements, such as connectivity, adaptability, safety, privacy, security, and scalability. Finally, they present experimental results from many projects and investigate the suitability of existing communications technologies to support reliable aerial networks.
\begin{figure*}
	\centering
	\includegraphics[scale=0.4]{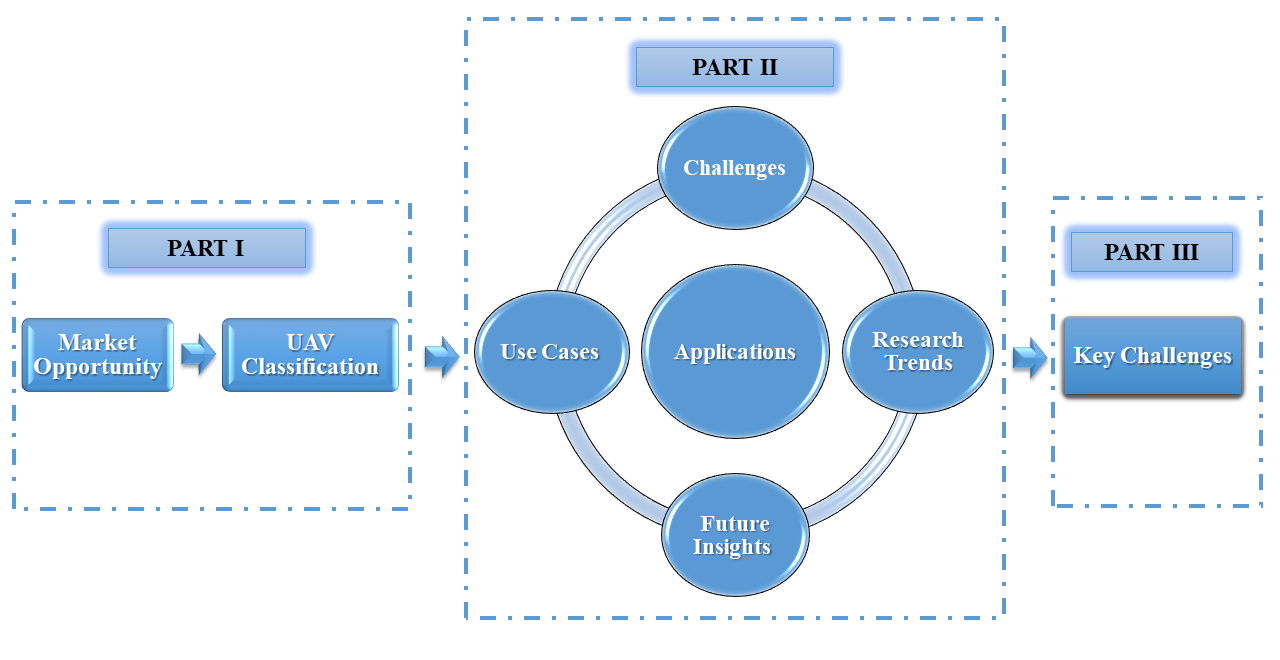}
	\caption{Overall Structure of the Survey.}
	\label{hazoverall}
\end{figure*}

\begin{table*}
	\scriptsize
	\renewcommand{\arraystretch}{1}
	\caption{\uppercase{Comparison of related work on UAV surveys in terms of applications}}
	\label{tableint1}
	\centering
	\begin{tabular}{|c|c|c|c|c|c|c|c|c|}
		\hline 
		Reference&Providing wireless &Remote sensing&Real-Time &Search and &Delivery of &Surveillance &Precision& Infrastructure \\
		&coverage&&monitoring&Rescue&goods&&agriculture&inspection\\
		\hline
		\cite{hayat2016survey}	&\checkmark&&&\checkmark&\checkmark&&&\checkmark\\
		\hline
		\cite{gupta2016survey}	&\checkmark&\checkmark&&&&&&\\
		\hline
		\cite{motlagh2016low}	&\checkmark&&\checkmark&&\checkmark&\checkmark&&\checkmark\\
		\hline
		\cite{mozaffari2018tutorial}	&\checkmark&&&&&&&\\
		\hline
		\cite{khawaja2018survey}	&\checkmark&&&&&&&\\
		\hline
		\cite{aja2018}	&\checkmark&&&&&&&\\
		\hline
		This work&\checkmark&\checkmark&\checkmark&\checkmark&\checkmark&\checkmark&\checkmark&\checkmark\\
		\hline
	\end{tabular}
\end{table*}
\begin{table*}
	\scriptsize
	\renewcommand{\arraystretch}{1}
	\caption{\uppercase {Comparison of related work on UAV surveys in terms of new technology trends}}
	\label{tableint2}
	\centering
	\begin{tabular}{|c|c|c|c|c|c|c|c|c|}
		\hline 
		Reference&Collision&mmWave&Free space &Cloud &Machine &NFV&SDN &Image \\
		&avoidance&&optical&computing&learning&&&processing\\
		\hline
		\cite{hayat2016survey}	&\checkmark&&&&&&&\checkmark\\
		\hline
		\cite{gupta2016survey}	&\checkmark&&&&&&\checkmark&\\
		\hline
		\cite{motlagh2016low}	&\checkmark&&&&&&&\checkmark\\
		\hline
		\cite{mozaffari2018tutorial}	&&\checkmark&&\checkmark&\checkmark&&&\\
		\hline
		\cite{khawaja2018survey}	&\checkmark&&&&&&&\\
		\hline
		\cite{aja2018}	&&&&&&&&\\
		\hline
		This work&\checkmark&\checkmark&\checkmark&\checkmark&\checkmark&\checkmark&\checkmark&\checkmark\\
		\hline
	\end{tabular}
\end{table*}
In~\cite{gupta2016survey}, the authors attempt to focus on research in the areas of routing, seamless handover and energy efficiency. First, they distinguish between infrastructure and ad-hoc UAV networks, application areas in which UAVs act as servers or as clients, star or mesh UAV networks and whether the deployment is hardened against delays and disruptions. Then, they focus on the main issues of routing, seamless handover and energy efficiency in UAV networks. The authors in~\cite{bekmezci2013flying} survey Flying Ad-Hoc Networks (FANETs) which are ad-hoc networks connecting the UAVs. They first clarify the differences between FANETs, Mobile Ad-hoc Networks (MANETs) and Vehicle Ad-Hoc Networks (VANETs). Then, they introduce the main FANET design challenges and discuss open research issues. In~\cite{zeng2016wireless}, the authors provide an overview of UAV-aided wireless communications by introducing the basic networking architecture and main channel characteristics. They also highlight the key design considerations as well as the new opportunities to be explored.  

The authors of~\cite{kumbhar2017survey} present an overview of legacy and emerging public safety communications technologies along with the spectrum allocation for public safety usage across all the frequency bands in the United States. They conclude that the application of UAVs in support of public safety communications is shrouded by privacy concerns and lack of comprehensive policies, regulations, and governance for UAVs. In~\cite{chmaj2015distributed}, the authors survey the applications implemented using cooperative swarms of UAVs that operate as distributed processing system. They classify the distributed processing applications into the following categories: 1) general purpose distributed processing applications, 2) object detection, 3) tracking, 4) surveillance, 5) data collection, 6) path planning, 7) navigation, 8) collision avoidance, 9) coordination, 10) environmental monitoring. However, this survey does not consider the challenges facing UAVs in these applications and the potential role of new technologies in UAV uses. The authors of~\cite{motlagh2016low} provide a comprehensive survey on UAVs, highlighting their potential use in the delivery of Internet of Things (IoT) services from the sky. They describe their envisioned UAV-based architecture and present the relevant key challenges and requirements.

In~\cite{mozaffari2018tutorial}, the authors provide a comprehensive study on the use of UAVs in wireless networks. They investigate two main use cases of UAVs; namely, aerial base stations and cellular-connected users. For each use case of UAVs, they present key challenges, applications, and fundamental open problems. Moreover, they describe mathematical tools and techniques needed for meeting UAV challenges as well as analyzing UAV-enabled wireless networks. The authors of~\cite{khawaja2018survey} provide a comprehensive survey on available Air-to-Ground channel measurement campaigns, large and small scale fading channel models, their limitations, and future research directions for UAV communications scenarios. In~\cite{aja2018}, the authors provide a survey on the measurement campaigns launched for UAV channel modeling using low altitude platforms and discuss various channel characterization efforts. They also review the contemporary perspective of UAV channel modeling approaches and outline some future research challenges in this domain. 

UAVs are projected to be a prominent deliverer of civil services in many areas including farming, transportation, surveillance, and disaster management. In this paper, we review several UAV civil applications and identify their challenges. We also discuss the research trends for UAV uses and future insights. The reason to undertake this survey is the lack of a survey focusing on these issues. Tables~\ref{tableint1} and~\ref{tableint2} delineate the closely related surveys on UAV civil applications and demonstrate the novelty of our survey relative to existing surveys. Specifically, the contributions of this survey can be delineated as:
\begin{itemize}
	\item Present the global UAV payload market value. The payload covers all equipment which are carried by UAVs
	such as cameras, sensors, radars, LIDARs, communications equipment, weaponry, and others. We also present the market value of UAV uses.
	\item Provide a classification of UAVs based on UAV endurance, maximum altitude, weight, payload, range, fuel type, operational complexity, coverage range and applications.
	\item Present UAV civil applications and challenges facing UAVs in each application domain. We also discuss the research trends for UAV uses and future insights. The UAV civil applications covered in this survey include: real-time monitoring of road traffic, providing wireless coverage, remote sensing, search and rescue, delivery of goods, security and surveillance, precision agriculture, and civil infrastructure inspection.
	\item Discuss the key challenges of UAVs across different application domains, such as charging challenges, collision avoidance and swarming challenges, and networking and security challenges.
\end{itemize}
Our survey is beneficial for future research on UAV uses, as it comprehensively serves as a resource for UAV applications, challenges, research trends and future insights, as shown in Figure~\ref{hazoverall}. For instance, UAVs can be utilized for providing wireless coverage to remote areas such as Facebook’s Aquila UAV~\cite{zuckerberg2014connecting}. In this aplication, UAVs need to return periodically to a charging station for recharging, due to their limited battery capacity. To overcome this challenge, solar panels installed on UAVs harvest the received solar energy and convert it to electrical energy for long endurance flights~\cite{sun2018resource}. We can envisage Laser power-beaming as a future technology to provide supplemental energy at night when solar energy is not available or is minimal at high latitudes during winter. This would enable such UAVs to fly day and night for weeks or possibly months without landing~\cite{sheet2014beamed}.

\begin{figure*}
	\centering
	\includegraphics[scale=0.42]{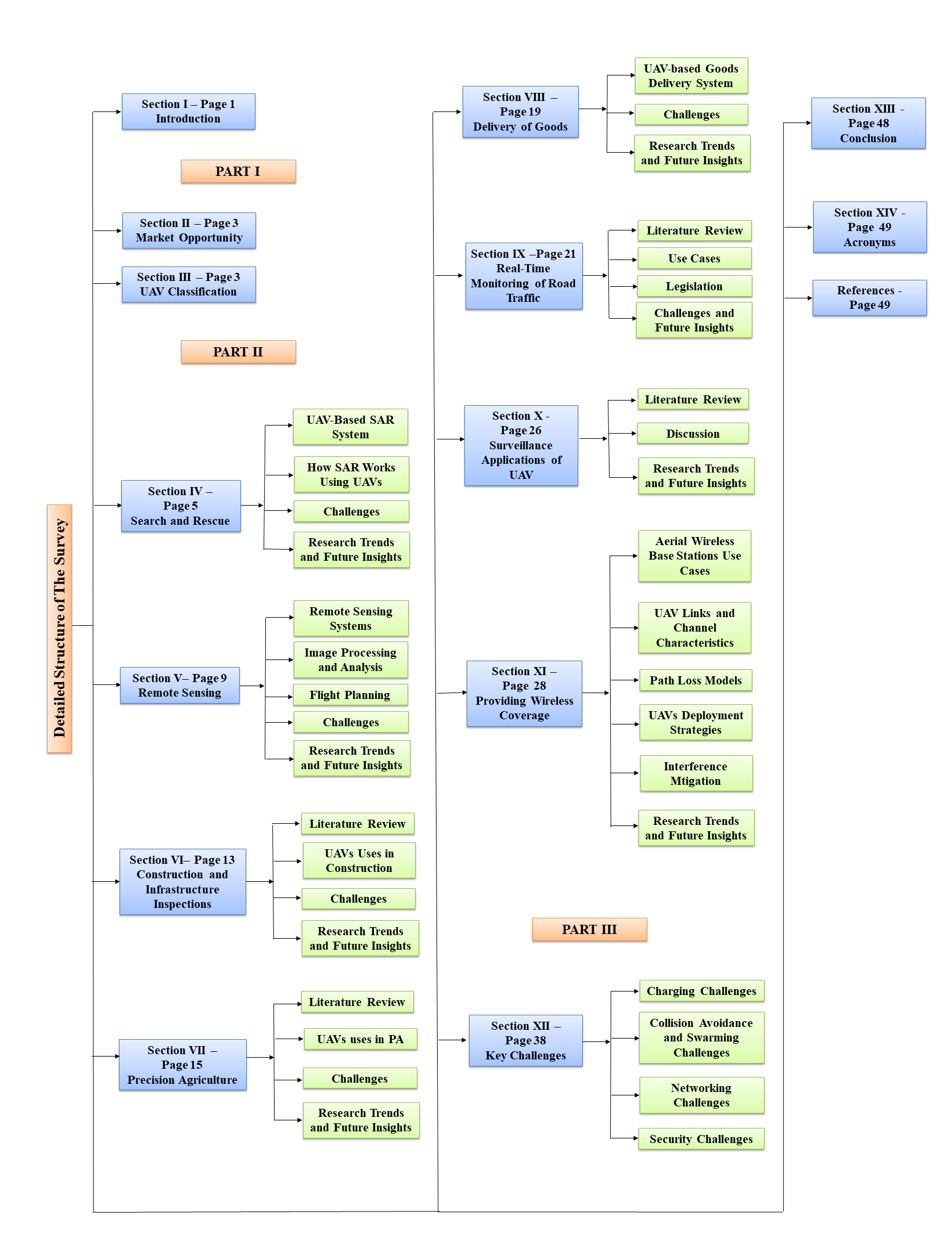}
	\caption{Detailed Structure of the Survey.}
	\label{enable2}
\end{figure*}

The rest of the survey is organized as follows. We start with an overview of the global UAV market value and UAV classification in Part I of this survey (Sections II and III). In Part II (Sections IV-XI), we present UAV civil applications and the challenges facing UAVs in each application domain, and we also discuss the research trends and future insights for UAV uses. In Part III (Sections XII and XIII), we discuss the key challenges of UAV civil applications and conclude this study. Finally, a list of the acronyms used in the survey is presented in Section XIV. To facilitate reading, Figure~\ref{enable2} provides a detailed structure of the survey.

\section*{PART I: MARKET OPPORTUNITY and UAV Classification }
\section{MARKET OPPORTUNITY}
\label{Market}
\begin{figure}
	\centering
	\includegraphics[scale=0.53]{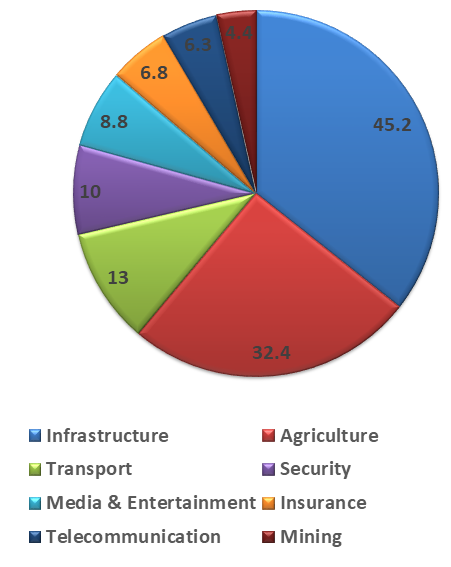}
	\caption{
		Predicted Value of UAV Solutions in Key Industries (Billion).}
	\label{ma2}
\end{figure}

\begin{figure}
	\centering
	\includegraphics[scale=0.53]{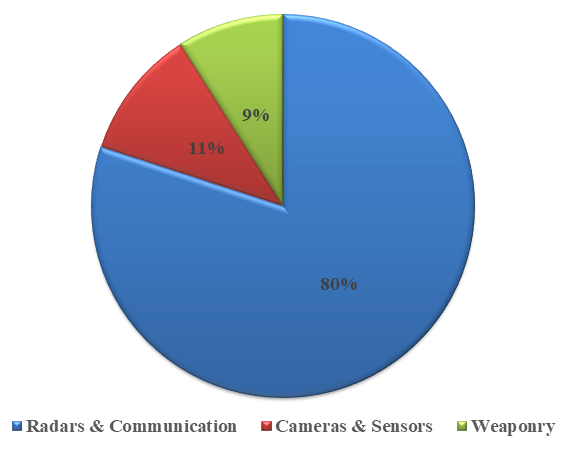}
	\caption{
		Global UAV Payload Market Predictions 2027.}
	\label{ma1}
\end{figure}

UAVs offer a great market opportunity for equipment manufacturers, investors and business service providers. According to the PwC report~\cite{pwc18}, the addressable market value of UAV uses is over $\$127$ billion as shown in Figure~\ref{ma2}. Civil infrastructure is expected to dominate the addressable market value of UAV uses, with  market value of  $\$45$  billion.  A report released by the Association for Unmanned Vehicle Systems International, expects more than $100,000$ new jobs in unmanned aircrafts by $2025$~\cite{Tiffany}. To know how to operate a UAV and meet job requirements, a person should attend training programs at universities or specialized institutes. Global UAV payload market value is expected to reach $\$3$ billion by $2027$ dominated by North America, followed by Asia-Pacific and Europe~\cite{SDI}. The payload covers all equipment which are carried by UAVs such as cameras, sensors, radars, LIDARs, communications equipment, weaponry, and others~\cite{grandviewresearch}. Radars and communications equipment segment is expected to dominate the global UAV payload market with a market share of close to $80\%$, followed by cameras and sensors segment with around over $11\%$ share and weaponry segment with almost $9\%$ share~\cite{SDI} as shown in Figure~\ref{ma1}.

Business Intelligence expects sales of UAVs to reach $\$12$ billion in $2021$, which is up by a compound annual growth rate of $7.6\%$ from $\$8.5$ billion in $2016$~\cite{DivyaJoshi}. This future growth is expected to occur across three main sectors: 1) Consumer UAV shipments which are projected to reach $29$ million in $2021$; 2) Enterprise UAV shipments which are projected to reach $805,000$ in $2021$; 3) Government UAVs for combat and surveillance. According to Bard Center for the Study of UAVs, U.S. Department of Defense allocated a budget of $\$4.457$ billion for UAVs in $2017$~\cite{gettinger2016drone}.

All these statistics show the economic importance of UAVs and their applications in the near future for equipment manufacturers, investors and business service providers. Smart UAVs will provide a unique opportunity for UAV manufacturers to utilize new technological trends to overcome current challenges of UAV applications. To spread UAV services globally, a complete legal framework and institutions regulating the commercial use of UAVs are needed~\cite{pwc17}.

\section{UAV Classification}
\label{S:1}
Unmanned vehicles can be classified into five different types according to their operation. These five types are unmanned ground vehicles, unmanned aerial vehicles, unmanned surface vehicles (operating on the surface of water), unmanned underwater vehicles, and unmanned spacecrafts. Unmanned vehicles can be either remote guided or autonomous vehicles \cite{Uncrewed-vehicle:Online}.

There has been many research studies on unmanned vehicles that reported progress towards \textit{autonomic systems} that do not require human interactions. In line with the concept of autonomy with respect to humans and societies, technical systems that claim to be autonomous must be able to make decisions and react to events without direct interventions by humans \cite{sebbane2015smart}. 
Therefore, some fundamental elements are common to all autonomous vehicles. These elements include: ability of sensing and perceiving the environment, ability of analyzing, communicating, planning and decision making using on-board computers, as well as acting which requires vehicle control algorithms.


UAV features may vary depending on the application in order for them to fit their specific task. Therefore, any classification of UAVs needs to take into consideration their various features as they are widely used for a variety of civilian operations \cite{korchenko2013generalized}. The use of UAVs as an aerial base station in communications networks, can be categorized based on their operating platform, which can be a Low Altitude Platform (LAP) \cite{al2014modeling,valcarce2013airborne,reynaud2012deployable} or High Altitude Platform (HAP) \cite{tozer2001high,thornton2001broadband,karapantazis2005broadband}.

LAP is a quasi-stationary aerial communications platform that operates at an altitude of less than $10$ km. Three main types of UAVs that fall under this category are vertical take-off and landing (VTOL) vehicles, aircrafts, and balloons.

\begin{figure*}[!h]
	\centering
	\begin{forest}
		for tree={
			align=center,
			parent anchor=south,
			child anchor=north,
			font=\footnotesize,
			edge={thick, -{Stealth[]}},
			l sep+=10pt,
			edge path={
				\noexpand\path [draw, \forestoption{edge}] (!u.parent anchor) -- +(0,-10pt) -| (.child anchor)\forestoption{edge label};
			},
			if level=0{
				inner xsep=0pt,
				tikz={\draw [thick] (.south east) -- (.south west);}
			}{}
		}
		[  UAV classification based on  \\ communication platform  
		[LAP
		[Balloon
		[Tethered Helikite\\ \cite{chandrasekharan2016designing,bucaille2013rapidly}]
		]
		[VTOL
		[Quadrotor
		[The FALCON \cite{Sentinel:Online}]
		]
		]
		[Aircraft
		[Viking aircraft\\ \cite{matolak2014initial,sun2017air,matolak2017air}]
		]
		]
		[HAP
		[Aircraft
		[Helios\\(NASA) \cite{tozer2001high}\\Heliplat(Europe)\\ \cite{thornton2001broadband,grace2003european}]
		]
		[Balloon
		[Project Loon\\Balloon\\(Google)\\\cite{katikala2014google}
		]
		]
		[Airship
		[Zeppelin\\NT \\ \cite{tozer2001high}]
		]
		]
		]
	\end{forest}
	\caption{UAV Classification.}
	\label{fig:ahmad1-1}
\end{figure*}
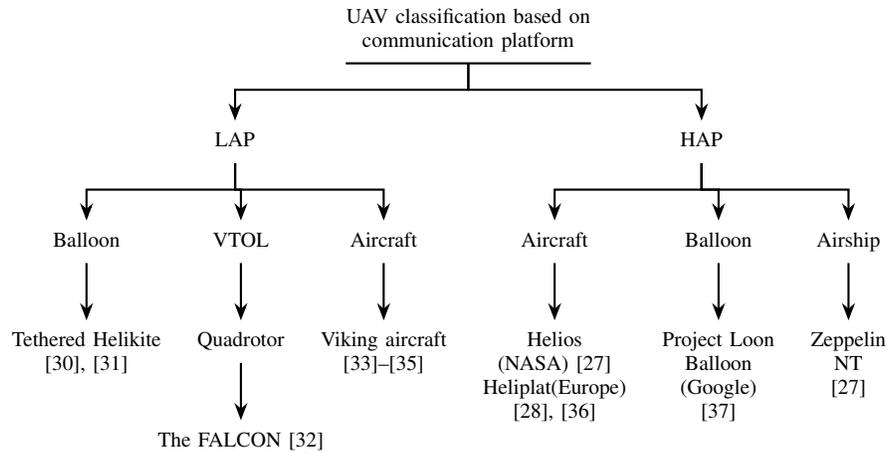

On the other hand, HAP operates at very high altitude above $10$ km, and vehicles utilizing this platform are able to stay for a long time in the upper layers of the stratosphere. Airships, aircrafts, and balloons are the main types of UAVs that fall under this category. These two categories of aerial communications platforms are shown in Figure ~\ref{fig:ahmad1-1}. More specifically, in each platform the main UAV types and examples of each type are illustrated in this figure.
A comparison between HAP and LAP is presented in Table~\ref{table:ahmad1-2} . This table also summarizes the main UAV types for each platform, and its performance parameters. Specifically, these parameters are UAV endurance, maximum altitude, weight, payload, range, deployment time, fuel type, operational complexity, coverage range, applications and examples for each UAV type is also shown in this table.\\
\begin{table*}[!t]
	\footnotesize
	\renewcommand{\arraystretch}{1.5}
	\label{table1}
	\centering
	\caption{\uppercase{Platform classification of UAV types, and performance parameters}}
	\begin{adjustbox}{width=1.0\textwidth}
		\begin{tabular}{|l||l|l|l||l|l|l|}
			\hline \hline
			Issues &\multicolumn{3}{|c||}{HAP}&\multicolumn{3}{|c|}{LAP} \\
			\hline \hline
			Type &Airship &Aircraft&Balloon &VTOL&Aircraft &Balloon \\
			\hline
			\hline 
			Endurance &long endurance &- 15-30 hours JP-fuel
			
			&Long endurance &Few hours&Few hours&From 1 day\\
			&&- $>$7 days Solar&Up to 100 days&&&To few days\\
			\hline
			Max. Altitude  &Up to 25 km &15-20 km&17-23 km&Up to 4 km&Up to 5 km&Up to 1.5 km\\
			\hline 
			Payload (kg)  &Hundreds of  kg's &Up to 1700 kg&Tens of kg's&Few kg's&Few kg's&Tens of kg's\\
			\hline
			Flight Range &Hundreds of km's &From 1500 to &Up to &Tens of km's&Less than 200 km &Tethered Balloon\\
			&&25000 km&17 million km&&&\\
			\hline
			Deployment time &Need Runway&Need Runway&custom-built &Easy to deploy &Easy to launched&Easy to deploy \\
			&&&Auto launchers&&by catapult&10-30 minutes\\
			\hline
			Fuel type &Helium Balloon&JP-8 jet fuel&Helium Balloon&Batteries&Fuel injection&Helium\\
			&&Solar panels&Solar panels&Solar panels&engine&\\
			\hline
			Operational  &Complex&Complex&Complex&Simple&Medium&Simple	 \\
			complexity&&&&&&\\
			\hline
			Coverage area  &Hundreds of km's&Hundreds of km's&Thousands of km's&Tens of km's&Hundreds of km's&Several tens \\
			&&&&&&of km's\\
			\hline
			UAV Weight  &Few hundreds&Few thousands &Tens of kg's&Few of kg's&Tens of kg's&Tens of kg's\\
			&of kg's&of kg's&&&&\\
			\hline
			Public safety  &Considered safe&Considered safe&Need global regulations &Need safety regulations&Safe&Safe\\
			\hline
			Applications  &Testing environmental&GIS Imaging&Internet Delivery&Internet Delivery&Agriculture &Aerial 
			\\
			&effects&&&&application&base station\\
			\hline
			Examples  &HiSentinel80 \cite{smith2011hisentinel80}&Global Hawk \cite{gupta2013review}&Project Loon&LIDAR \cite{Sentinel:Online}&EMT Luna &Desert Star \\
			&&& Balloon (Google)\cite{katikala2014google}&& X-2000 \cite{Luna2000}&34cm Helikite \cite{chandrasekharan2016designing}\\
			
			\hline
			\hline

		\end{tabular}
	\end{adjustbox}
	
	\label{table:ahmad1-2}
\end{table*}

\section*{PART II: UAV Applications}

\section{Search and Rescue (SAR) }
In the wake of new scientific developments, speculations shot up with regard to the future potential of UAVs in the context of public and civil domains. UAVs are believed to be of immense advantage in these domains, especially in support of public safety, search and rescue operations and disaster management. In case of natural or man-made disasters like floods, Tsunamis, or terrorist attacks, critical infrastructure including water and power utilities, transportation, and telecommunications systems can be partially or fully affected by the disaster. This necessitates rapid solutions to provide communications coverage in support of rescue operations \cite{hayat2016survey}.
When the public communications networks are disrupted, UAVs can provide timely disaster warnings and assist in speeding up rescue and recovery operations. UAVs can also carry medical supplies to areas that are classified as inaccessible. 
In certain disastrous situations like poisonous gas infiltration, wildfires, avalanches, and search for missing persons, UAVs can be used to play a support role and speed up SAR operations \cite{silvagni2017multipurpose}. Moreover, UAVs can quickly provide coverage of a large area without ever risking the security or safety of the personnel involved.
\subsection{UAV-Based SAR System}
SAR operations using traditional aerial systems (e.g., aircrafts and helicopters) are typically very costly. Moreover, aircrafts require special training, and special permits for taking off and landing areas. However, using UAVs in SAR operations reduces the costs, resources and human risks. Unfortunately, every year large amounts of money and time are wasted on SAR operations using traditional aerial systems \cite{silvagni2017multipurpose}. UAVs can contribute to reduce the resources needed in support of more efficient SAR operations.

\begin{figure}[!h]
	\centering
	\includegraphics[scale=0.30]{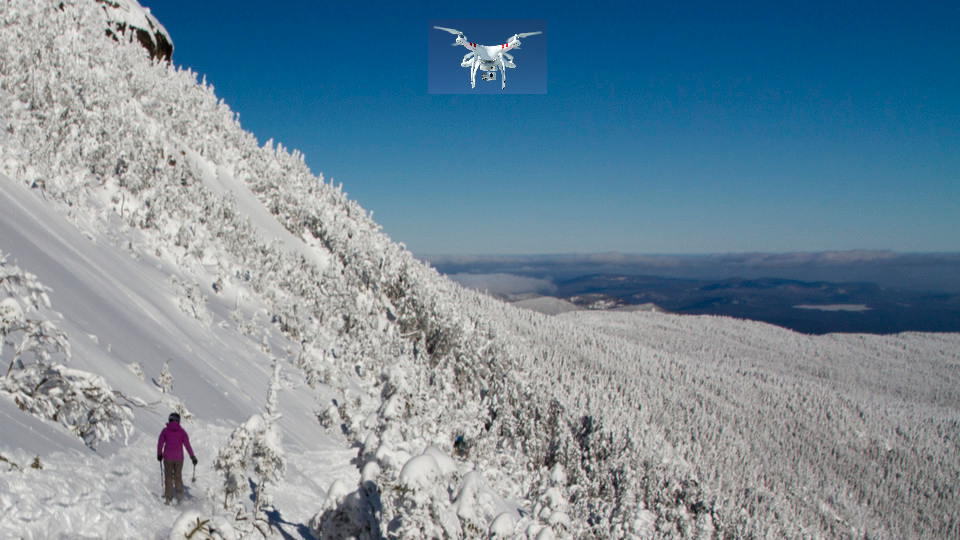}
	\caption{Use of Single UAV Systems in SAR Operations, Mountain Avalanche Events.  }
	\label{sar1}
\end{figure}

There are two types of SAR systems, single UAV systems, and Multi-UAV systems. A single UAV system is illustrated in Figure~\ref{sar1}. In the first step, the rescue team defines the search region, then the search operation is started by scanning the target area using a single UAV equipped with vision or thermal cameras. After that, real-time aerial videos/images from the targeted area are sent to the Ground Control System (GCS). These videos and images are analyzed by the rescue team to direct the SAR operations optimally \cite{doherty2007uav}.

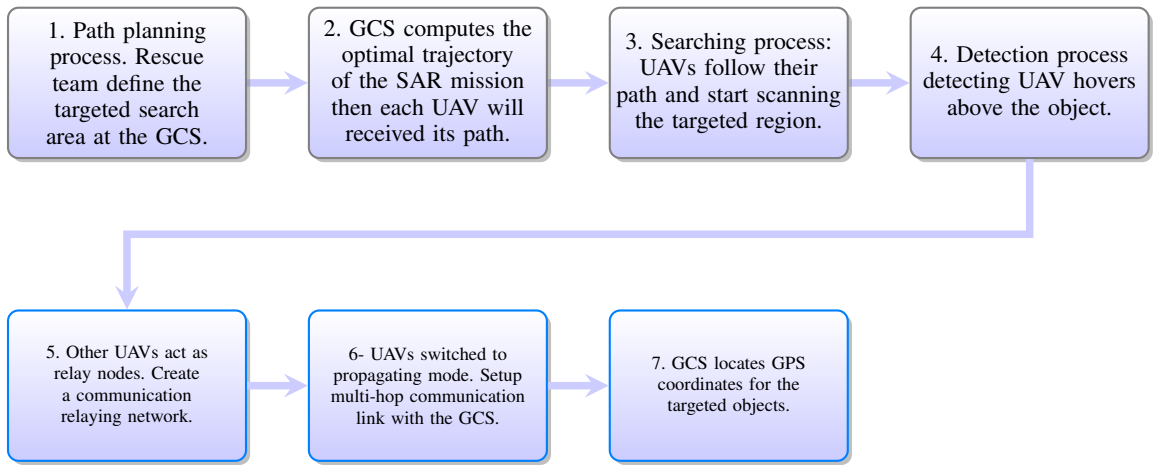
\begin{figure*}
	\begin{center}
		\scriptsize
		\smartdiagramset{
			uniform color list=blue!20!white for 6 items,
			back arrow disabled=true,
			module minimum width=2cm,
			module minimum height=2cm,
			module x sep=4cm,
			text width=3cm,
			additions={
				additional item offset=20mm,
				additional item width=2cm,
				additional item height=2cm,
				additional item text width=3cm,
				additional item shadow=drop shadow,
				additional item bottom color=blue!20!white,
				additional item border color=blue!50!cyan,
				additional arrow color=blue!20!white,
		}}
		\smartdiagramadd[flow diagram:horizontal]{
			1. Path planning process. Rescue team define the targeted search area at the GCS.
			, 2. GCS computes the optimal trajectory of the SAR mission then each UAV will received its path.
			, 3. Searching process: UAVs follow their path and start scanning the targeted region.,4. Detection process detecting UAV hovers above the object.
		}{
			below of module1/5. Other UAVs act as relay nodes. Create a communication relaying network.,below of module2/6- UAVs switched to propagating mode. Setup multi-hop communication link with the GCS.,below of module3/7. GCS locates GPS coordinates for the targeted objects.
		}
		\smartdiagramconnect{->}{additional-module1/additional-module2,additional-module2/additional-module3}
		\begin{tikzpicture}[remember picture,overlay]
		\draw[additional item arrow type] (module4) |- ([yshift=-10mm]module1.south) -- (additional-module1);
		\end{tikzpicture}
		\vspace{40mm}\par
		
	\end{center}
	
	\caption{ Use of Multi-UAV Systems in SAR Operations, Locate GPS Coordinates for the Missing Persons. }
	\label{sar-steps}
\end{figure*}

In Multi-UAV systems, UAVs with on-board imaging sensors are used to locate the position of missing persons. The following processes summarize the SAR operations that use these systems. Firstly, the rescue team conduct path planning to compute the optimal trajectory of the SAR mission. Then, each UAV receives its assigned path from the GCS. Secondly, the search process is started. During this process, all UAVs follow their assigned trajectories to scan the targeted region. This process utilizes object detection, video/image transmission and collision avoidance methods. Thirdly, the detection process is started. During this process, a UAV that detects an object hovers over it, while the other UAVs act as relay nodes to facilitate coordination between all the UAVs and communications with the GCS. Afterword, UAVs switch to data dissemination mode and setup a multi-hop communications link with the GCS. Finally, the location of the targeted object, and related videos and images are transmitted to the GCS. Figure~\ref{sar-steps}, illustrates the use of multi-UAV systems in support of SAR operations \cite{scherer2015autonomous}.   


\subsection{How SAR Operations Utilize UAVs}
SAR operations are one of the primary use-cases of UAVs. Their use in SAR operations attracted considerable attention and became a topic of interest in the recent past. SAR missions can utilize UAVs as follows:

\begin{enumerate}
	
	\item Taking high resolution images and videos using on-board cameras to survey a given target area \textit{(stricken region)}. Here, UAVs are used for post disaster aerial assessment or damage evaluation. This helps to evaluate the magnitude of the damage in the infrastructure caused by the disaster. After assessment, rescue teams can identify the targeted search area and commence SAR operations accordingly \cite{ruiz2017unmanned}.
	
	\item SAR operations using UAVs can be performed autonomously, accurately and without introducing additional risks \cite{silvagni2017multipurpose}. In the Alcedo project \cite{ALCEDO:Online}, a prototype was developed using a lightweight quadrotor UAV equipped with GPS to help in finding lost persons. In a Capstone project \cite{joern2015examining}, using UAVs in support of SAR operations in snow avalanche scenarios is explored. The used UAV utilizes thermal infrared imaging and Geographic Information System (GIS) data. In such scenarios, UAVs can be utilized to find avalanche victims or lost persons.
	
	\item UAV can be also used to deliver food, water and medicines to the injured. Although the use of UAVs in SAR operations can help to present potential dangers to crews of the flight, UAVs still suffer from capacity scale problems and limitations to their payloads. In \cite{jo2017development}, a UAV with vertical takeoff and landing capabilities was designed. Even, a high power propellant system has been added to allow the UAV to lift heavy cargo between $10$-$15$ Kg, which could include medicine, food, and water.

	\item UAVs can act as aerial base stations for rapid service recovery after complete communications infrastructure damage in disaster stricken areas. This helps in SAR operations as illustrated in Figures~\ref{SAR-out} and ~\ref{SAR-in} for outdoor and indoor environments, respectively.
	
	\begin{figure}[!h]
		\centering
		\includegraphics[scale=0.285]{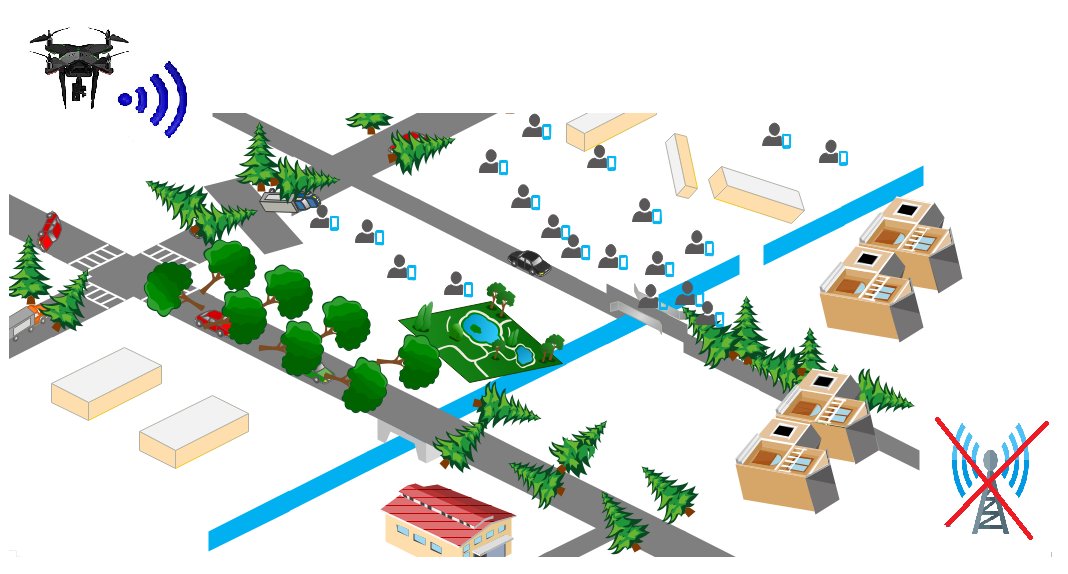}
		\caption{Use UAV to Provide Wireless Coverage for Outdoor Users. }
		\label{SAR-out}
	\end{figure}	
	\begin{figure}[!h]
		\centering
		\includegraphics[scale=0.35]{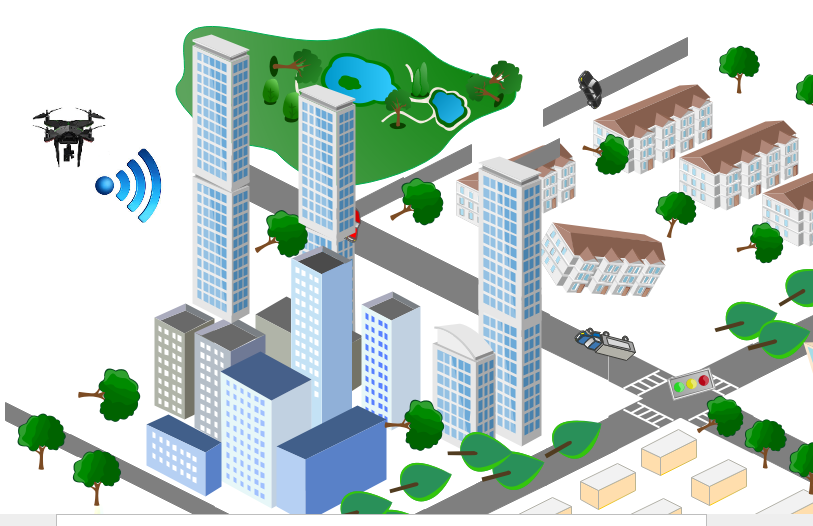}
		\caption{Use UAV to Provide Wireless Coverage for Indoor Users.}
		\label{SAR-in}
	\end{figure}
\end{enumerate}

\subsection{Challenges }
\subsubsection{Legislation}
In the United States, the FAA does not currently permit the use of swarms of autonomous UAVs for commercial applications. But it is possible to adjust the regulations to allow this type of use. Swarms of UAVs can be used to coordinate the operations of SAR teams \cite{smith2015regulating}.

\subsubsection{Weather}
Weather conditions pose a challenge to UAVs as they might result in deviations in their predetermined paths. In cases of natural or man-made disasters, such as Tsunamis, Hurricanes, or terrorist attacks, weather becomes a tough and cardinal challenge. In such scenarios, UAVs may fail in their missions as a result of the detrimental weather conditions \cite{jordan2015bird}. 

\subsubsection{Energy Limitations}
Energy consumption is one of the most important challenges facing UAVs. Usually, UAVs are battery powered. UAV batteries are used for UAV hovering, wireless communications, data processing and image analysis. 
In some SAR operations, UAVs need to be operated for extended periods of time over disaster stricken regions. Due to the power limitations of UAVs, a decision must be taken on whether UAVs should perform data and image analysis on-board in real-time, or data should be stored for later analysis to reduce the consumed power \cite{vergouw2016drone,gupta2016survey }. 

\begin{figure*}[!t]
	\centering
	\includegraphics[scale=0.4]{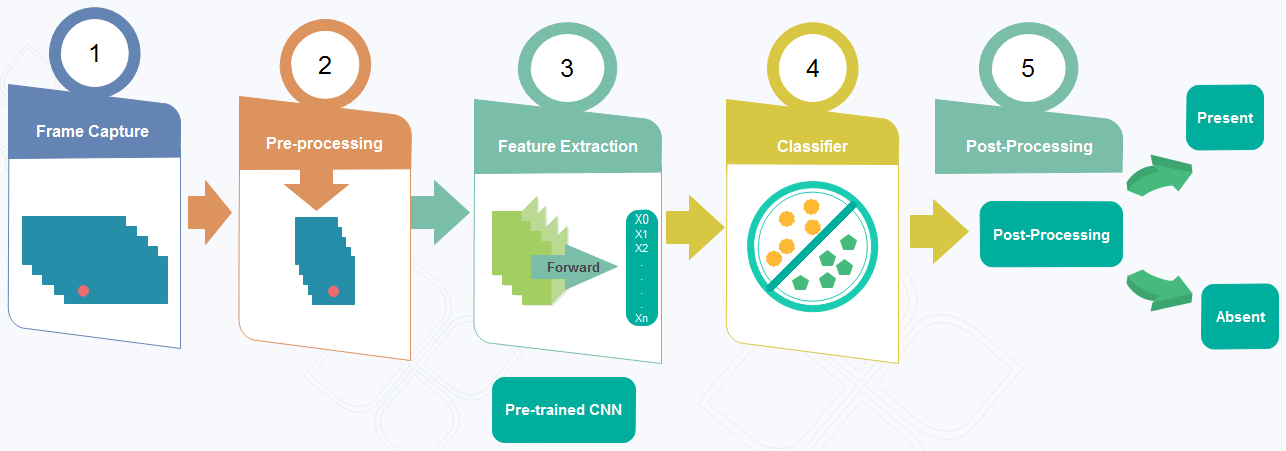}
	\caption{Block Diagram of SAR System Using UAVs with Machine Learning Technology.}
	\label{SAR-ML}
\end{figure*}

\subsection{Research Trends and Future Insights}
\subsubsection{Image Processing}
SAR operations using UAVs can employ image processing techniques to quickly and accurately find targeted objects. Image processing methods can be used in autonomous single and multi-UAV systems to locate potential targets in support of SAR operations. Moreover, location information can be augmented to aerial images of target objects \cite{ macke2013systems}.
UAVs can be integrated with target detection technologies including thermal and vision cameras. Thermal cameras (e.g., IR cameras) can detect the heat profile to locate missing persons. Two stage template based methods can be used with such cameras 
\cite{rudol2008human}. Vision cameras can also help in the detection process of objects and persons \cite{hernandez2012detecting, mikolajczyk2004human}. Methods that utilize a combination of thermal and vision cameras have been reported in the literature as in \cite{rudol2008human}.

In SAR operations, image processing can be done either at the GCS, post target identification, or at the UAV itself, using on-board processors with real-time image processing capabilities. In \cite{sun2016camera}, the authors implemented a target identification method using on-board processors, and the GCS. This method utilizes image processing techniques to identify the targeted objects and their coarse location coordinates. Using terrestrial networks, the UAV sends the images and their GPS locations to the GCS. Another possible approach requires the UAV to capture and save high resolution videos for later analysis at the GCS.

\subsubsection{Machine Learning}
Machine learning techniques can be applied on images captured by UAVs to help in SAR operations \cite{giusti2016machine, bejiga2017convolutional}.
In \cite{bejiga2017convolutional}, the authors propose  machine learning techniques applied to images captured by UAVs equipped with vision cameras. In their study, pre-trained Convolutional Neural Network (CNN) with trained linear Support Vector Machine (SVM) is used to determine the exact image/video frame at which a lost person is potentially detected. Figure~\ref{SAR-ML}. illustrates a block diagram of a SAR system utilizing UAVs in conjunction with machine learning technology \cite{bejiga2017convolutional}.

SAR operations employing machine learning technology face many challenges. UAVs are battery powered so there is significant limitation on their on-board processing capability. Protection against adversarial attacks on the employed machine learning techniques pose another important challenge 
Reliable and real-time communications with the GCS given QoS and energy constraints is another challenge \cite{bejiga2017convolutional}.

\subsubsection{Future Insights}
Based on the reviewed current literature focusing on SAR scenarios using UAVs, we believe there is a need for more research on the following:
\begin{itemize}

	\item Data fusion and decision fusion algorithms that integrate the output of multiple sensors. For example, GPS can be integrated with Forward looking infrared (FLIR) sensors and thermal sensors to realize more accurate detection solution \cite{rudol2008human}.
	
	\item While traditional machine learning techniques have demonstrated their success on UAVs, deep learning techniques are currently off limits because of the limitations on the on-board processing capabilities and power resources on UAVs. Therefore, there is a need to design and implement on-board, low power, and efficient deep learning solutions in support of SAR operations using UAVs \cite{carrio2017review}.
	
	\item Design and implementation of power-efficient distributed algorithms for the real-time processing of UAV swarm captured videos, images, and sensing data \cite{bejiga2017convolutional}. 
	
	\item New lighter materials, efficient batteries and energy harvesting solutions can contribute to the potential use of UAVs in long duration missions \cite{vergouw2016drone}.
	
	\item Algorithms that support UAV autonomy and swarm coordination are needed. These algorithms include: flight route determination, path planning, collision avoidance and swarm coordination \cite{alexopoulos2013comparative,pham2015survey}. 
	
	\item In Multi-UAV systems, there are many coordination and communications challenges that need to be overcome. These challenges include QoS communications between the swarm of UAVs over multi-hop communications links and with the GCS \cite{scherer2015autonomous}.
	
	\item More accurate localization and mapping systems and algorithms are required in support of SAR operations. Nowadays, GPS is used in UAVs to locate the coordinates UAVs and target objects but GPS is known to have coverage and accuracy issues. Therefore, new algorithms are needed for data fusion of the data received from multiple sensors to achieve more precise localization and mapping without coverage disruptions.   
	
	\item The use of UAVs as aerial base stations is still nascent. Therefore, more research is needed to study the use of such systems to provide communications coverage when the public communications network is disrupted or operating above its maximum capacity \cite{al2014modeling}.
	
\end{itemize}

\section{Remote Sensing}
UAVs can be used to collect data from ground sensors and deliver the collected data to ground base stations~\cite{tuyishimire2017cooperative}. UAVs equipped with sensors can also be used as aerial sensor network for environmental monitoring and disaster management~\cite{quaritsch2010networked}. Numerous datasets originating from UAVs remote sensing have been acquired to support the research teams, serving a broad range of applications:
crop monitoring, yield estimates, drought monitoring, water quality monitoring, tree species, disease detection, etc~\cite{vito}. In this section, we present UAV remote sensing systems, challenges during aerial sensing using UAVs, research trends and future insights.
\begin{figure*}[t]
	\centering
	\includegraphics[scale=0.4]{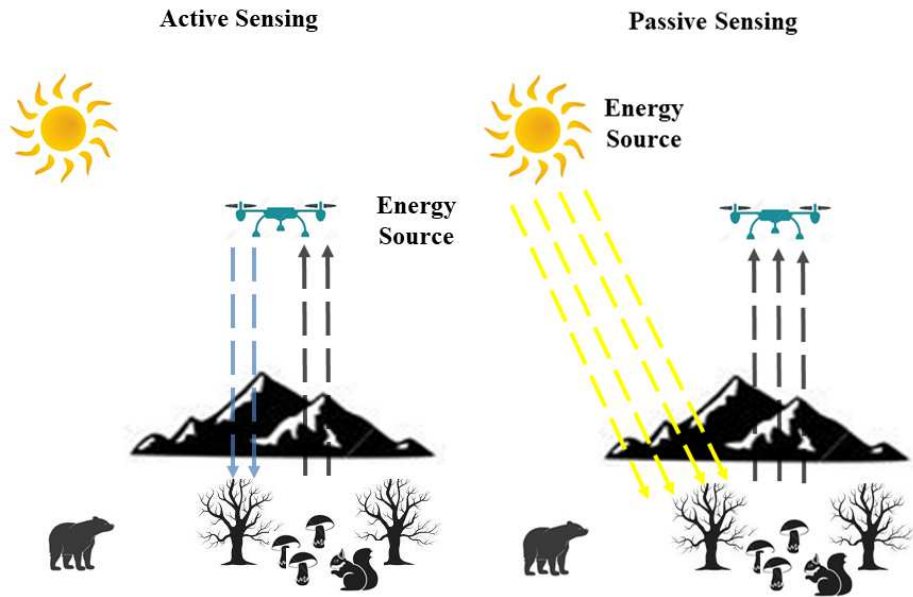}
	\caption{Active Vs. Passive Remote Sensing.}
	\label{ha1}
\end{figure*}

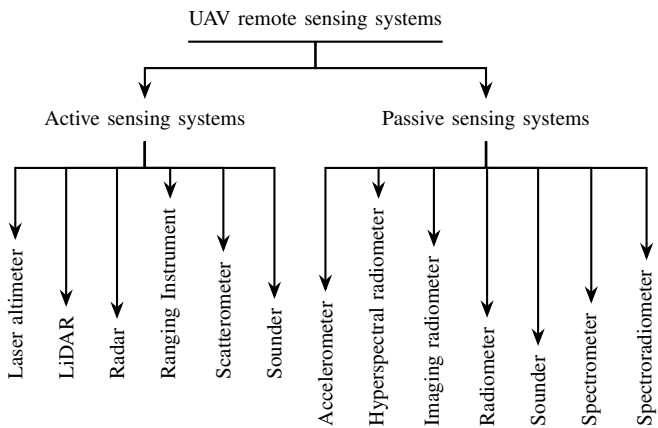
\begin{figure}
	\centering
	\begin{forest}
		for tree={
			align=center,
			parent anchor=south,
			child anchor=north,
			font=\footnotesize,
			edge={thick, -{Stealth[]}},
			l sep+=10pt,
			edge path={
				\noexpand\path [draw, \forestoption{edge}] (!u.parent anchor) -- +(0,-10pt) -| (.child anchor)\forestoption{edge label};
			},
			if level=0{
				inner xsep=0pt,
				tikz={\draw [thick] (.south east) -- (.south west);}
			}{}
		}
		[  UAV remote sensing systems 		
		[Active sensing systems 
		[\rotatebox{90}{Laser altimeter}		
		]
		[\rotatebox{90}{LiDAR}		
		]
		[\rotatebox{90}{Radar}		
		]
		[\rotatebox{90}{Ranging Instrument}		
		]
		[\rotatebox{90}{Scatterometer}
		]
		[\rotatebox{90}{Sounder}	
		]
		]
		[Passive sensing systems 
		[\rotatebox{90}{Accelerometer}
		]
		[\rotatebox{90}{Hyperspectral radiometer}		
		]
		[\rotatebox{90}{Imaging radiometer}		
		]
		[\rotatebox{90}{Radiometer}		
		]
		[\rotatebox{90}{Sounder}		
		]
		[\rotatebox{90}{Spectrometer}
		]
		[\rotatebox{90}{Spectroradiometer}
		]
		]
		]
	\end{forest}
	\caption{Classification of UAV Aerial Sensing Systems.}
	\label{ha2}
\end{figure}
\subsection{Remote Sensing Systems}
There are two primary types of remote sensing systems: active and passive remote sensing systems~\cite{NASA}. In active remote sensing system, the sensors are responsible for providing the source of energy required to detect the objects. The sensor transmits radiation toward the object to be investigated, then the sensor detects and measures the radiation that is reflected from the object. 
Most active remote systems used in remote sensing applications operate in the microwave portion of the electromagnetic spectrum and hence it makes them able to propagate through the atmosphere under most conditions~\cite{NASA}. The active remote sensing systems include laser altimeter, LiDAR, radar, ranging instrument, scatterometer and sounder. In passive remote sensing system, the sensor detects natural radiation that is emitted or reflected by the object as shown in Figure~\ref{ha1}.  The majority of passive sensors operate in the visible, infrared, thermal infrared, and microwave portions of the electromagnetic spectrum~\cite{NASA}. The passive remote sensing systems include accelerometer, hyperspectral radiometer, imaging radiometer, radiometer, sounder, spectrometer and spectroradiometer. In Figure~\ref{ha2}, we show the classifications of UAV aerial sensing systems. In Table~\ref{table1Remote}, we make a comparison among the UAV remote sensing systems based on the operating frequency and applications. The common active sensors in remote sensing are LiDAR and radar. A LiDAR sensor directs a laser beam onto the surface of the earth and determines the distance to the object by recording the time between transmitted and backscattered light pulses. A radar sensor produces a two-dimensional image of the surface by recording the range and magnitude of the energy reflected from all objects. The common passive sensor in remote sensing is spectrometer. A spectrometer sensor is designed to detect, measure, and analyze the spectral content of incident electromagnetic radiation.
\subsection{Image Processing and Analysis}
\begin{figure}
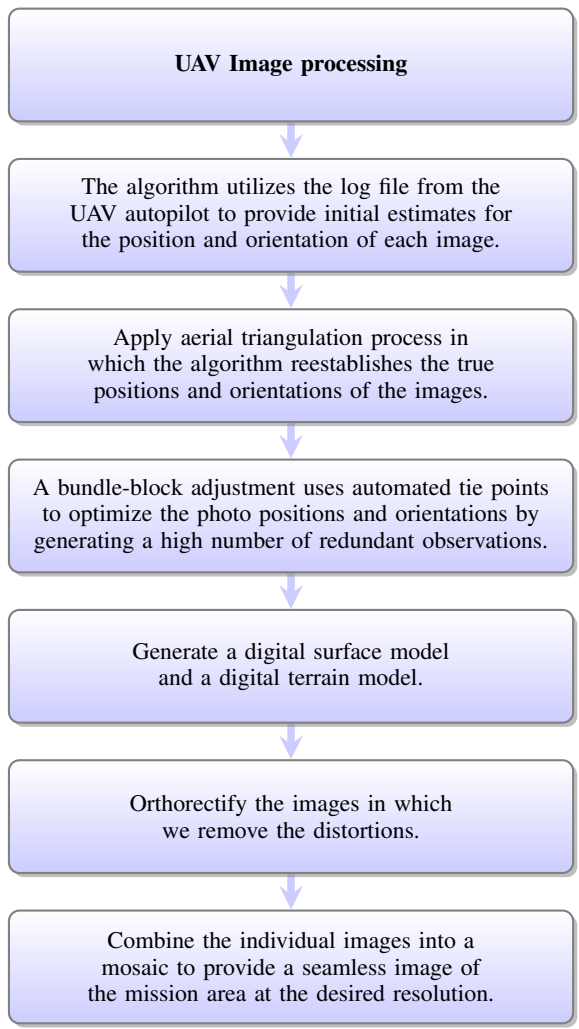

	
	\begin{center}
		\scriptsize
		\smartdiagramset{
			uniform color list=blue!20!white for 7 items,
			back arrow disabled=true,
			module minimum width=7.5cm,
			module minimum height=1.5cm,
			module y sep=2.0cm,
			text width=7.3cm,
		}
		\smartdiagram[flow diagram:vertical]{\textbf{UAV Image processing} 
			,The algorithm utilizes the log file from the UAV autopilot to provide initial estimates for the position and orientation of each image.
			,Apply aerial triangulation process in which the algorithm reestablishes the true positions and orientations of the images.,A bundle-block adjustment uses automated tie points to optimize the photo positions and orientations by generating a high number of redundant observations.,Generate a digital surface model and a digital terrain model.,Orthorectify the images in which we remove the distortions.,Combine the individual images into a mosaic to provide a seamless image of the mission area at the desired resolution.
		}
	\end{center}
	\caption{Image Processing for a UAV Remote Sensing Image}
	\label{ha3}
\end{figure}

The image processing steps for a typical UAV mission are described in details by the authors in~\cite{hugenholtz2013geomorphological} and~\cite{w2013low}. The process flow is the same for most remotely sensed imagery processing algorithms. First, the algorithm utilizes the log file from the UAV autopilot to provide initial estimates for the position and orientation of each image. The algorithm then applies aerial triangulation process in which the algorithm reestablishes the true positions and orientations of the images from an aerial mission. During this process, the algorithm generates a large number of automated tie points for conjugate points identified across multiple images. A bundle-block adjustment then uses these automated tie points to optimize the photo positions and orientations by generating a high number of redundant observations, which are used to derive an efficient solution through a rigorous least-squares adjustment. To provide an independent check on the accuracy of the adjustment, the algorithm includes a number of check points. Then, the oriented images are used to create a digital surface model, which provides a detailed representation of the terrain surface, including the elevations of raised objects, such as trees and buildings. The digital surface model generates a dense point cloud by matching features across multiple image pairs~\cite{w2013low}. At this stage, a digital terrain model can be generated, which is referred to as a bare-earth model. A digital terrain model is a more useful product than a surface model, because the high frequency noise associated with vegetation cover is removed. After the algorithm generates a digital terrain model, orthorectification process can then be performed to remove the distortion in the original images. After orthorectification process, the algorithm combines the individual images into a mosaic, to provide a seamless image of the mission area at the desired resolution~\cite{whitehead2014remote}. Figure~\ref{ha3} summarizes the image processing steps for remotely sensed imagery. 
\begin{table*}[!h]
	\scriptsize
	\renewcommand{\arraystretch}{1}
	\caption{\uppercase{UAV aerial sensing systems}}
	\label{table1Remote}
	\centering
	\begin{tabular}{|c|c|c|c|}
		\hline 
		Type of remote sensing&Operating Frequency&Type of sensor& Applications\\ 
		\hline
		&&Laser altimeter&It measures the height of a UAV with respect to the mean Earth’s\\
		&&&surface to determine the topography of the underlying surface.~~~~\\
		\hhline{|~|~|-|-|} 
		&&LiDAR&It determines the distance to the object by recording the time~~  \\
		&&&between transmitted and backscattered light pulses.~~~~~~~~~~~~~~~~~\\ 
		\hhline{|~|~|-|-|} 
		&&Radar &It produces a two-dimensional image of the surface by recording \\
		Active	&Microwave portion of the &&the range and magnitude of the energy reflected from all objects.\\ 	
		\hhline{|~|~|-|-|} 
		&electromagnetic spectrum &Ranging Instrument &It determines the distance between identical microwave~~~~~~~~~~~~  \\
		&&&instruments on a pair of platforms.~~~~~~~~~~~~~~~~~~~~~~~~~~~~~~~~~~~~~\\ 	
		\hhline{|~|~|-|-|} 
		&&Scatterometer &It derives maps of surface wind speed and direction by measuring \\
		&&& backscattered radiation in the microwave spectral region.~~~~~~~~~~~~\\ 
		\hhline{|~|~|-|-|} 
		&&Sounder &It measures vertical distribution of precipitation, temperature,~~~~~\\
		&&& humidity, and cloud composition.~~~~~~~~~~~~~~~~~~~~~~~~~~~~~~~~~~~~~~\\ 
		\hline	
		&&Accelerometer &It measures two general types of accelerometers: 1) The~~~~~~~~~~   \\
		&&&translational accelerations (changes in linear motions); 2) The~~~~\\ 
		&&&angular accelerations (changes in rotation rate per unit time).~~~~~\\
		\hhline{|~|~|-|-|} 
		&&Hyperspectral radiometer &It discriminates between different targets based on their spectral~\\
		&&&  response in each of the narrow bands.~~~~~~~~~~~~~~~~~~~~~~~~~~~~~~~\\
		\hhline{|~|~|-|-|} 
		&&Imaging radiometer &It provides a two-dimensional array of pixels from which ~~~~~~~\\
		&&&  an image may be produced.~~~~~~~~~~~~~~~~~~~~~~~~~~~~~~~~~~~~~~~~~~~\\
		\hhline{|~|~|-|-|} 
		Passive&Visible, infrared, thermal &Radiometer&It measures the intensity of electromagnetic radiation in some~~~\\
		&infrared, and microwave &&  bands within the spectrum.~~~~~~~~~~~~~~~~~~~~~~~~~~~~~~~~~~~~~~~~~~~~\\
		\hhline{|~|~|-|-|} 
		&portions of the  &Sounder&It measures vertical distributions of atmospheric parameters~~~ \\
		& electromagnetic spectrum&&such as  temperature, pressure, and composition from~~~~~~~~~~~ \\
		&&&multispectral information.~~~~~~~~~~~~~~~~~~~~~~~~~~~~~~~~~~~~~~~~~~~\\
		\hhline{|~|~|-|-|} 
		&&Spectrometer &It designs to detect, measure, and analyze the spectral content\\
		&&&   of incident electromagnetic radiation.~~~~~~~~~~~~~~~~~~~~~~~~~~~~~~~\\
		\hhline{|~|~|-|-|} 
		&&Spectroradiometer &It measures the intensity of radiation in multiple wavelength~~  \\
		&&&  bands. It designs for remotely sensing specific geophysical~~~ \\
		&&&parameters.~~~~~~~~~~~~~~~~~~~~~~~~~~~~~~~~~~~~~~~~~~~~~~~~~~~~~~~~~~~~\\
		\hline
	\end{tabular}
\end{table*}

\subsection{Flight Planning}
Although each UAV mission is unique in nature, the same steps and processes are normally followed. Typically, a UAV mission starts with flight planning~\cite{hugenholtz2013geomorphological}. This step depends on specific flight-planning algorithm and uses a background map or satellite image as a reference to define the flight area. Extra data is then included, for example, the desired flying altitude, the focal length and orientation of the camera, and the desired flight path. The flight-planning algorithm will then find an efficient way to obtain overlapping stereo imagery covering the area of interest. During the flight-planning process, the algorithm can adjust various parameters until the operator is satisfied with the flight plan. As part of the mission planning process, the camera shutter speed settings must satisfy the different lighting conditions. If exposure time is too short, the imagery might be too dark to discriminate among all key features of interest, but if it is too long, the imagery will be blurred or will be bright.  Next, the generated flight plan is uploaded to the UAV autopilot. The autopilot uses the instructions contained in the flight plan to find climb rates and positional adjustments that enable the UAV to follow the planned path as closely as possible. The autopilot reads the adjustments from the global navigation and satellite system and the initial measurement unit several times per second throughout the flight. After the flight completion, the autopilot download a log file, this file contains information about the recorded UAV 3D placements throughout the flight, as well as information about when the camera was triggered. The information in log file is used to provide initial estimates for image centre positions and camera orientations, which are then used as inputs to recover the exact positions of surface points~\cite{whitehead2014remote}. 
\subsection{Challenges}
\subsubsection{{Hostile Natural Environment}}
UAVs can be utilized to study the atmospheric composition, air quality and climate parameters, because of their ability to access hazardous environments, such as thunderstorms, hurricanes and volcanic plumes~\cite{austin2011unmanned}. The researchers used UAVs for conducting environmental sampling and ocean surface temperature studies in the Arctic~\cite{villa2016overview}. The authors in~\cite{curry2004applications} modify and test the Aerosonde UAV in extreme weather conditions, at very low temperatures (less than $-20\,^{\circ}\mathrm{C}$) to ensure a safe flight in the Arctic. The aim of the work was to modify and integrate sensors on-board an Aerosonde UAV to improve the UAV’s capability for its mission under extreme weather conditions such as in the Arctic. The steps to customize the UAV for the extreme weather conditions were: 1) The avionics were isolated; 2) A fuel-injection engine was used to avoid carburetor icing; 3) A servo-system was adopted to force ice breaking over the leading edge of the air-foil. In Barrow, Alaska, the modified UAVs successfully demonstrated their capabilities to collect data for 48 hours along a 30 $km^{2}$ rectangular geographical area. When a UAV collects data from a volcano plume, a rotary wing UAV was particularly beneficial to hover inside the plume~\cite{mcgonigle2008unmanned}. On the other hand, a fixed wing UAV was suitable to cover longer distances and higher altitudes to sense different atmospheric layers~\cite{saggiani2007uav}.  In~\cite{lin2008eyewall}, the authors presented a successful eye-penetration reconnaissance flight by Aerosonde UAV into Typhoon Longwang (2005). The 10 hours flight was split into four flight legs. In these flight legs, the UAV measured the wind field and provided the tangential and radial wind profiles from the outer perimeter into the eye of the typhoon at the 700 hPa layer. The UAV also took a vertical sounding in the eye of the typhoon and measured the strongest winds during the whole flight mission~\cite{villa2016overview}. 

\subsubsection{{Camera Issues}}
The radiometric and geometric limitations imposed by the current
generation of lightweight digital cameras are outstanding issues that need to be addressed. The current UAV digital cameras are designed for the general market and are not optimized for remote sensing applications. The current commercial instruments tend to be too bulky to be used with current lightweight UAVs, and for those that do exist, there is still a question of calibration process with conventional sensors. Spectral drawbacks include the fact that spectral response curves from cameras are usually poorly calibrated, which makes it difficult to convert brightness values to radiance. However, even cameras designed specifically for UAVs may not meet the required scientific benchmarks~\cite{whitehead2014remote}. Another drawback is that the detectors of camera may also become saturated when there are high contrasts, for example when an image covers both a dark forest and a snow covered field. Another drawback is that many cameras are prone to vignette, where the centres of images appear brighter than the edges. This is because rays of light in the centres of the image have to pass through a less optical thickness of the camera lens, and are thus low attenuated than rays at the edges of the image. There are a number of techniques that can be taken into account to improve the quality of image: 1) micro-four-thirds cameras with fixed interchangeable lenses can be used instead of having a retractable lens, which allows for much improved calibrations and image quality; 2) a simple step that can make a big difference in the processing stage is to remove images that are blurred, under or overexposed, or saturated~\cite{whitehead2014remote}.

\subsubsection{{Illumination Issues}}
The shadows on a sunny day are clear and well defined. These weather conditions can cause critical problems for the automated image matching algorithms used in both triangulation process and digital elevation model generation~\cite{whitehead2014remote}. When clouds move rapidly, shaded areas can vary between images obtained during the same mission, therefore the aerial triangulation process will fail for some images, and also will result in errors for automatically generated digital elevation models. Furthermore, the automated color balancing algorithms used in the creation of image mosaics may be affected by the patterns of light and shade across images. This can cause mosaics with poor visual quality. Another generally observed illumination effect is the presence of image hotspots, where a bright points appear in the image. Hotspots occur at the antisolar point due to the effects of bidirectional reflectance, which is dependent on the relative placement of the image sensor and the sun~\cite{whitehead2014remote}.
\subsection{Research Trends and Future Insights}
\subsubsection{{Machine Learning}}
\begin{figure}
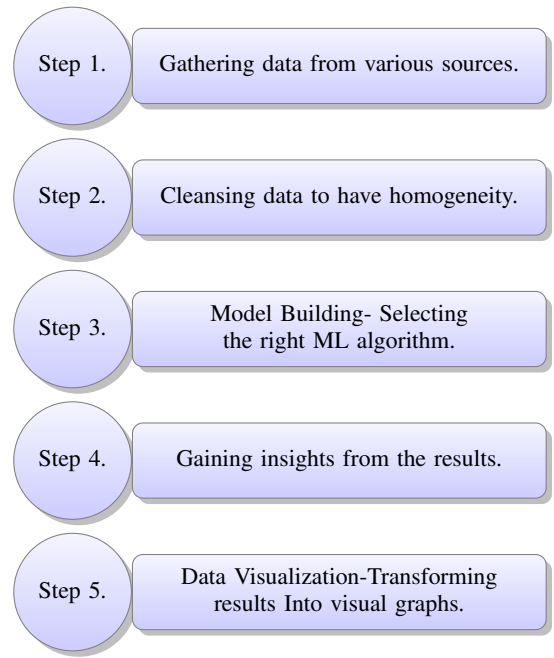

	\begin{center}
		\begin{center}
			\smartdiagram[descriptive diagram]
			{
				{Step 1.,Gathering data from
					various sources.
				},
				{Step 2., {Cleansing data to have homogeneity.
				}},
				{Step 3., Model Building- Selecting the right 
					ML algorithm.
				},
				{Step 4., {Gaining insights from the results.
				}},
				{Step 5.,Data Visualization-Transforming results
					Into visual graphs.
				},
			}
		\end{center}
		
	\end{center}
	
	\caption{ Machine Learning in UAV Remote Sensing. }
	\label{ha4}
\end{figure}

In remote sensing, the machine learning process begins with data collection using UAVs. The next step of machine learning is data cleansing, which includes cleansing up image and/or textual-based data and making the data manageable. This step sometimes might include reducing the number of variables associated with a record. The third step is selecting the right algorithm, which includes getting acquainted with the problem we are trying to solve. There are three famous algorithms being used in remote sensing:1) Random forest; 2) Support vector machines; 3) Artificial neural networks. An algorithm is selected depending on the type of problem being solved. In some scenarios, where there are multiple features but limited records, support vector machines might work better. If there are a lot of records but less features, neural networks might yield a better prediction/classification accuracy. Normally, several algorithms will be applied on a dataset and the one that works best is selected. In order to achieve a higher accuracy of the machine learning results, a combination of multiple algorithms can also be employed, which is referred to as ensemble. Similarly, multiple ensembles will need to be applied on a dataset, in order to select the ensemble that works the best. It is practical to choose a subset of candidate algorithms based on the type of problem and then use the narrowed down algorithms on a part of the dataset and see which one performs best. The first challenge in machine learning is that the training segment of the dataset should have an unbiased representation of the whole dataset and should not be too small as compared to the testing segment of the dataset. The second challenge is overfitting which can happen when the dataset that has been used for algorithm training is used for evaluating the model. This will result in a very high prediction/classification accuracy. However, if a simple modification is performed, then the prediction/classification accuracy takes a dip~\cite{pythontips}. The machine learning steps utilized by UAV remote sensing are shown in Figure~\ref{ha4}.

\subsubsection{{Combining Remote Sensing and Cloud Technology}}
Use of digital maps in risk management as well as improving data visualization and decision making process has become a standard for businesses and insurance companies. For instance, the insurance company can utilize UAV to generate a normalized difference vegetation index (NDVI) map in order to have an overview of the hail damage in corn. The geographic information system (GIS) technology in the cloud utilizes the NDVI map generated from UAV images to provide an accurate and advanced tool for assisting with crop hail damage insurance settlements in minimal time and without conflict, while keeping expenses low~\cite{giscloud}.
\subsubsection{{Free Space Optical}}
FSO technology over an UAV can be utilized in armed forces, where military wireless communications demand for secure transmission of information on the battlefield. Remote sensing UAVs can utilize this technology to disseminate large amount of images and videos to the fighting forces, mostly in a real time. Near Earth observing UAVs can be utilized to provide high resolution images of surface contours using synthetic aperture radar and light detection and ranging. Using FSO technology, aerial sensors can also transmit the collected data to the command center via on-board satellite communication sub-system~\cite{kaushal2017optical}.
\subsubsection{Future Insights }

Some of the future possible directions for this application are: 

\begin{itemize}
	\item Camera stabilization during flight~\cite{whitehead2014remote} is one of the issues that needs to be addressed, in the employment of UAV for remote sensing.
	\item Battery weight and charging time are critical issues that affect the duration of UAV missions~\cite{madden15future}. The development and incorporation of lightweight solar powered battery components of UAV can improve the duration of UAV missions and hence it reduces the complexity of flight planning. 
	\item The temporal digital surface models produced from aerial imagery using UAV as the platform, can become practical solution in mass balance studies for example mass balance of any particular metal in sanitary landfills, chloride in groundwater and sediment in a river. More specifically, the UAV-based mass balance in a debris-covered glacier was estimated at high accuracy, when using high-resolution digital surface models differencing. Thus, the employment of UAV save time and money when compared with the traditional method of stake drilling into the glaciers which was labor-intensive and time consuming ~\cite{immerzeel2014high,bhardwaj2016uavs}.
	\item The tracking methods utilized on high-resolution images can be practical to find an accurate surface velocity and glacier dynamics estimates. In~\cite{sam2016remote}, the authors suggested a differential band method for estimating velocities of debris and non-debris parts of the glaciers. The on-demand deployment of UAV to obtain high-resolution images of a glacier has resulted in an efficient tracking methods when compared with satellite imagery which depends on the satellite overpass~\cite{bhardwaj2016uavs}.
	\item UAV remote sensing can be as a powerful technique for field-based
	phenotyping with the advantages of high  efficiency, low cost and suitability for complex environments. The adoption of multi-sensors coupled with advanced data analysis techniques for retrieving crop phenotypic traits have attracted great attention in recent years~\cite{yang2017unmanned}.
	\item It is expected that with the advancement of UAVs with larger
	payload, longer flight time, low-cost sensors, improved
	image processing algorithms for Big data, and effective
	UAV regulations, there is potential for wider applications
	of the UAV-based field crop phenotyping~\cite{yang2017unmanned}.
	\item UAV remote sensing for field-based crop phenotyping provides data at high resolutions, which is needed for accurate crop parameter estimations. The derivation of crop phenotypic traits based on the spectral reflection information using UAV as the platform has shown good accuracy under certain conditions. However, it showed a low accuracy in the research on the non-destructive acquisition of complex traits that were indirectly related to the spectral information~\cite{yang2017unmanned}; 
	\item Image processing of UAV imagery faces a number of challenges, such as variable scales, high amounts of overlap, variable image orientations, and high amounts of relief displacement arising from the low flying altitudes relative to the variation in topographic relief~\cite{whitehead2014remote}. Researchers need to find efficient ways to overcome these challenges in future studies.
\end{itemize}

\section{Construction \& Infrastructure Inspection}

As already mentioned in the market opportunity section \ref{Market}, the net market value of the deployment of UAV in support of construction and infrastructure inspection applications is about 45\% of the total UAV market. So there is a growing interest in UAV uses in large construction projects monitoring \cite{liu2014review} and power lines, gas pipelines and GSM towers infrastructure inspection \cite{deng2014unmanned,mohamadi2014vertical}. In this section, we first present a literature review. Then, we show the uses of UAVs in support of infrastructure inspection. Finally, we present the challenges, research trends and future insights.

\subsection{Literature Review}

In construction and infrastructure inspection applications, UAVs can be used for real-time monitoring construction project sites   \cite{gheisari2014uas4safety}. So, the project managers can monitor the construction site using UAVs with better visibility about the project progress without any need to access the site \cite{liu2014review}.

Moreover, UAVs can also be utilized for high voltage inspection of the power transmission lines. In  \cite{jones2005power,luque2014power,sampedro2014supervised,li2010towards}, the authors used the UAVs to perform an autonomous navigation for the power lines inspection. The UAVs was deployed to detect, inspect and diagnose the defects of the power line infrastructure.

In \cite{larrauri2013automatic}, the authors designed and implemented a fully automated UAV-based system for the real-time power line inspection. More specifically, multiple images and data from UAVs were processed to identify the locations of trees and buildings near to the power lines, as well as to calculate the distance between trees, buildings and power lines. Furthermore, TIR camera was employed for bad conductivity detection in the power lines.
UAVs can also be used to monitor the facilities and infrastructure, including gas, oil and water pipelines. In \cite{mohamadi2014vertical}, the authors proposed the deployment of small-UAV (sUAV) equipped with a gas controller unit to detect air and gas content. The system provided a remote sensing to detect gas leaks in oil and gas pipelines.

Table~\ref{cons_t1} summarizes some of the construction and infrastructure inspection applications using UAVs. More specifically, this table presents several types of UAV used in construction and infrastructure inspection applications, as well as the type of sensors deployed for each application and the corresponding UAV specifications in terms of payload, altitude and endurance. 

\begin{table*}[!h]
	\footnotesize
	\renewcommand{\arraystretch}{1.7}
	\caption{\uppercase{Summary of UAV specifications, applications and technology used in construction and infrastructure inspection}}
	\label{cons_t1}
	\centering
	\begin{adjustbox}{width=1.0\textwidth}
		\begin{tabu} to 1.0\textwidth { | X[l] | X[l] | X[l] |X[l] |X[l] | }
			\hline 
			\textbf{UAV Type} & \textbf{Applications}& {Payload}\textbf{/}{Altitude}\textbf{/}{Endurance}&\textbf{Sensor Type}&\textbf{References} \\
			\hline \hline
			AR.Drone French Company Parrot.
			& 
			Use UAVs to enhance the safety on on construction sites by providing a real-time visual view for these sites.
			& 
			No /  50 $m$ / 12 min.& On-board HD camera, Wi-Fi connection.
			&\cite{gheisari2014uas4safety}
			\\
			\hline
			A Multi Rotor UAV.
			& 
			Use UAVs with image processing methods for crack detection and assessment of surface degradation.
			& 
			100 g / LAP / 20 min.& Color imaging sensors.
			&\cite{sankarasrinivasan2015health}
			\\
			\hline
			MikroKopter L4-ME Quadcopter \cite{MikroKopter-L4-ME:Online}.
			& 
			Use UAVs for vertical inspection for high rise infrastructures such as street lights, GSM towers or high rise buildings.
			& 
			Up to 500 g /  Up to 247 $m$ / 13-20 min. \cite{MikroKopter-L4-ME:Online}& Laser scanner.
			&\cite{sa2014vertical}
			\\
			\hline
			Quadrotor Helicopter, UAV.
			& 
			Inspection of the high voltage of power transmission lines.
			& 
			Less than 1 kg/ LAP/ Less than 1 hour. & Color and TIR cameras, GPS, IMU.
			&\cite{luque2014power}
			\\
			\hline
			
			Fixed Wing Aircraft, UAV.
			& 
			Sketchy inspection, identifying the defects of the power transmission lines.
			& 
			Less than 3 kg/ Up to 500 $m$/ Up to 50 min (50 $km$). & HD ultra-wide angle video camera.
			&\cite{deng2014unmanned}
			\\
			\hline
			Quadrotor, UAV.
			& 
			
			Use cooperative UAVs platform for inspection and diagnose of the power lines infrastructure.
			& 
			Less than 6 kg/ Up to 200 $m$/ Up to 25 min (10 $km$). & TIR cameras, GPS.
			&\cite{deng2014unmanned}
			\\
			\hline
			
			Quadrotor (VTOL), sUAV.
			& 
			provide a remote sensing to detect gas leaks in gas pipelines.
			& 
			NA/ LAP/ 30-50 min. & Gas controller unit, GPS.
			&\cite{mohamadi2014vertical,bretschneider2015uav}
			\\
			\hline

		\end{tabu}
		
	\end{adjustbox}
	
	\label{table:ahmad1-4}
\end{table*}

\subsection{The Deployment of UAVs  for Construction \& Infrastructure Inspection Applications}

\begin{figure*}[h] 
	\centering
	\begin{tikzpicture}[font=\scriptsize]
	\tikzset{every node/.style=
		{align=center, minimum height=46pt, text width=80pt}}
	\node[,draw=black] (b1) {Oil/gas and wind turbine inspection
	};
	\node[right=5pt,draw=black] (b2) at (b1.east) {Critical land 
		building inspection 
		(e.g., cell tower)
	};
	\node[right=5pt,draw=black] (b3) at (b2.east) {Infrastructure 
		internal inspection
		(e.g., pipe)
	};  
	\node[right=5pt,draw=black] (b4) at (b3.east) {Extreme condition inspection
	};
	\node[above=10pt, text width=160pt,draw=black] (top) at ($(b2.north)!.5!(b3.north)$) {\small{Construction 
			\& Infrastructure Inspection Using UAVs}};
	\coordinate (atop) at ($(top.south) + (0,-5pt)$);
	\coordinate (btop) at ($(b3.south) + (0,-5pt)$);
	\draw[thick] (top.south) -- (atop)
	(b1.north) |- (atop) -| (b4.north)
	(b2.north) |- (atop) -| (b3.north);
	
	
	
	
	\end{tikzpicture}
	\caption{The Deployment of UAVs for Construction 
		and Infrastructure Inspection.
	}
	\label{uses}
\end{figure*}
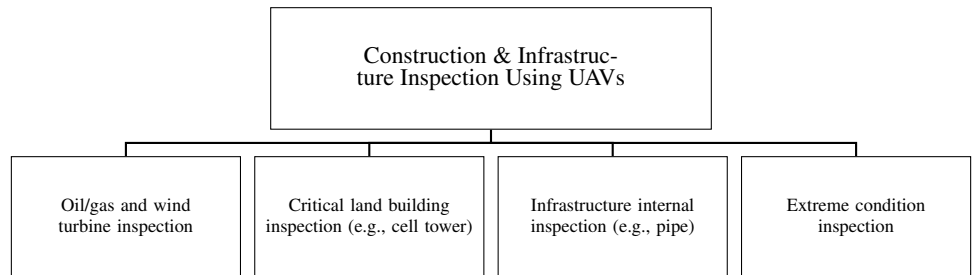

In this section, we present several specific example of the deployment of UAV for construction and infrastructure inspection. Figure \ref{uses} illustrates the classification of these deployments.


\begin{itemize}
	\item {Oil/gas and wind turbine inspection:} 
	In 2016, it was reported that Pacific Gas and Electric Company (PG\&E) performed drone tests to inspect its electric and gas services for better safety and reliability with the authorization from  Federal Aviation Administration (FAA) \cite{peg2016}. The inspections focused on hard-to-reach areas to detect methane leaks across its 70,000-square-mile service area. In the future, PG\&E plans to extend the drone tests for storm and disaster response.
	
	Cyberhawk is one of world's premier oil and gas company that uses UAV for inspection\cite{Cyberhawk2008}. Furthermore, it has completed more than 5,000 structural inspections, including: 1) oil and gas; 2) wind turbine; and 3) live flare. UAV inspection conducted by Cyberhawk provides a bunch of photos, as well as conducts close visual and thermal inspections of the inspected assets.
	
	Industrial SkyWorks \cite{SkyWorks2018} uses drones for building inspections and oil/gas inspections in North America. Furthermore, a powerful machine learning algorithm, BlueVu, has been developed to efficiently handle the captured data. To sum up, it provides the following solutions:
	\begin{itemize}
		\item Asset inspections and data acquisition;
		\item Advanced data processing with 2D and 3D images;
		\item Detailed reports of the inspected asset (i.e., annotations, inspector comments, and recommendations).
	\end{itemize}
	
	\item {Critical land building inspection (e.g., cell tower):}
	AT\&T owns about 65,000 cell towers that need to be inspected, repaired or installed. The video analytic team at AT\&T Labs has collaborated with other forces (e.g., Intel, Qualcomm, etc.) to develop faster, better, more efficient, and fully automated cell tower inspection using UAVs \cite{att2017}. One of the approaches is to employ deep learning algorithm on high definition (HD) videos to detect defects and anomalies in real time.

	Honeywell InView inspection service has been launched to provide industrial critical structure inspections \cite{honeywell2017}. It combines the Intel® Falcon™ 8+ UAV system with Honeywell aerospace and industrial technology solutions. Specifically, the Honeywell InView inspection service can achieve: 1) safety of the workers; 2) improved efficiency; and 3) advanced data processing.

	\item {Infrastructure internal inspection:}
	Maverick has provided industrial UAV inspection services for equipment, piping, tanks, and stack internals in western Canada since 1994~\cite{Maverick2018}. It provides a dedicated services for assets internal inspection using Flyability ELIOS. Maverick also provides post data processing that analyzes data using measurement software and CAD modeling.
	
	\item {Extreme condition inspection:}
	Bluestream offers UAV inspection services for onshore and offshore assets \cite{Alongside2018}. It services are particularly suitable for: 1) onshore and offshore live flare inspections; 2) topside, splash zone and under deck inspections; and 3) hard to access infrastructure inspection.
	
\end{itemize}

\subsection{Challenges}
There are several challenges in utilizing UAVs for construction and infrastructure inspection:
\begin{itemize}
	
	\item Some of the challenges in using UAVs for infrastructure inspection are the limited energy, short flight time and limited processing capabilities \cite{gheisari2014uas4safety}.
	
	\item Limited payload capacities for sUAVs is a big challenge. The on-board loads could include optical wavelength range camera, TIR camera, color and stereo vision cameras, different types of sensors such as gas detection, GPS, etc., \cite{bretschneider2015uav}.

	\item There is a lack of research attention to multi-UAV cooperation for construction and infrastructure inspection applications. Multi-UAV cooperation could provide wider inspection scope, higher error tolerance, and faster task completion time. 
	
	\item Another challenge is to allow autonomous UAVs that can maneuver an indoor environment with no access to GPS signals \cite{dupont2017potential}.
	
\end{itemize}

\subsection{Research Trends and Future Insights}

\subsubsection{Machine Learning}
Machine learning has become an increasingly important artifical intelligence approach for UAVs to operate autonomously. Applying advanced machine learning algorithms (e.g., deep learning algorithm) could help the UAV system to draw better conclusions. For example, due to its improved data processing models, deep learning algorithms could help to obtain new findings from existing data and to get more concise and reliable analysis results. UAV inspection program at AT\&T uses deep learning algorithms on HD videos to detect defects and anomalies in real time~\cite{att2017}. Industrial SkyWorks introduces advanced machine learning algorithms to process 2D and 3D images \cite{SkyWorks2018}. 

\begin{figure}[!h]
	\centering
	\includegraphics[scale=0.25]{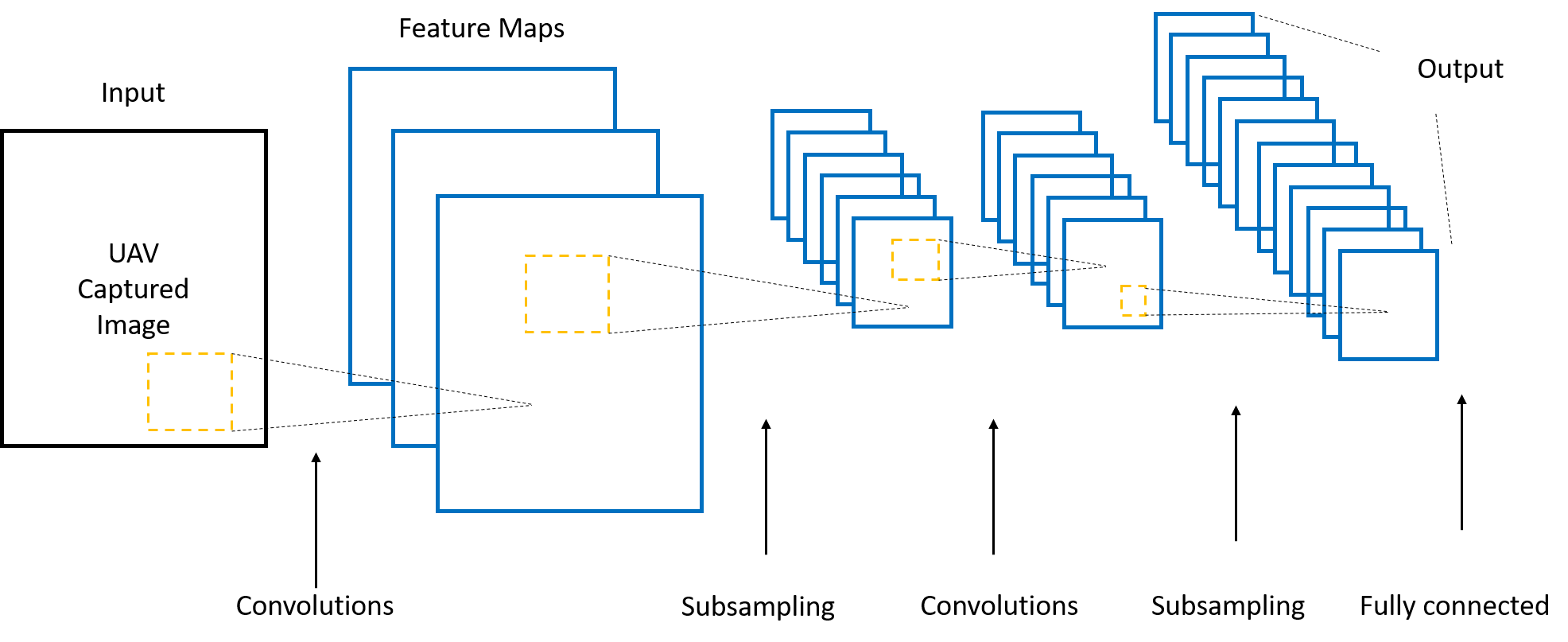}
	\caption{Illustration of How CNNs Work~\cite{carrio2017review}.}
	\label{cnn}
\end{figure}

More specifically, deep learning is useful for feature extraction from the raw measurements provided by sensors on-board a UAV (details about UAV sensor technology is presented in the Precision Agriculture section). Convolutional Neural Networks (CNNs) is one of the main deep learning feature extractors used in the area of image recognition and classification which has been proven very effective \cite{carrio2017review}. Figure \ref{cnn} illustrates one example of how CNNs works.

\subsubsection{Image Processing}
Construction and infrastructure inspection using UAVs equipped with an on-board cameras and sensors, can be efficiently operated when employing image processing techniques. The employment of image processing techniques allows for monitoring and assessing the construction projects, as well as performing inspection of the infrastructure such as surveying construction sites, work progress monitoring,  inspection of bridges, irrigation structures monitoring, detection of construction damage and surfaces degradation.
\cite{sankarasrinivasan2015health,ham2016visual}.

\begin{figure}[!h]
	\centering
	\includegraphics[scale=0.3]{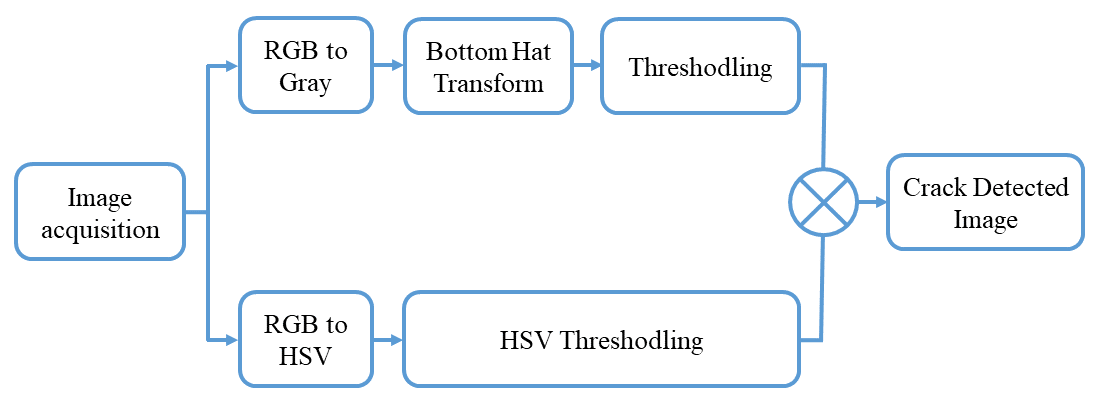}
	\caption{Proposed Approach for Crack Detection Algorithm~\cite{sankarasrinivasan2015health}.}
	\label{crack}
\end{figure}

In \cite{sankarasrinivasan2015health}, presented an integrated data acquisition and image processing platform mounted on UAV. It was proposed to be used for the inspection of infrastructures and real time structural health monitoring (SHM). In the proposed framework, a real time image and data will be sent to the GCS. Then the images and data will be processed using image processing unit in the GCS, to facilitate the diagnosis and inspection process.   
The authors proposed to combine HSV thresholding and hat transform for cracks detection on the concrete surfaces. Figure~\ref{crack} presents the crack detection algorithm block diagram proposed in \cite{sankarasrinivasan2015health}.

In \cite{luque2014power}, the authors proposed for the deployment of UAV with a vision-based system that consists of a color camera, TIR camera and a transmitter to send the captured images to the GCS. Then these images will be processed in the GCS, to be used in the inspection and estimation of the real temperature of the power lines joints.

In \cite{larrauri2013automatic}, a fully automatic system to determine the distance between power lines and the trees, buildings and any others obstacles was proposed. 
The authors designed and developed vision-based algorithms for processing the HD images that were obtained using HD camera equipped on a UAV. 
The captured video was sent to the computer in the GCS. At the GCS, the video was converted into the consecutive images and were further processed to calculate the distance between the power line and the obstacles.

\subsubsection{Future Insights}
\label{cons-ins}

Based on the reviewed articles on construction and infrastructure inspection applications using UAVs, we suggest these future possible directions:
\begin{itemize}
	\item More research is required to improve UAVs battery life time to allow longer distance and to increase the UAV flight time \cite{gheisari2014uas4safety}.
	
	\item For future research, it is important to propose and develop an accurate, autonomous and real-time power lines inspection approaches using UAVs, including ultrasonic sensors, TIR or color cameras, image processing and data analysis tools. More specifically, to propose and develop techniques to monitor, detect and diagnose any power lines defects automatically \cite{pagnano2013roadmap}.
	
	\item More advanced data collection, sharing and processing algorithms for multi-UAV cooperation are required, in order to achieve faster and more efficient inspections.
	
	\item The researchers should also focus on the improvement of  the autonomy and safety for UAVs to maneuver in the congested and indoor environment with no or weak GPS signals   \cite{dupont2017potential}.
	
\end{itemize}

\section{Precision Agriculture}

UAVs can be utilized in precision agriculture (PA) for crop management and monitoring \cite{huang2013development,muchiri2016review }, weed detection \cite{ kazmi2011adaptive }, irrigation scheduling \cite{gonzalez2013using}, disease detection \cite{ garcia2013comparison }, pesticide spraying \cite{huang2013development} and gathering data from ground sensors (moisture, soil properties, etc.,)  \cite{mathur2016data}. 
The deployment of UAVs in PA is a cost-effective and time saving technology which can help for improving crop yields, farms productivity and profitability in farming systems. Moreover, 
UAVs facilitate agricultural management, weed monitoring, and pest damage,  thereby they help to meet these challenges quickly \cite{primicerio2012flexible}.

In this section, we first present a literature review of UAVs in PA. Then, we show the deployment of UAV in PA. Moreover, we present the challenges, as well as research trends future insights.

\subsection{Literature Review}

UAVs can be efficiently used for small crop fields at low altitudes with higher precision and low-cost compared with traditional manned aircraft. Using UAVs for crop management can provide precise and real time data about specific location. Moreover, UAVs can offer a high resolution images for crop to help in crop management such as disease detection, monitoring agriculture, detecting the variability in crop response to irrigation, weed management and reduce the amount of herbicides \cite{muchiri2016review,jensen2003assessing,huang2013development,hunt2010acquisition,sullivan2007evaluating}. In Table~\ref{PA-t0}, a comparison between UAVs, traditional manned aircraft and satellite based system is presented in terms of system cost, endurance, availability, deployment time, coverage area, weather and working conditions, operational complexity, applications usage and finally we present some examples from the literature.

\begin{table}[!h]
	\scriptsize
	\renewcommand{\arraystretch}{1.7}
	\caption{\uppercase{A comparison between UAVs, traditional manned aircraft and satellite based system for PA}}
	\label{PA-t0}
	\centering
	\begin{tabular}{|c|c|c|c|}
		\hline
		\textbf{Issues}&UAVs &Manned Aircraft&Satellite System\\
		\hline
		Cost&Low&High&Very High\\
		\hline
		Endurance&Short-time&Long-time&All the times \\
		\hline 
		Availability&When needed&Some times&All the times \\
		\hline 
		Deployment time&Easy&Need runway&Complex \\
		\hline 
		Coverage area&Small&Large&Very large \\
		\hline 
		Weather and&Sensitive&Low sensitivity&Require clear sky \\
		working conditions&&&for imaging \\
		\hline 
		Payload&Low &Large&Large \\
		\hline
		Operational &Simple&Simple&Very complicated \\
		complexity&&& \\
		\hline
		Applications &Carry small&Spraying UAV  &high resolution   \\
		and usage&digital, thermal&system pesti-&images for \\
		&cameras \& sensors&cide spraying&specific-area \\
		\hline
		Examples &\cite{ muchiri2016review, jensen2003assessing }&\cite{ akesson1974use }&\cite{ reed1994measuring } \\
		\hline

	\end{tabular}
\end{table}

Table~\ref{PA_t1} summarizes some of the precision agriculture applications using UAVs. More specifically, this table presents several types of UAV used in precision agriculture applications, as well as the type of sensors deployed for each application and the corresponding UAV specifications in terms of payload, altitude and endurance. 

\begin{table*}[!h]
	\footnotesize
	\renewcommand{\arraystretch}{1.7}
	\caption{\uppercase{Summary of UAV specifications, applications and technology used in precision agriculture}}
	\label{PA_t1}
	\centering
	\begin{adjustbox}{width=1.0\textwidth}
		\begin{tabu} to 1.0\textwidth { | X[l] | X[l] | X[l] |X[l] |X[l] | }
			\hline 
			\textbf{UAV Type} & \textbf{Applications}& {Payload}\textbf{/}{Altitude}\textbf{/}{Endurance}&\textbf{Sensor Type}&\textbf{References} \\
			\hline \hline
			Yamaha
			Aero Robot "R-50.
			& 
			Monitoring Agriculture, spraying UAV systems.
			& 
			20 kg / LAP / 1 hour.& Azimuth and Differential Global Positioning System (DGPS) sensor system.
			&\cite{ muchiri2016review }
			\\
			\hline
			Yanmar KG-135, YH300 and AYH3.
			& 
			Pesticide spraying over crop fields.
			& 
			22.7 kg / 1500m / 5 hours.& Spray system with GPS sensor system.
			&\cite{huang2013development}
			\\
			\hline
			RC model fixed-wing airframe.
			& 
			Imaging small sorghum fields to assess the attributes of a grain crop.
			& 
			Less than 1kg / LAP / less than 1 hour.& Image sensor digital camera.
			&\cite{ jensen2003assessing, huang2013development}
			\\
			\hline
			Vector-P UAV.
			& 
			Crop management (e.g. winter wheat) for site-specific agriculture, a correlation is investigated between leaf area index and the green normalized difference vegetation index (GNDVI).
			& 
			Less than 1kg /105m-210m/1-6 hours deepening on the payload.& Digital color-infrared camera with a red-light-blocking filter.
			&\cite{ hunt2010acquisition }
			\\
			\hline
			Fixed-wing UAV.
			& 
			Detect the variability in crop response to irrigation  (e.g. cotton).
			& 
			Lightweight camera/ 90m / Less than 1 hour.& Thermal camera , Thermal Infrared (TIR) imaging sensor.
			&\cite{ sullivan2007evaluating }
			\\
			\hline
			Multi-rotor micro UAV.
			& 
			Agricultural management, 
			disease detection for citrus (citrus greening, Huanglongbing (HLB)).
			& 
			Less than 1kg / 100 m / 10-20 min.& Multi-band imaging sensor, 6-channel multispectral camera.
			&\cite{ garcia2013comparison }
			\\
			\hline
			Vario XLC helicopter.
			& 
			Weed management, reduce  the amount  of  herbicides using aerial images for crop.
			& 
			7 kg / LAP / 30 min.& Advanced  vision  sensors  for  3D  and  multispectral  imaging.
			&\cite{ kazmi2011adaptive }
			\\
			\hline
			VIPtero UAV.
			& 
			Crop management, they used UAV to acquire high resolution multi-spectral images for vineyard management.
			& 
			1 Kg / 150 m / 10 min.& Tetracam ADC-lite camera, GPS.
			&\cite{primicerio2012flexible}
			\\
			\hline
			Fieldcopter UAV.
			& 
			Water assessment. UAVs was used for acquiring high resolution  images and  to assess  vineyard water  status, which can help for irrigation processes.
			& 
			Less than 1 Kg / LAP /NA.& Multispectral and thermal cameras on-board UAV.
			&\cite{ baluja2012assessment}
			\\
			\hline
			Multi-rotor hexacopter ESAFLY A2500-WH
			& 
			Cultivations analysis, processing multi spectral data of the surveyed sites to create tri-band ortho-images  used to extract some Vegetation Indices (VI). 
			
			& 
			Up to 2.5 kg Kg / LAP /12-20 min.& Tetracam camera on-board UAV.
			&\cite{ candiago2015evaluating}
			\\
			\hline
		\end{tabu}
		
	\end{adjustbox}
	
	\label{table:ahmad1-4}
\end{table*}
\subsection{The Deployment of UAV in Precision Agriculture}
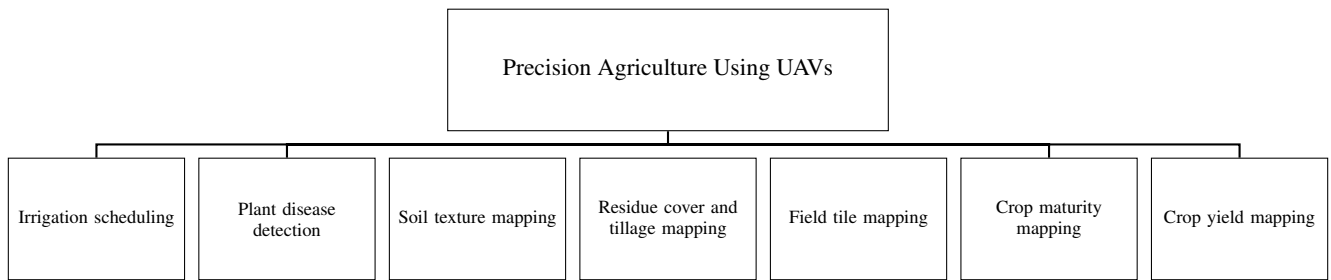
\begin{figure*}[h] 
	\centering
	\begin{tikzpicture}[font=\scriptsize]
	\tikzset{every node/.style=
		{align=center, minimum height=46pt, text width=60pt}}
	\node[,draw=black] (b1) {Irrigation scheduling
		
	};
	\node[right=5pt,draw=black] (b2) at (b1.east) {Plant disease detection
	};
	\node[right=5pt,draw=black] (b3) at (b2.east) {Soil texture mapping
		
	};  
	\node[right=5pt,draw=black] (b4) at (b3.east) {Residue cover and tillage mapping
	};
	\node[right=5pt,draw=black] (b5) at (b4.east) {Field tile mapping
	};
	\node[right=5pt,draw=black] (b6) at (b5.east) {Crop maturity mapping
	};
	\node[right=5pt,draw=black] (b7) at (b6.east) {Crop yield mapping
	};
	\node[above=10pt, text width=160pt,draw=black] (top) at ($(b2.north)!.5!(b6.north)$) {\small{Precision Agriculture Using UAVs
	}};
	\coordinate (atop) at ($(top.south) + (0,-5pt)$);
	\coordinate (btop) at ($(b5.south) + (0,-5pt)$);
	\draw[thick] (top.south) -- (atop)
	(b1.north) |- (atop) -| (b7.north)
	(b2.north) |- (atop) -| (b6.north);
	
	
	
	
	\end{tikzpicture}
	\caption{The Deployment of UAVs in Precision Agriculture Applications.
	}
	\label{pa1}
\end{figure*}

In~\cite{khanal2017overview}, the authors presented the deployment of UAVs in precision agriculture applications as summarized in Figure~\ref{pa1}. The deployment of UAV in precision agriculture are discussed in the following:
\begin{itemize}
	\item {Irrigation scheduling:} There are four factors that needs to be monitored, in order to determine a need for irrigation: 1) Availability of soil water; 2) Crop water need, which represents the amount of water needed by the various crops to grow optimally; 3) Rainfall amount; 4) Efficiency of the irrigation system~\cite{natinoal}. These factors can be quantified by utilizing UAVs to measure soil moisture, plant-based temperature, and evapotranspiration. For instance, the spatial distribution of surface soil moisture can be estimated using high-resolution multi-spectral imagery captured by a UAV, in combination with ground sampling~\cite{hassan2015assessment}. The crop water stress index can also be estimated, in order to determine water stressed areas by utilizing thermal UAV images~\cite{gonzalez2013using}.
	\item {Plant disease detection:} In the U.S., it is estimated that crop losses caused by plant diseases result in about \$33 billion in lost revenue every year~\cite{pimentel2005update}. UAVs can be used for thermal remote sensing to monitor the spatial and temporal patterns of crop diseases pre-symptomatically during various disease development phases and hence farmers may reduce the crop losses. For instance, aerial thermal images can be used to detect early stage development of soil-borne fungus~\cite{calderon2013high}.
	\item {Soil texture mapping:} Some soil properties, such as soil texture, can be an indicative of soil quality which in turn influences crop productivity. Thus, UAV thermal images can be utilized to quantify soil texture at a regional scale by measuring the differences in land surface temperature under a relatively homogeneous climatic condition~\cite{de2012mapping,wang2015retrieval}.
	\item {Residue cover and tillage mapping:} Crop residues is essential in soil conservation by providing a protective layer on agricultural fields that shields soil from wind and water. Accurate assessment of crop residue is necessary for proper implementation of conservation tillage practices~\cite{Department}. In~\cite{sullivan2004evaluation}, the authors demonstrated that aerial thermal images can explain more than 95\% of the variability in crop residue cover amount compared to 77\% using visible and near IR images.
	\item {Field tile mapping:} Tile drainage systems remove excess water from the fields and hence it provides ecological and economic benefits~\cite{hofstrand2015economics}. An efficient monitoring of tile drains can help farmers and natural resource managers to better mitigate any adverse environmental and economic impacts. By measuring temperature differences within a field, thermal UAV images can provide additional opportunities in field tile mapping~\cite{mapping}.
	\item {Crop maturity mapping:} UAVs can be a practical technology to monitor crop maturity for determining the harvesting time, particularly when the entire area cannot be harvested in the time available. For instance, UAV visual and infrared images from barley trial areas at Lundavra, Australia were used to map two primary growth stages of barley and demonstrated classification accuracy of 83.5\%~\cite{jensen2009crop}.
	\item {Crop yield mapping:}
	Farmers require accurate, early estimation of crop yield  for a number of reasons, including crop insurance, planning of harvest and storage requirements, and cash flow budgeting. In~\cite{swain2010adoption}, UAV images were utilized to estimate yield and total biomass of rice crop in Thailand. In~\cite{geipel2014combined}, UAV images were also utilized to predict corn grain yields in the early to mid-season crop growth stages in Germany.
\end{itemize}

The authors in~\cite{sankaran2015low} presented several types of sensor that were used in UAV-based precision agriculture, as summarized in Table~\ref{pa2}.

\subsection{Challenges}
There are several challenges in the deployment of UAVs in PA:
\begin{itemize}
	\item Thermal cameras have poor resolution and they are expensive. The price ranges from \$2000-\$50,000 depending on the quality and functionality, and the majority of thermal cameras have resolution of 640 pixels by 480 pixels~\cite{khanal2017overview}.
	\item Thermal aerial images can be affected by many factors, such as the moisture in the atmosphere, shooting distance, and other sources of emitted and reflected thermal radiation. Therefore, calibration of aerial sensors is critical to extract scientifically reliable surface temperatures of objects~\cite{khanal2017overview}.
	\item Temperature readings through aerial sensors can be affected by crop growth stages. At the beginning of the growing season, when plants are small and sparse, temperature measurements can be influenced by reflectance from the soil surface~\cite{khanal2017overview}.
	\item In the event of adverse weather, such as extreme wind, rain and storms, there is a big challenge of UAVs deployment in PA applications. In these conditions, UAVs may fail in their missions. Therefore, small UAVs cannot operate in extreme weather conditions and even cannot take readings during these conditions.
	
	\item One of the key challenges is the ability of lightweight  UAVs to carry a high-weight payload, which will limit the ability of UAVs to carry an integrated system that includes multiple sensors, high-resolution and thermal cameras\cite{anderson2013lightweight}.
	
	\item UAVs have short battery life time, usually less than 1 hour. Therefore, the power limitations of UAVs is one of the challenges of using UAVs in PA. Another challenge, when UAVs are used to cover large areas, is that it needs to return many times to the charging station for recharging. \cite{ garcia2013comparison, jensen2003assessing, huang2013development }.

\end{itemize}

\subsection{Research Trends and Future Insights}
\subsubsection{Machine Learning }
The next generation of UAVs will utilize the new technologies in precision agriculture, such as machine learning. Hummingbird is a UAV-enabled data and imagery analytics business for precision agriculture~\cite{hummingbirdtech}. It utilizes machine learning to deliver actionable insights on crop health directly to the field. The process flow begins by performing UAV surveys on the agricultural land at critical decision-making points in the growing season. Then, UAV images is uploaded to the cloud, before being processed with machine learning techniques. Finally, the mobile app and web based platform provides farmers with actionable insights on crop health. The advantages of utilizing UAVs with machine learning technology in precision agriculture are: 1) Early detection of crop diseases; 2) Precision weed mapping; 3) Accurate yield forecasting; 4) Nutrient optimization and planting; 5) Plant growth monitoring~\cite{hummingbirdtech}.
\begin{table*}[!h]
	\scriptsize
	\renewcommand{\arraystretch}{1}
	\caption{\uppercase{UAV sensors in precision agriculture applications}}
	\label{pa2}
	\centering
	\begin{tabular}{|c|c|c|c|}
		\hline 
		Type of sensor&Operating Frequency&Applications& Disadvantages\\ 
		\hline
		Digital camera&Visible region&Visible properties, outer defects, greenness, growth.&- Limited to visual spectral bands and properties. ~~~~~~~~\\
		\hline
		Multispectral camera&Visible-infrared region&Multiple plant responses to nutrient deficiency, &- Limited to few spectral bands.~~~~~~~~~~~~~~~~~~~~~~~~~~~~~\\
		&&water stress, diseases among others.&\\
		\hline
		Hyperspectral camera&Visible-infrared region&Plant stress, produce quality, and safety control.&- Image processing is challenging.~~~~~~~~~~~~~~~~~~~~~~~~~~ \\
		&&&- High cost sensors.~~~~~~~~~~~~~~~~~~~~~~~~~~~~~~~~~~~~~~~~~~~~\\
		\hline
		Thermal camera&Thermal infrared region&Stomatal conductance,
		plant responses to&- Environmental conditions affect the performance.~~~~~~~\\
		&&water stress and diseases.&- Very small temperature
		differences are not
		detectable.~
		\\
		&&&- High resolution cameras are heavier.~~~~~~~~~~~~~~~~~~~~~~\\
		\hline
		Spectrometer&Visible-near infrared region&Detecting disease, stress and crop responses.&- Background such as soil
		may affect the data quality.~~
		\\
		&&&- Possibilities of spectral
		mixing.~~~~~~~~~~~~~~~~~~~~~~~~~~~
		\\
		&&&- More applicable for
		Ground sensor  systems.~~~~~~~~~~~~~\\
		\hline
		3D camera&Infrared laser region&Physical attributes such as plant height&- Lower accuracies.~~~~~~~~~~~~~~~~~~~~~~~~~~~~~~~~~~~~~~~~~~~~\\
		&&and canopy density.&- Limited field applications.~~~~~~~~~~~~~~~~~~~~~~~~~~~~~~~~~~\\
		\hline
		LiDAR&Laser region&Accurate estimates of plant/tree height &- Sensitive to small variations in path length.~~~~~~~~~~~~~~\\
		&& and volume.&\\
		\hline
		SONAR&Sound propagation&Mapping and quantification of the canopy volumes,&- Sensitivity limited by acoustic absorption, background \\
		&&digital control of application rates in sprayers or&noise, etc.~~~~~~~~~~~~~~~~~~~~~~~~~~~~~~~~~~~~~~~~~~~~~~~~~~~\\
		&&fertilizer spreader.&- Lower sampling rate than laser-based sensing.~~~~~~~~~~\\
		\hline
	\end{tabular}
\end{table*}
\subsubsection{Image Processing}
UAV-based systems can be used in PA to acquire high-resolution images for farms, crops and rangeland. It can also be utilized as an alternative to satellite and manned aircraft imaging system. Processing of these images is one of the most rapidly developing fields in PA applications.
The Vegetation Indices (VI) can be produced using image processing techniques for the prediction of the agricultural crop yield, agricultural analysis, crop and weed management and in diseases detection. Moreover, the VIs can be used to create vigor maps of the specific-site and for vegetative covers evaluation using spectral measurements \cite{candiago2015evaluating,bannari1995review}.

In \cite{bannari1995review,glenn2008relationship,candiago2015evaluating,hunt2010acquisition}, most of the VIs found in the literature have been summarized and discussed. Some of these VIs are:
\begin{itemize}
	\item Green Vegetation Index (GVI).
	\item Normalized Difference Vegetation Index (NDVI).
	\item Green Normalized Difference Vegetation index (GNDVI).
	\item Soil Adjusted Vegetation Index (SAVI).
	\item Perpendicular Vegetation Index (PVI).
	\item Enhanced Vegetation Index (EVI).
	
\end{itemize}

Many researchers utilized VIs that are obtained using image processing techniques in PA.
The authors in \cite{ candiago2015evaluating} presented agricultural analysis for vineyards and tomatoes crops. A UAV with Tetracam multi-spectral camera was deployed to take aerial image for crop. These images were processed using PixelWrench2 (PW2) software which came with the camera and it will be exported in a tri-band TIFF image. Then from the contents of this images VIs such as NDVI \cite{ rouse1974monitoring}, GNDVI \cite{ gitelson1996use}, SAVI \cite{huete1988soil} can be extracted.

In~\cite{laliberte2010acquisition}, the authors used UAVs to take aerial images for rangeland to identify rangeland VI for different types of plant in Southwestern Idaho. In the study, image processing and analysis was performed in three steps as shown in Figure~\ref{PA-image}. More specifically, the three steps were:
\begin{figure}
	\centering
	\includegraphics[scale=0.3]{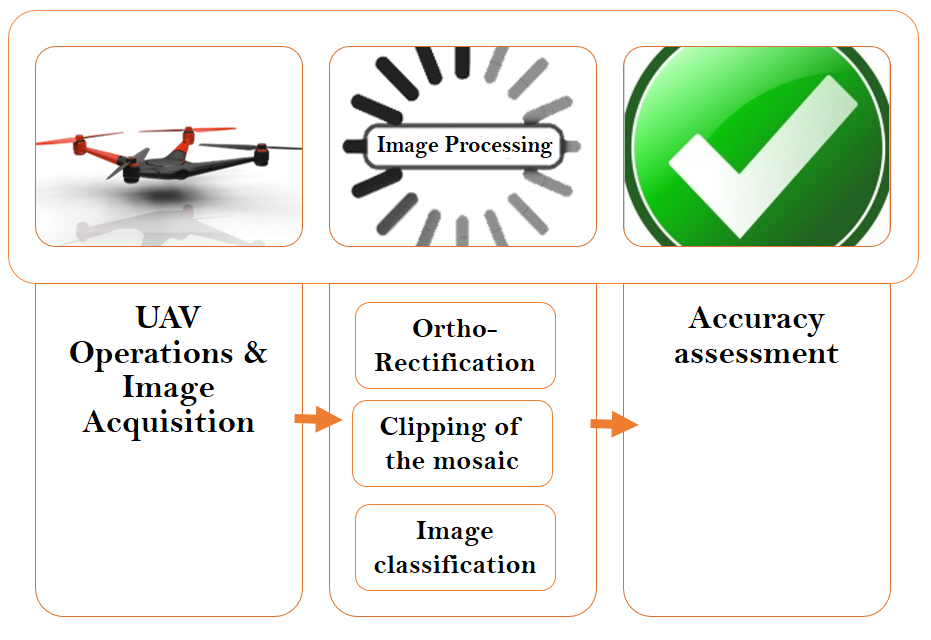}
	\caption{Steps of Image Processing and Analysis  for Identifying Rangeland VI \cite{laliberte2010acquisition}.}
	\label{PA-image}
\end{figure}
\begin{itemize}
	\item Ortho-Rectification and mosaicing of UAV imagery \cite{laliberte2010acquisition}. A semi-automated ortho-rectification approach were developed using PreSync procedure \cite{laliberte2008procedure}.
	
	\item Clipping of the mosaic to the 50$m$$\times$50$m$ plot areas measured on the ground. In this step, image classification and segmentation was performed using an object-based image analysis (OBIA) program with Definiens Developer 7.0 \cite{definiens2007definiens}, where the acquisition image was segmented into homogeneous areas \cite{laliberte2010acquisition}.
	
	\item Image classification: In this step, hierarchical classification scheme along with a rule based masking approach were used \cite{laliberte2010acquisition}. 
	
\end{itemize}

\subsubsection{Future Insights}
Based on the reviewed articles focusing on PA using UAVs, we suggest these future possible directions:
\begin{itemize}
	\item With relaxed flight regulations and improvement in image processing, geo-referencing, mosaicing, and classification algorithms, UAV can provide a great potential for soil and crop monitoring~\cite{khanal2017overview,zhang2012application}.
	\item The next generation of UAV sensors, such as 3p sensor~\cite{SLANTRANGE}, can provide on-board image processing and in-field analytic capabilities, which can give farmers instant insights in the field, without the need for cellular connectivity and cloud connection~\cite{next}.
	\item More precision agricultural researches are required towards designing and implementing special types of cameras and sensors on- board UAVs, which have the ability of remote crop monitoring and detection of soil and other agricultural characteristics in real time scenarios \cite{primicerio2012flexible}.
	\item UAVs can be used for obtaining high-resolution images for plants to study plant diseases and traits using image processing techniques \cite{patil2011advances}.
	
\end{itemize}

\section{Delivery of Goods}
\begin{figure*}[h]
	\centering
	\includegraphics[scale=0.57]{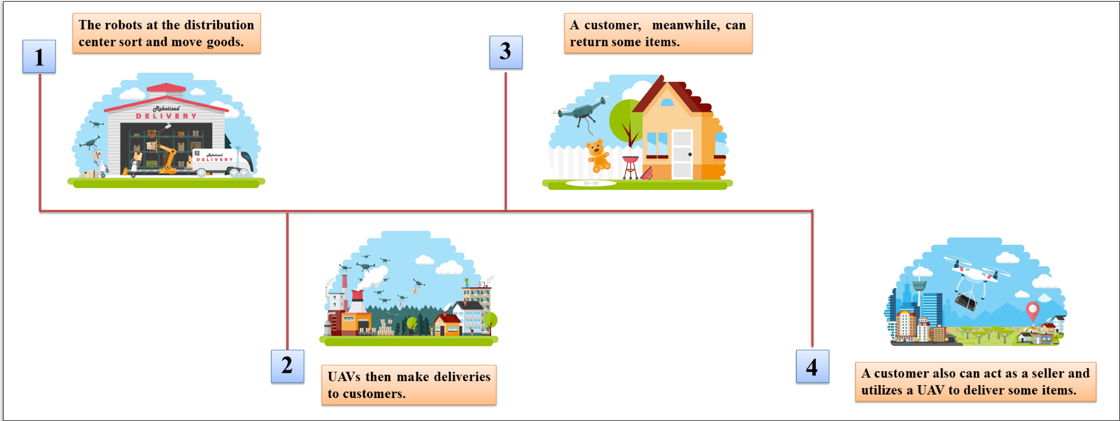}
	\caption{Autonomous E-Commerce.}
	\label{dg1}
\end{figure*}

\begin{figure}
	\centering
	\includegraphics[scale=0.35]{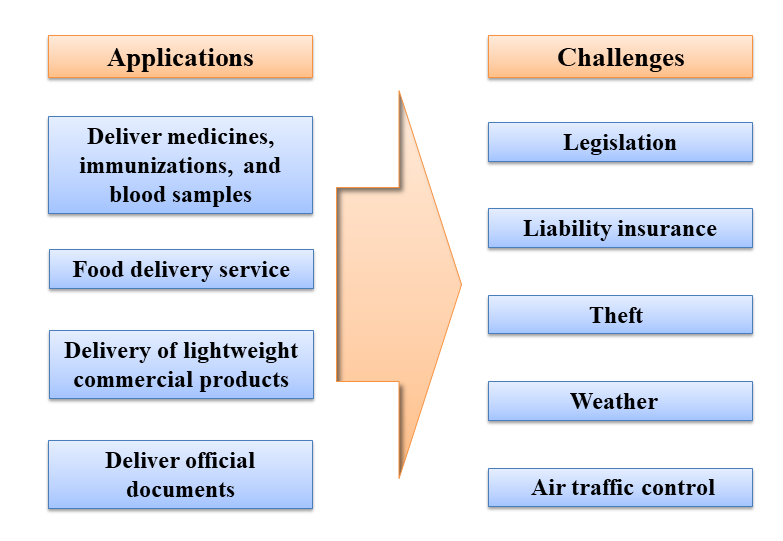}
	\caption{UAV Delivery Applications and Challenges.}
	\label{dg2}
\end{figure}

UAVs can be used to transport food, packages and other goods ~\cite{fandetti2015method,pwc,redstagfulfillment,tworedstagfulfillment}
as shown in Figure~\ref{dg1}. In health-care field, ambulance drones can deliver medicines, immunizations, and blood samples, into and out of unreachable places. They can rapidly transport medical instruments in the crucial few minutes after cardiac arrests. They can also include live video streaming services allowing paramedics to remotely observe and instruct on-scene individuals on how to use the medical instruments~\cite{wikipedia}. In July 2015, the Federal Aviation Administration (FAA) approved the first delivery of medical supplies using UAVs at Wise, Virginia~\cite{howell2016first}. With the rapid demise of snail mail and the massive growth of e-Commerce, postal companies have been forced to find new methods to expand beyond their traditional mail delivery business models. Different postal companies have undertaken various UAV trials to test the feasibility and profitability of UAV delivery services~\cite{unmannedcargo}. In this section, we present the UAV-based goods delivery system and its challenges as shown in Figure~\ref{dg2}.

\begin{figure}
	\centering
	\includegraphics[scale=0.35]{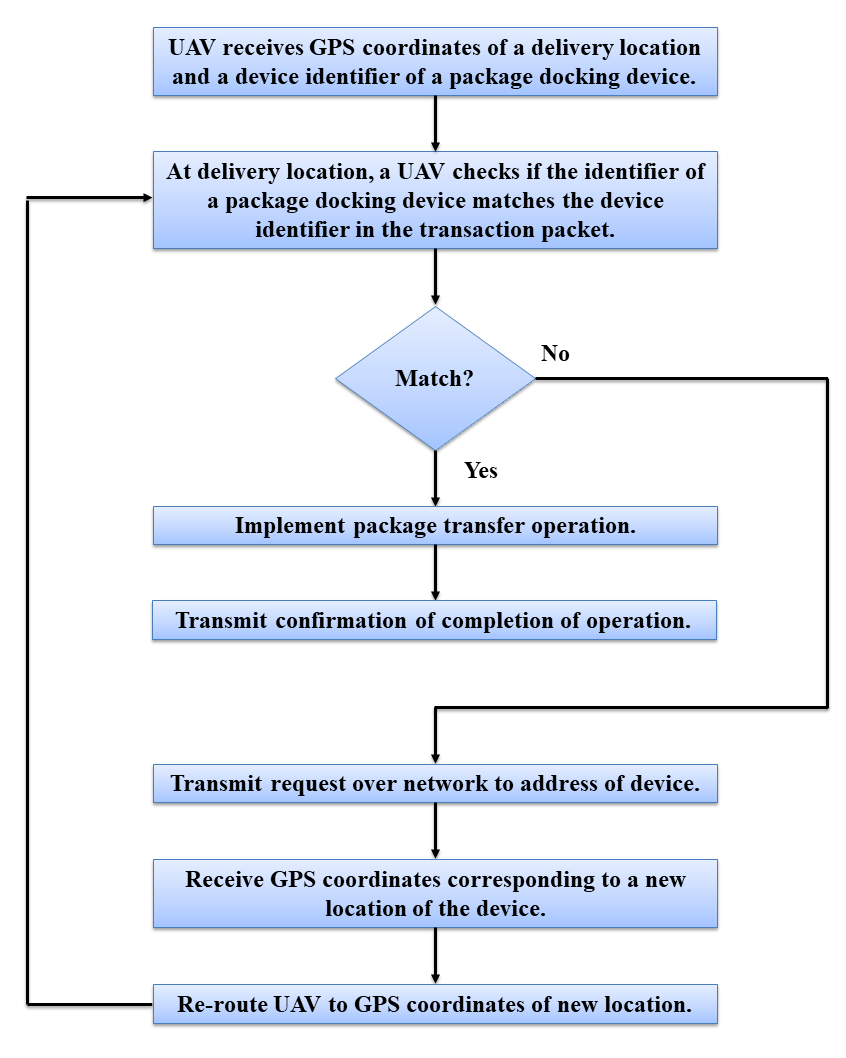}
	\caption{UAV Delivery of Goods System.}
	\label{dg3}
\end{figure}
\subsection{UAV-Based Goods Delivery System}
In UAV-based goods delivery system, a UAV is capable of traveling between a pick up location and a delivery location. The UAV is equipped with control processor and GPS module. It receives a transaction packet for the delivery operation that contains the GPS coordinates and the identifier of a package docking device associated with the order. Upon arrival of a UAV at the delivery location, the control processor checks if the identifier of a package docking device matches the device identifier in the transaction packet, performs the package transfer operation, and sends confirmation of completion of the operation to an originator of the order~\cite{hoareau2017package}. If the identifier of a package docking device at the delivery point does not match the device identifier in the transaction packet, the UAV communication components transmit a request over a short-range network such as bluetooth or Wi-Fi. The request may contain the device identifier, or network address of the package docking device. Under the assumption that the package docking device has not moved outside of the range of UAV communication, the package docking device having the network address transmits a signal containing the address of a new location. The package docking device may then transmit updated GPS coordinates to the UAV. The UAV is re-routed to the new address based on the updated GPS location~\cite{hoareau2017package}. In Figure~\ref{dg3}, we present the flowchart of UAV delivery system.

\subsection{Challenges}
\subsubsection{Legislation}
In the United States, the FAA regulation blocked all attempts at commercial use of UAVs, such as the Tacocopter company for food delivery~\cite{gilbert2012tacocopter}. As of 2015, delivering of packages with UAVs in the United States is not allowed~\cite{mashable}. Under current rules, companies are permitted to operate commercial UAVs in the United States, but only under certain conditions. Their UAVs must be flown within a pilot’s line of sight and those pilots must get licenses. Commercial operators also are restricted to fly their UAVs during daylight hours. Their UAVs are limited in size, altitude and speed, and UAVs are generally not allowed to fly over people or to operate beyond visual line of sight~\cite{recode}.
\subsubsection{Liability Insurance}
Some UAVs can weigh up to 25 kg and travel at speeds approaching 45 m/s. A number of media reports describes severe lacerations, eye loss, and soft tissue injuries caused by UAV accidents. In addition to the risk of injuries or property damage from a UAV crash, ubiquitous UAV uses also create other types of accidents, such as automobile accidents due to distraction from low-flying UAVs, injuries caused by dropped cargo, liability for damaged goods, or accidents resulting from a UAV's interference with aircraft. Liability for UAV use, however, is not limited to personal injury or property damage claims, UAVs present an enormous threat to individual privacy~\cite{law}.
\begin{figure*}[!t]
	\centering
	\includegraphics[scale=0.55]{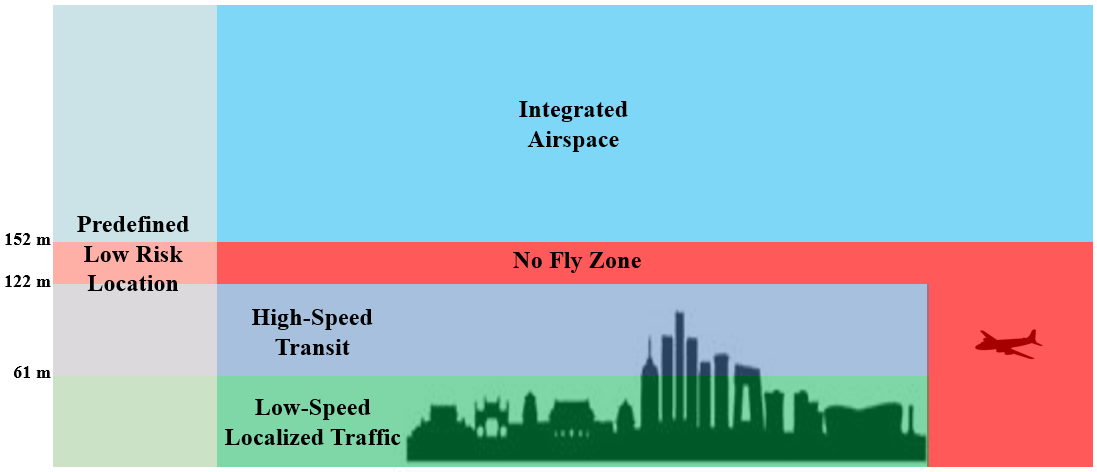}
	\caption{Amazon Airspace Model for the Safe Integration of UAV Systems.}
	\label{dg4}
\end{figure*}
\subsubsection{Theft}
The main concerns of utilizing UAVs for data gathering and wireless delivery, are cyber liability and hacking. UAVs that are used to gather sensitive information might become targets for malicious software seeking to steal data. A hacker might even usurp control of the UAV itself for the purpose of illegal activities, such as theft of its cargo or stored data, invasion of privacy, or smuggling. Liability for utilizing a UAV does not fit neatly into the coverage offered by the types of liability insurance policies that most individuals and businesses currently possess~\cite{law}.
\subsubsection{Weather}
Similar to light aircrafts, UAVs cannot hover in all weather conditions. The capability to resist certain weather conditions is determined by the specifications of the UAV~\cite{fornace2014mapping}. In pre-flight planning, it is clear that advanced weather data will play an essential role in ensuring that UAVs can fly their weather-sensitive missions safely and efficiently to deliver commercial goods. During flight operations, weather data affects flight direction, path elevation, operation duration and other in-flight variables. Wind speeds in particular are an essential component for a smooth UAV-based operations and thus should be factored in the operation planning and deployment phases. In post-flight analysis, by analyzing data through advanced weather visualization dashboards, we can improve the UAV flight operations to ensure future mission success~\cite{unmanned}.
\subsubsection{Air Traffic Control}
Air traffic control is an essential condition for coordinating large fleets of UAVs, where regulators will not permit large-scale UAV delivery missions without such systems in place~\cite{businessinsider}. Amazon designs an airspace model for the safe integration of UAV systems as shown in Figure~\ref{dg4}. In this proposed model, the “low-speed localized traffic” will be reserved for: 1) Terminal non-transit operations such as surveying, videography and inspection; 2) Operations for lesser-equipped UAVs, e.g. ones without sophisticated sense-and-avoid technology. The “high-speed transit” will be reserved for well equipped UAVs as determined by the relevant performance standards and rules. The “no fly zone” will serve as a restricted area in which UAV operators will not be allowed to fly, except in emergencies. Finally, the “predefined low risk locations” will include areas like designated academy of model aeronautics airfields, altitude and equipage restrictions in these locations will be established in advance by aviation authorities~\cite{air2015revising}.
\subsection{Research Trends and Future Insights}
\subsubsection{{Machine Learning}}
With machine learning, UAVs can fly autonomously without knowing the objects they may encounter which is important for large-scale UAV delivery missions. Qualcomm’s tech shows more advanced computing that can actually understand what the UAV encountered in mid-air and creates a flight route. They show how the UAV processing and decision-making technology is nimble enough to allow UAVs to operate in unpredictable settings without using any GPS. All of the UAV computational tasks, like the machine learning and flight control, happens on 12 grams processor without any off-board computing~\cite{recodejaz}. Some of challenges are : 1) We need to put a lot of effort to find efficient methods to do unsupervised learning, where collecting large amounts of unlabeled data is nowadays becoming economically less expensive~\cite{carrio2017review}; 2) Real-world problems with high number of states can turn the problem intractable with current techniques, severely limiting the development of real applications. An efficient method for coping with these types of problems remains as an unsolved challenge~\cite{carrio2017review}.
\subsubsection{{Navigation System}}
Researchers are developing navigation systems that do not utilize GPS signals. This could enable UAVs to fly autonomously over places where GPS signals are unavailable or unreliable. Whether delivering goods to remote places or handling emergency tasks in hazardous conditions, this type of capability could significantly expand UAVs' usefulness~\cite{theconversation}. Researchers from GPU maker Nvidia are currently working on a navigation system that utilizes visual recognition and computer learning to make sure UAVs don't get lost. The team believes that the system has already managed the most stable GPS-free flight to date~\cite{newatlas}. Some of challenges are: 1) The design of computing devices with low-power consumption, particularly GPUs, is a challenge and active working field for embedded hardware developers~\cite{carrio2017review}; 2) While a few UAVs can already travel without a UAV operators directing their routes, this technology is still emerging. Over the next few years, system-failure responses, adaptive routing, and handoffs between user and UAV controllers should be improved~\cite{mckinsey}. 

\subsubsection{Future Insights}
Some of the future possible directions for this application are:
\begin{itemize}
	\item The energy density of lithium-ion batteries is improving by 5\%-8\% per year, and their lifetime is expected to double by 2025. These improvements will make commercial UAVs able to hover for more than an hour without recharging, enabling UAVs to deliver more goods~\cite{mckinsey}.
	\item Detect-and-avoid systems which help UAVs to avoid collisions and obstacles, are still in development, with strong solutions expected to emerge by 2025~\cite{mckinsey}.
	\item UAVs currently travel below the height of commercial aircraft due to the collision potential. The methods that can track UAVs and communicate with air-traffic-control systems for typical aircraft are not expected to be available before 2027, making high-altitude missions impossible until that time~\cite{mckinsey}.
	\item To make UAV delivery practical, automation research is required to address UAVs design. UAVs design covers creating aerial vehicles that are practical, can be used in a wide range of conditions, and whose capability rivals that of commercial airliners; this is a significant undertaking that will need many experiments, ingenuity and contributions from experts in diverse areas~\cite{d2014guest}.
	\item More research is needed to address localization and navigation. The localization and navigation problems may seem like simple problems due to the many GPS systems that already exist, but to make drone delivery practical in different operating conditions, the integration of low cost sensors and localization systems is required~\cite{d2014guest}.
	\item More research is needed to address UAVs coordination. Thousands of UAV operators in the air, utilizing the same resources such as charging stations and operating frequency, will need robust coordination which can be studied by simulation~\cite{d2014guest}. 
\end{itemize}

\section{Real-Time Monitoring of Road Traffic}

Automation of the overall transportation system cannot be automated through vehicles only\cite{menouar2017uav}. In fact, other components of the end-to-end transportation system, such as tasks of field support teams, traffic police, road surveyors, and rescue teams, also need to be automated. Smart and reliable UAVs can help in the automation of these components.

UAVs have been considered as a novel traffic monitoring technology to collect information about traffic conditions on roads. Compared to the traditional monitoring devices such as loop detectors, surveillance video cameras and microwave sensors, UAVs are cost-effective, and can monitor large continuous road segments or focus on a specific road segment \cite{ke2017real}. Data generated by sensor technologies are somewhat aggregated in nature and hence do not support an effective record of individual vehicle tracks in the traffic stream. This restricts the application of these data in individual driving behavior analysis as well as calibrating and validating simulation models \cite{GUIDO2016136}. 

Moreover, disasters may damage computing, communications infrastructure or power systems. Such failures can result in a complete lack of the ability to control and collect data about the transportation network \cite{Leitloff_2014}.

\subsection{Literature Review} \label{related-work}
UAVs are getting accepted as a method to hasten the gathering of geographic surveillance data \cite{smartcities}.
As autonomous and connected vehicles become popular, many new services and applications of UAVs will be enabled \cite{menouar2017uav}.

Recognition of moving vehicles using UAVs is still a challenging problem.
Moving vehicle detection methods depend on the accuracy of image registration methods, since the background in the UAV surveillance platform changes frequently. Accurate image registration methods require extensive computing power, which affecta the real-time capability of these methods \cite{qu2016moving}. 
In \cite{qu2016moving}, Qu et al. studied the problem of moving vehicle detection using UAV cameras. In their proposed approach, they used convolutional neural networks to identify vehicles more accurately and in real-time. The proposed approach consists of three steps to detect moving vehicles: First, adjacent frames are matched. Then, frame pixels are classified as background or candidate targets. Finally, a deep convolutional neural network is trained over candidate targets to classify them into vehicles or background. They achieved detection accuracy of around $90\%$ when evaluating their method using the CATEC UAV dataset.

In \cite{WANG2016294}, the authors introduce a vehicle detection and tracking system based on imagery data collected by a UAV. This system uses consecutive frames to generate the vehicle's dynamic information, such as positions and velocities over time. Four major modules have been developed in this study: image registration, image feature extraction, vehicle shape detecting, and vehicle tracking. Some unique features have been introduced into this system to customize the vehicle and traffic flow and use them together in multiple contiguous images to increase the system's accuracy of detecting and tracking vehicles.

A framework is presented in \cite{ke2017real} to support real-time and accurate collection of traffic flow parameters, including speed, density, and volume, in two travel directions simultaneously. The proposed framework consists of the following four features: (1) A framework for estimating multi-directional traffic flow parameters from aerial videos (2) A method combining the Kanade–Lucas–Tomasi (KLT) tracker, k-means clustering, and connected graphs for vehicle detection and counting (3) Identifying traffic streams and extracting traffic information in a real-time manner (4) The system works in daytime and nighttime settings, and is not sensitive to UAV movements (i.e., regular movement, vibration, drifting, changes in speed, and hovering). A challenge that this framework faces is that their algorithm sometimes recognizes trucks, buses, and other large/heavy vehicles as multiple passenger cars.

A real-time framework for the detection and tracking of a specific road segment using low and mid-altitude UAV video feeds was presented in \cite{zhou2015efficient}. This framework can be used for autonomous navigation, inspection, traffic surveillance and monitoring. For road detection, they utilize the GraphCut algorithm abecause of its efficient and powerful segmentation performance in 2-D color images. For road tracking, they develop a tracking technique based on homography alignment to adjust one image plane to another when the moving camera takes images of a planar scene. 

In \cite{Leitloff_2014}, the authors develop a processing procedure for fast vehicle detection, which consists of three stages; pre-classification with a boosted classifier, blob detection, and final classification using SVM. 

In \cite{puri2007statistical}, the authors propose to integrate collected video data from UAVs with traffic simulation models to enhance real-time traffic monitoring and control. This can be performed by transforming collected video data into ‘useful traffic measures’ to generate essential statistical profiles of traffic patterns, including traffic parameters such as mean-speed, density, volume, turning ratio, etc. However, a main issue with this approach is the limitation of flying time for UAVs which could hover to obtain data for a few hours a day.

The work in \cite{reshma2016security} addresses the security issues of road traffic monitoring systems using UAVs. In this work, the role of a UAV in a road traffic management system is analyzed and various situational security issues that occur in traffic management are mitigated. In their proposed approach, the UAV's intelligent systems analyze real-time traffic as well as security issues and provide the appropriate mitigation commands to the traffic management control center for re-routing. Instead of image processing, the authors used sensor networks and graph theory for representing the road network. They also devised different situational security scenarios to assess the road traffic management.For example, a car without an RFID tag that enters a defense area or government building area is considered a potential security risk.

Based on a research study by Kansas Department of Transportation (KDOT) \cite{Kansas}, the use of UAVs for KDOT's operations could lead to improved safety, efficiency, as well as reduced costs. The study also recommends the use of UAVs in a range of applications including bridge inspection, radio tower inspection, surveying, road mapping, high-mast light tower inspection, stockpile measurement, and aerial photography. However, the study indicates that UAVs are not recommended to replace the current methods of traffic data collection in KDOT operations which rely on different kinds of sensors (e.g., weight, loop, piezo). However, UAVs can complement existing data collection projects to gather data in small increments of time in certain traffic areas. Based on their survey, battery life and flight time of UAVs may limit the samples of traffic data. For example, current UAV technology cannot collect 24-hour continuous data.

Yu Ming Chen et al. \cite{chen2007real} proposed a video relay scheme for traffic surveillance systems using UAVs. The proposed communications scheme is straightforward to implement because of the typical availability of mobile broadband along highways. Their results show that the proposed scheme can transfer quality videos to the traffic management center in real-time. They also implemented two types of data communications schemes to transmit captured videos through existing public mobile broadband networks: (1) Video stream delivered directly to clients. (2) Video stream delivered to clients through a server. In their experiments, they were able to transmit video signals with an image size of $320\times280$, at a rate of $112$ Kbps, and $15$ frames per second.

In \cite{ro2007lessons}, a platform which operates autonomously and delivers high-quality video imagery and sensor data in real-time is utilized. In their scenario, the authors employ a $10$ lbs aircraft to fly up to $6$ hours with a telemetry range of $1$ mile and payload capacity of $4$ lbs. Their system consists of five components: (1) a GPS signal receiver (2) a radio control transmitter (3) a modem for flight data (4) a PC to display the UAV on a map. (5) a real-time video down-link.

Apeltauer et al. \cite{apeltauer2015automatic} present an approach for moving vehicle detection and tracking through the intersection of aerial images captured by UAVs. Overall, the system follows three steps: pre-processing, vehicle detection, and tracking. For pre-processing, images are undistorted geo-registered against a user-selected reference frame. For the detection step, the boosting technique is used to improve the training phase which employs Multi-scale Block Local Binary Patterns (MB-LBP). Finally for tracking, the system uses a set of Bootstrap particle filters, one per vehicle.

\begin{figure*}[!t]
	\centering
	\includegraphics[width=1.0\textwidth]{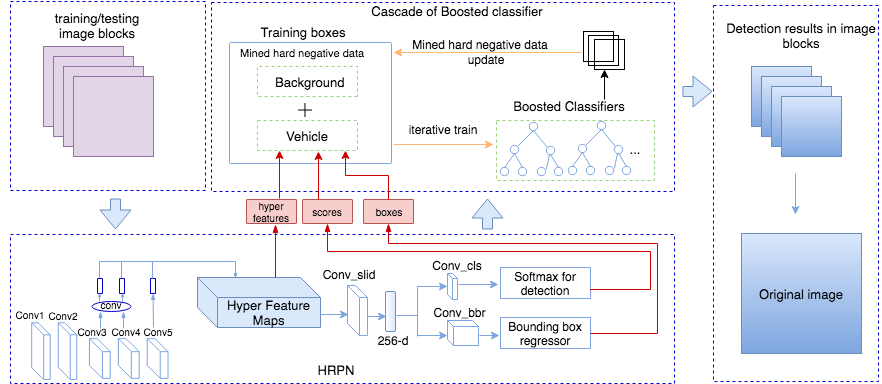}
	\caption{\label{fig:HRPN} Proposed Vehicle Detection Framework in \cite{tang2017vehicle}.}
\end{figure*}

An improved vehicle detection method based on Faster R-CNNs is proposed in \cite{tang2017vehicle}. The overall vehicle detection method is illustrated in Figure \ref{fig:HRPN}. For training, the method crops the original large-scale images into segments and augments the number of image segments with four angles (i.e., {0}, {90}, {180}, and {270}). Then, all the training image blocks that constitute the HRPN input are processed to produce candidate region boxes, scores and corresponding hyper features. 
Finally, the results of the HRPN are used to train a cascade of boosted classifiers, and a final classifier is obtained.
For testing, a large-scale testing image is cropped into image blocks. Then, HRPN takes these image blocks as input and generates potential outputs as well as hyper feature maps. The final classifier checks these boxes using hyper features. Finally, all the detection results of segments are gathered to integrate the original image. The authors tested their method on UAV images successfully.

\begin{table*}[!h]
	\centering
	\caption{\uppercase{Summary of related literature}}
	\label{my-label}
	\begin{tabular}{|l|l|l|l|l|}
		\hline
		\textbf{Project}                 & \textbf{Goal}                                                                                              & \textbf{Hardware}     & \textbf{Dataset}                                                                  & \textbf{Link to dataset}                                                                                                 \\ \hline
		Qu et al. \cite{qu2016moving}        & Moving vehicle detection                                                                          & UAV + Camera & CATEC UAV                                                                &  -                                                                                                               \\ \hline
		Wang et al. \cite{WANG2016294}     & \begin{tabular}[c]{@{}l@{}}Vehicle detection and \\ tracking\end{tabular}                         & UAV + Camera & not avaiable                                                             & -                                                                                                                \\ \hline
		Ke et al. \cite{ke2017real}       & \begin{tabular}[c]{@{}l@{}}Extraction of traffic\\  flow parameters\end{tabular}                  & UAV + Camera & \begin{tabular}[c]{@{}l@{}}Taken from \\ Beihang University\end{tabular} & -                                                                                                                \\ \hline
		Zhou et al. \cite{zhou2015efficient}     & \begin{tabular}[c]{@{}l@{}}Real-time road detection\\ and tracking\end{tabular}                   & UAV + camera & Self-collected                                                           & https://sites.google.com/site/hailingzhouwei                                                                    \\ \hline
		Leitloff et al. \cite{Leitloff_2014} & Fast vehicle detection                                                                            & UAV + camera & Self-collected                                                           &-                                                                                                                 \\ \hline
		Puri et al. \cite{puri2007statistical}    & \begin{tabular}[c]{@{}l@{}}real-time traffic \\ monitoring and control\end{tabular}               & UAV + camera & Self-collected                                                           &  -                                                                                                               \\ \hline
		Reshma et al. \cite{reshma2016security}  & \begin{tabular}[c]{@{}l@{}}address the security issues\\  of road traffic monitoring\end{tabular} & UAV + RFID   & \begin{tabular}[c]{@{}l@{}}Proteus\\ simulator\end{tabular}              &  -                                                                                                               \\ \hline
		M-Chen et al. \cite{chen2007real}  & traffic surveillance                                                                              & UAV + camera & self-collected                                                           & -                                                                                                                \\ \hline
		Tang et al. \cite{tang2017vehicle}    & vehicle detection                                                                                 & UAV + camera & Munich vehicle dataset                                                   & \begin{tabular}[c]{@{}l@{}}http://pba-freesoftware.eoc.dlr.de/\\ 3K\_VehicleDetection\_dataset.zip\end{tabular} \\ \hline
		Apeltauer et al. \cite{apeltauer2015automatic}     & vehicle trajectory extraction                                                                     & UAV + camera & self-collected                                                           & -                                                                                                                \\ \hline
	\end{tabular}
\end{table*}


\subsection{Use Cases}
\begin{figure}[h]
	\centering
	\includegraphics[width=0.45\textwidth]{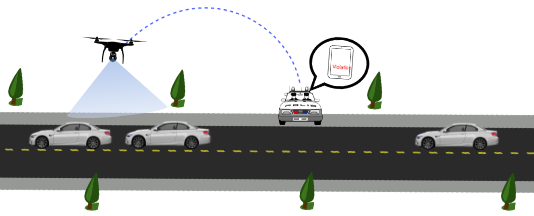}
	\caption{\label{fig:use-caseA}A
		UAV is Used By Police to Catch Traffic Violators \cite{menouar2017uav}.}
\end{figure}

\begin{figure}[t]
	\centering
	\includegraphics[width=0.45\textwidth,]{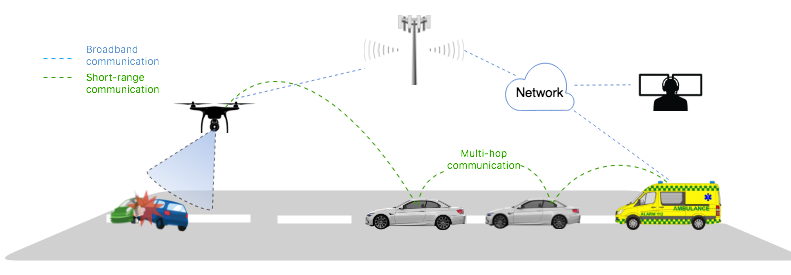}
	\caption{\label{fig:use-caseB}A UAV is Used as a Flying RSU That Broadcasts a Warning About Road Hazards that have been Detected
		in an Area not Pre-Equipped with an RSU (Flying Roadside Units) \cite{menouar2017uav}. \newline}
	\centering
	\includegraphics[width=0.45\textwidth]{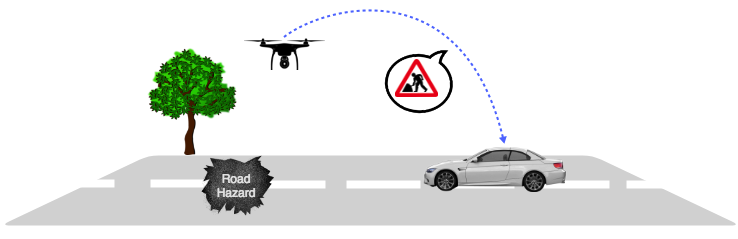}
	\caption{\label{fig:use-caseC}A UAV is Used to Provide the Rescue Team an Advance Report Prior to Reaching the Incident Scene \cite{menouar2017uav}.}
\end{figure}

The major applications of UAVs in transportation include security surveillance, traffic monitoring, inspection of road construction projects, and survey of traffic, rivers, coastlines, pipelines, etc. \cite{6851187}.
Some of the ITS applications that can be enabled by UAVs are as follows: 
\begin{itemize}
	\item Flying Accident Report Agents: Rescue teams can use UAVs to quickly reach accident locations. Flying accident report agents can also be used to deliver first aid kits to accident locations while waiting for rescue teams to arrive \cite{menouar2017uav}. 
	\item Flying Police Eyes: UAVs can be used to fly over different road segments in order to stop vehicle for traffic violations. The UAV can change the traffic light in front of the vehicle to stop it or relay a message to a specific vehicle to stop \cite{menouar2017uav} \cite{menouar2017uav}.
	\item  Flying Roadside Unit: A UAV can be complemented with DSRC to enable a flying RSU. The flying RSU can fly to a specific position to execute a specific application. For example, consider an accident on the highway at a specific segment that is not equipped with any RSU. Then the traffic management center can activate a UAV to fly to the accident location and land at the proper location to broadcast the information and warn all approaching vehicles about a specific incident \cite{menouar2017uav}.
	\item Behavior Recognition Method: UAVs can be used to recognize suspicious or abnormal behavior of ground vehicles moving along with the road traffic \cite{oh2014behaviour}. 
	\item Monitor Pedestrian Traffic: Sutheerakul et al. \cite{SUTHEERAKUL20171717} used UAVs as an alternative data collection technique to monitor pedestrian traffic and evaluate demand and supply characteristics. In fact, they classified their collected data into four areas: the measurement of pedestrian demand, pedestrian characteristics, traffic flow characteristics, and walking facilities and environment.
	\item Flying Dynamic Traffic Signals \cite{menouar2017uav}.
\end{itemize}
UAVs may also be employed for a wide range of transportation operations and planning applications such as following \cite{puri2005survey}:
\begin{itemize}
	\item Incident response.
	\item Monitor freeway conditions.
	\item Coordination among a network of traffic signals.
	\item Traveler information.
	\item Emergency vehicle guidance.
	\item Measurement of typical roadway usage.
	\item Monitor parking lot utilization.
	\item Estimate Origin-Destination (OD) flows.
\end{itemize}
Figures \ref{fig:use-caseA}, \ref{fig:use-caseB}, and \ref{fig:use-caseC} illustrate different applications use-cases of UAVs in smart cities.

\subsection{Legislation}
The Federal Aviation Administration (FAA) approves the civil use of UAVs \cite{FAA}. They can be utilized for public use provided that the UAVs are flown at a certain altitude. 
For maintaining the safety of manned aircrafts and the public, the FAA in the United States has developed rules to regulate the use of small UAVs \cite{vattapparamban2016drones}. For example, the FAA requires small Unmanned Aircraft Systems (UAS) that weigh more than $0.55$ lbs and below $55$ lbs to be registered in their system.
Regulations are broadly organized into two categories; namely, prescriptive regulations and performance-based regulations (PBRs). Prescriptive regulations define what must not be done whereas PBRs indicate what must be attained. $20\%$ of FAA regulations are
PBRs \cite{NAKAMURA2017}. Table X 
describes the rules for operating UAS in the US\cite{FAA}.

\begin{table}
	\caption{ \uppercase{Rules for Operating a UAS.\newline *These Rules are Subject to Waiver.}}
	\label{eyad-1} 
	\begin{tabular}{|l| p{4.75cm}|}
		\hline
		& \begin{center}
			\textbf{Fly for work}
		\end{center} 
		\\ \hline
		Pilot Requirements & Must have Remote Pilot Airman Certificate Must be 16 years old Must pass TSA vetting    \\ \hline
		Aircraft Requirements & Must be less than 55 lbs. Must be registered if over 0.55 lbs. (online) Must undergo pre-flight check to ensure UAS is in condition for safe operation  \\ \hline
		Location Requirements & Class G airspace*   \\ \hline
		Operating Rules & Must keep the aircraft in sight (visual line-of-sight)*
		\newline Must fly under 400 feet* 
		\newline Must fly during the day* Must fly at or below 100 mph*
		\newline Must yield right of way to manned aircraft* \newline Must NOT fly over people*
		\newline Must NOT fly from a moving vehicle* \\ \hline
		Example Applications & Flying for commercial use (e.g. providing aerial surveying or photography services).\newline incidental to a business (e.g. doing roof inspections or real estate photography) \\ \hline
		Legal or Regulatory Basis & Title 14 of the Code of Federal Regulation (14 CFR) Part 107 \\ \hline
	\end{tabular}
	
\end{table}

When violating the FAA regulations, owners of drones can face civil and criminal penalties \cite{vattapparamban2016drones}. The FAA has also provided a smartphone application B4UFLY5 which provides drone users important information about the limitations that pertain to the location they where drone is being operated. The "Know Before You Fly" campaign by the FAA aims to educate the public about UAV safety and responsibilities. For civil operations, the FAA authorization can be received either through Section 333 Exemption (i.e., by issuing a COA), or through a Special Airworthiness Certificate (SAC) in which applicants describe their design, software development, control, along with how and where they intend to fly. The FAA also enforces their regulations along with Law Enforcement Agencies (LEAs) to deter, detect, investigate, and stop unauthorized and unsafe UAV operations \cite{vattapparamban2016drones}.

Overall, the FAA regulations for small UAVs require flying under $400$ feet without obstacles in their vicinity, such that operators maintain a line of sight with the operated UAV at all times. It also requires UAVs not to fly within $5$ miles from an airport unless permission is received from the airport and the control tower; thus, avoiding the endangerment of people and aircrafts. Other FAA regulations require drones not to be operated over public infrastructure (e.g., stadiums) as that may pose dangers to the public.

{The Federal UAS regulations final rule requires drone pilots to keep unmanned aircrafts within visual line of sight and operations are only allowed during daylight and during half-light if the drone is equipped with “anti-collision lights.” The new regulations also initiate height and speed constraints and other operational limits, such as prohibiting flights over unprotected people on the ground who aren’t directly participating in the UAS operation. There is a process through which users can apply to have some of these restrictions waived, while those users currently operating under section 333 exemptions (which allowed commercial use to take place prior to the new rule) are still able to operate depending on their exemptions \cite{regulations}.}

{In Section 107.51 of the FAA regulations \cite{federalregulations}, it is mentioned that a remote pilot in command and the person manipulating the flight controls of the small unmanned aircraft system must consent with all of the following operating limitations when operating a small unmanned aircraft system:
	\begin{itemize}
		\item The ground speed of a small unmanned aircraft may not exceed $87$ knots (i.e., $100$ miles per hour).
		\item The altitude of the small unmanned aircraft cannot be higher than $400$ feet above ground level, unless the small unmanned aircraft:
		(1) Is flown within a $400$-foot radius of a structure; and
		(2) Does not fly higher than $400$ feet above the structure’s immediate uppermost limit.
		\item The minimum flight visibility, as observed from the location of the control station must be no less than $3$ statute miles.
	\end{itemize}

	For critical infrastructure, a legislation has been developed to protect such infrastructure from rogue drone operators. UAVs should not be close to such places when these critical infrastructure exist. The classification of critical infrastructure differs by state, but generally includes facilities such as petroleum refineries, chemical manufacturing facilities, pipelines, waste water treatment facilities, power generation stations, electric utilities, chemical or rubber manufacturing facilities, and other similar facilities \cite{regulationsncls}. 
}

\subsection{Challenges and Future Insights}

There are several challenges and needed future extensions to facilitate the use of UAVs in support of ITS applications:
\begin{itemize}
	
	\item One of the important challenges is to preserve the privacy of sensitive information (e.g., location) from other vehicles and drones \cite{menouar2017uav}. Since usually there is no encryption on UAVs on-board chips, they can be hijacked and subjected to man-in-middle attacks originating up to two kilometers away \cite{vattapparamban2016drones}.
	\item countries should devise registration mechanisms for UAVs operated on their geographic areas. Commercial airplanes and their navigation might be affected by UAVs. So countries must implement rules and regulations for their proper use \cite{mohammed2014uav}.
	\item Developing precise coordination algorithms is one of the challenges that need to be considered to enable ITS UAVs \cite{menouar2017uav}.
	\item Wireless sensors can be utilized to smooth the operations of UAVs. For example, surveillance and live feeds from wireless sensors can be developed for control traffic systems \cite{mohammed2014uav}.
	\item Data fusion of information from diverse sensors, automating image data compression, and stitching of aerial imagery are required techniques\cite{mohammed2014uav}.
	\item Enabling teams of operators to control the UAV and retrieve imagery and sensor information in real-time. To achieve this goal, the development of network-centric infrastructure is required \cite{mohammed2014uav}.
	\item Limited energy, processing capabilities, and signal transmission range \cite{menouar2017uav} are also some of the main issues in UAVs that require more development to contribute to the maturity of the UAV technology
	\item UAVs have slower speeds compared to vehicles driving on highways. However, a possible solution might entail changing the regulations to allow UAVs to fly at higher altitudes. Such regulations would allow UAVs to benefit from high views to compensate the limitation in their speed. Finding the optimal altitude change in support of ITS applications is a research challenge \cite{menouar2017uav}.
	\item Battery technology that allow UAVs to achieve long operational times beyond half an hour is another challenge. Several UAVs can fly together and form a swarm, by which they could overcome their individual limitations in terms of energy efficiency through optimal coordination algorithms \cite{menouar2017uav}.
	Another alternative is for UAVs to use recharge stations. UAVs can be recharged while on the ground, or their depleted batteries can be replaced to minimize interruptions to their service. To this end,  deployment of UAVs, recharge stations, and ground RSUs jointly becomes an interesting but complicated optimization problem \cite{menouar2017uav}.
	\item The detection of multiple vehicles at the same time is another challenge. In \cite{zhang2007video}, Zhang et al. developed several computer-vision  based algorithms and applied them to extract the background image from a video sequence, identify vehicles, detect and remove shadows, and compute pixel-based vehicle lengths for classification.
	\item Truly autonomous operations of UAV swarms are a big challenge, since they need to recognize other UAVs, humans and obstacles to avoid collisions. Therefore, the development of swarm intelligence algorithms that fuse data from diverse sources including location sensors, weather sensors, accelerometers, gyoscopes, RADARs, LIDARs, etc. are needed.
\end{itemize}

\section{Surveillance Applications of UAVs}

In this section, we first present a detailed literature review of surveillance applications of UAVs. Then, based on the review, we summarize the advantages, disadvantages and important concerns of UAV uses in surveillance. Also, we discuss several takeaway lessons for researchers and practitioners in the area, which we believe will help guide the development of effective UAV surveillance applications.
\subsection{Literature review}

The report in \cite{haddal2010homeland} discusses the advantages and disadvantages of employing UAVs along US borders for surveillance and introduces several important issues for the Congress. The advantages include: (1) The usage of UAVs for border surveillance improves the coverage along the remote border sections of the U.S.; and (2) The UAVs provide a much wider coverage than current approaches for border surveillance (e.g., border agents on patrol, stationary surveillance equipment, etc.). On the other hand, the disadvantages include: (1) The high accident rates of UAVs (e.g., inclement weather conditions); and (2) The significant operating costs of a UAV, which are often more than double compared to the costs of a manned aircraft. The report also highlights other issues of UAVs that should be considered by the Congress including: UAV effectiveness, lack of information, coordination with USBP agents, safety concerns, and implementation details.
This report provides a clear view of the advantages, disadvantages and important concerns for the border surveillance using UAVs.  However, it lacks details/examples on each aspect discussed (i.e., only high level summaries are included). Also, other important aspects are missing, for example, the robustness benefits (e.g., human error tolerance) of using the UAVs compared to manned aircraft.

The authors in \cite{maza2011experimental} present a multi-UAV coordination system in the framework of the AWARE project (distributed decision-making architecture suitable for multi-UAV coordination). Generally speaking, a set of tasks are generated and allocated to perform a given mission in an efficient manner according to planning strategies. These tasks are sub-goals that are necessary for achieving the overall goal of the system, and that can be achieved independent of other sub-goals. The key issues include:
\begin{itemize}
	\item {Task allocation}: determines the place (i.e., in which UAV node) that each task should be executed to optimize the performance and to ensure appropriate co-operation among different UAVs.
	\item {Operative perception}: generates and maintains a consistent view of all operating UAVs (e.g., number of sensors equipped).
\end{itemize}
In addition, several types of field experiments are conducted, including: (1) Multi-UAV cooperative area surveillance; (2) Wireless sensor deployment; (3) Fire threat confirmation and extinguishing; (4) Load transportation and deployment with single and multiple UAVs; and (5) People tracking.
To sum up, this paper presents details of the proposed algorithm and shows the results of several real filed experiments. However, it lacks detailed analysis or formal proof of the proposed approach to demonstrate its correctness and effectiveness.

In \cite{wall2011surveillance}, the authors perform analysis for UAVs deployments within war-zones (i.e., Afghanistan, Iraq, and Pakistan), border-zones and urban areas in the USA. The analysis highlights the benefits of UAVs in such scenarios which include: 
\begin{itemize}
	\item {Safety}: Drones insulates operators and allies from direct harm and subjects targets to 'precise' attack.
	\item {Robustness}:  Drones reduce human errors (e.g., moral ambiguity) from political, cultural, and geographical contexts.
\end{itemize}
Most importantly, the analysis uncovers a major limitation in UAV surveillance practices. The usage of drones for surveillance has difficulties in exact target identification and control in risk societies. Potential blurred identities include: (1) insurgent and civilian; (2) criminal and undocumented migrant; and (3) remotely located pilot and front-line soldier.
To sum up, this paper puts more focus on safety and robustness benefits of using UAVs in battle fields and urban areas. Also, it presents the limitation of exact target identification and control. However, it overlooks the coverage benefits and the deployment/implementation limitations of using UAV systems for surveillance.

In \cite{finn2012unmanned}, the authors show the impact of UAV-based surveillance in civil applications on privacy and civil liberties. First, it states that current regulatory mechanisms (e.g., the US Fourth amendment, EU legislation and judicial decisions, and UK legislation) do not adequately address privacy and civil liberties concerns. This is mainly because UAVs are complex, multi-modal surveillance systems that integrate a range of technologies and capabilities. Second, the inadequacy of current legislation mechanisms results in disproportionate impacts on civil liberties for already marginalized population. In conclusion, multi-layered regulatory mechanisms that combine legislative protections with a bottom-up process of privacy and ethical assessment offer the most comprehensive way to adequately address the complexity and heterogeneity of unmanned aircraft systems and their intended deployments.
To sum up, this paper focuses on the law enforcement of privacy and civil liberty aspects of using UAVs for surveillance. However, the regulation recommendation provided is high level and lacks sufficient details and convincing analysis.

In \cite{kingston2008decentralized}, the authors introduce a cooperative perimeter surveillance problem and offer a decentralized solution that accounts for perimeter growth (expanding or contracting) and insertion/deletion of team members. The cooperative perimeter surveillance problem is defined to gather information about the state of the perimeter and transmit that data back to a central base station with as little delay and at the highest rate possible. The proposed solution is presented in Algorithm \ref{alg1}. 
The proposed scheme is described comprehensively and a simple formal proof is provided. However, there is no related works or comparative evaluations provided to justify the effectiveness of the proposed scheme.

\begin{table*}[]
	\normalsize
	\centering
	\caption{\uppercase{Summary of UAV applications for security surveillance}}
	\label{t1}
	\begin{tabular}{|l|l|c|c|}
		\hline
		\multicolumn{1}{|c|}{\textbf{Surveillance}} & \multicolumn{1}{c|}{\textbf{Year}} & \textbf{Research Focus}                                                                                                           & \textbf{Category}            \\ \hline
		\cite{kingston2008decentralized}                                                & 2008                               & UAVs cooperative perimeter surveillance                                                                                           & Multi-UAV cooperation        \\ \hline
		\cite{haddal2010homeland}                                                & 2010                               & UAV with border surveillance                                                                                                      & Homeland security concerns   \\ \hline
		\cite{maza2011experimental}                                                & 2011                               & \begin{tabular}[c]{@{}c@{}}Multi-UAV coordination for disaster management \\ and civil security applications\end{tabular}         & Multi-UAV cooperation       \\ \hline
		\cite{wall2011surveillance}                                                & 2011                               & Concerns of UAVs in battle filed                                                                                                  & Military concerns            \\ \hline
		\cite{birk2011safety}                                                & 2011                               & \begin{tabular}[c]{@{}c@{}}UAV system filed tests with improved \\ photo mapping algorithm\end{tabular}                           & Algorithm improvement        \\ \hline
		\cite{finn2012unmanned}                                                & 2012                               & \begin{tabular}[c]{@{}c@{}}UAV surveillance in civil applications impacts \\ upon privacy and other civil liberties\end{tabular} & Privacy and civil liberties \\ \hline
		\cite{wada2015surveillance}                                                & 2015                               & A small UAV system with related technologies                                                                                      & Product introduction         \\ \hline
		\cite{motlagh2017uav}                                                & 2017                               & UAV-based IoT platform for crowd surveillance                                                                                     & New use case                 \\ \hline
	\end{tabular}
\end{table*}

\begin{algorithm}
	\begin{algorithmic}
	\STATE {\textbf{If}~~~ There is an agreement with the neighbor: \textbf{Then}},
	\STATE (1) Calculate shared border position ;
	\STATE (2) Travel with neighbor to shared border ;
	\STATE (3) Set direction to monitor own segment;
\STATE~~~~	{\textbf{If} Reached perimeter endpoint:~~~ \textbf{Then}}
	\STATE~~~~ Reverse direction.
	\STATE~~~~ {\textbf{ENDIF}} 
\STATE	{\textbf{ENDIF}}
	\end{algorithmic}
	\caption{Proposed Algorithm in \cite{kingston2008decentralized}.}
	\label{alg1}
\end{algorithm}

The authors of \cite{birk2011safety} present the results of using a UAV in two field tests, the 2009 European Land Robot Trials (ELROB-2009) and the 2010 Response Robot Evaluation Exercises (RREE-2010), to investigate different realistic response scenarios. They transplant an improved photo mapping algorithm which is a variant of the Fourier Mellin Invariant (FMI) transform for image representation and processing. They used FMI with the following two modifications: 
\begin{itemize}
	\item A logarithmic representation of the spectral magnitude of the FMI descriptor is used;
	\item A filter on the frequency where the shift is supposed to appear is applied.
\end{itemize}
To sum up, this report mainly discusses two field tests with a description of a transplanted improved photo mapping algorithm. However, the proof/analysis of the improved algorithm is missing. 

In \cite{motlagh2017uav}, the authors discuss a potential application of UAVs in delivering IoT services. A high-level overview is presented and a use case is introduced to demonstrate how UAVs can be used for crowd surveillance based on face recognition. They also study the offloading of video data processing to an MEC (Mobile Edge Computing) node compared to the local processing of video data on board UAVs. The results demonstrate the efficiency of the MEC-based offloading approach in energy consumption, processing time of recognition, and in promptly detecting suspicious persons.
To sum up, this paper successfully demonstrates the benefits of using a single UAV system with IoT devices for crowd surveillance with the introduced video data processing offloading technique. However, further analysis of the coordination of multiple UAVs is missing. 

In \cite{wada2015surveillance}, the authors present the use of small UAVs for specific surveillance scenarios, such as assessment after a disaster. This type of surveillance applications utilizes a system that integrates a wide range of technologies involving: communication, control, sensing, image processing and networking.
This article is a product technical report and the corresponding  manufacturer successfully applied multiple practical techniques in a real UAV surveillance product.

Table \ref{t1} summarizes the eight papers we have reviewed including their research focus and category.

\subsection{Discussion}

\subsubsection*{State-of-the-art Research}

Table \ref{t2} summarizes the advantages, disadvantages and important concerns for the UAV surveillance applications based on our literature review.  We classify the UAV surveillance research works into six categories based on their focus and content:
\begin{itemize}
	\item Multi-UAV cooperation: this line of research focuses on developing new orchestration algorithms or architectures in the presence of multiple UAVs.
	\item Homeland/military security concerns: this line of research provides analysis and suggestions for the authorities with the considerations for Homeland/military security.
	\item Algorithm improvement: this line of research develops new algorithms for more efficient post-processing of the data (e.g., videos or photos captured by the UAVs).
	\item New use case: this line of research integrates UAV system into new application scenarios for potential extra benefits.
	\item Product introduction: this line of research describes mature techniques leveraged by a real UAV surveillance product.
\end{itemize}

\begin{table}[]
	\centering
	\caption{\uppercase{Summary of Advantages, Disadvantages and Important Concerns for Surveillance applications of UAVs}}
	\label{t2}
	\begin{tabular}{|l|l|}
		\hline
		
		\textbf{Advantages}         & \begin{tabular}[c]{@{}l@{}} surveillance coverage and range improvement\\  better safety for human operators\\ robustness and efficiency in surveillance\end{tabular} \\ \hline
		\textbf{Disadvantages}      & \begin{tabular}[c]{@{}l@{}} high accidental rate\\  high operating costs\\  difficulties in surveillance of risk societies.\end{tabular}                               \\ \hline
		\textbf{\begin{tabular}[c]{@{}l@{}}Important \\ Concerns\end{tabular}} & \begin{tabular}[c]{@{}l@{}} co-operation of multi-UAVs\\ post-processing algorithm improvements\\ privacy concerns\\ law enforcement\end{tabular}                 \\ \hline
	\end{tabular}
\end{table}

\subsection{Research Trends and Future Insights}
\subsubsection{Research Trends}
Based on the reviewed literature and our analysis, we identify that the use of research trends (e.g., machine learning algorithms, nano-sensors \cite{chen2013infrared}, short-range communications technologies, etc.) could bring further benefits in the area of UAV surveillance applications:
\begin{itemize}
	\item There are several preferred features of UAV surveillance applications, especially for some military or homeland security use cases: (1) small size: this feature not only reduces the physical attack surface of the UAV but also achieves more secrete UAV patrolling. (2) high resolution photos: this feature helps the data post processing algorithms to draw more accurate conclusions or obtain more insight findings; (3) low energy consumption: this feature allows the UAV to operate for a longer time of period such that to achieve more complex tasks.
	
	Therefore, to capture more accurate data in a more efficient and secret way, advanced sensor technologies should be considered. Such as nano-sensors (small size), ultra-high-resolution image sensors (accurate data) \cite{jiang2015ultrahigh}, and energy efficient sensors (low energy consumption) \cite{folea2015low}.

	\item To develop more efficient and accurate multi-UAV cooperation algorithms or data post-processing algorithms, advanced machine learning algorithms (e.g., deep learning \cite{lecun2015deep}) could be utilized to achieve better performance and faster response.
	Specifically, multi-UAV cooperation brings more benefits than single UAV surveillance, such as more wider surveillance scope, higher error tolerance, and faster task completion time. However, multi-UAV surveillance requires more advanced data collection, sharing and processing algorithms. Processing of huge amount data is time consuming and much more complex. Applying advanced machine learning algorithms (e.g., deep learning algorithm) could help the UAV system to draw better conclusions in a short period of time. For example, due to its improved data processing models, deep learning algorithms could help to obtain new findings from existing data and to get more concise and reliable analysis results.

\end{itemize}    

\subsubsection{{Future Insights}} based on our literature review and analysis, to develop or deploy effective UAV surveillance applications, we suggest the following:
\begin{itemize}
	\item Select/design appropriate UAV system based on the state/federal laws and individual budget. For example, the design of UAVs surveillance systems should take the privacy and security requirements enforced by the local laws;
	\item Develop efficient customized algorithms (e.g., co-operation algorithm of multi-UAVs, photo mapping algorithms, etc.) based on system requirements. Multi-UAVs co-operation algorithm could achieve more efficient surveillance (e.g., optimum patrolling routes with minimum power consumption). Also, advanced data post-processing algorithms help the users to draw more accurate conclusions;
	\item Conduct various field experiments to verify newly proposed systems. Extensive field experiments in various surveillance conditions (e.g., day time, night time, sunshine day, cloudy day, etc.) should be conducted to verify the effectiveness and robustness of the UAVs systems.
\end{itemize}

\section{Providing Wireless Coverage}
\label{comm}
UAVs can be used to provide wireless coverage during
emergency cases where each UAV serves as an aerial wireless
base station when the cellular network goes down~\cite{bupe2015relief}. They can also be used to supplement the ground base station in order to provide better coverage and higher data rates for users~\cite{bor2016efficient}. In this section, we present the aerial wireless base stations use cases, UAV links and UAV channel characteristics. We also show the path loss models for UAVs, classify them based on environment, altitude and telecommunication link, and present some challenges facing these path loss models. Moreover, we present UAV deployment strategies that optimize several objective functions of interest. Then, we discuss the challenges facing UAV interference mitigation techniques. Finally, we present the research trends and future insights for aerial wireless base stations.

\begin{figure}[t]
	\centering
	\includegraphics[scale=0.29]{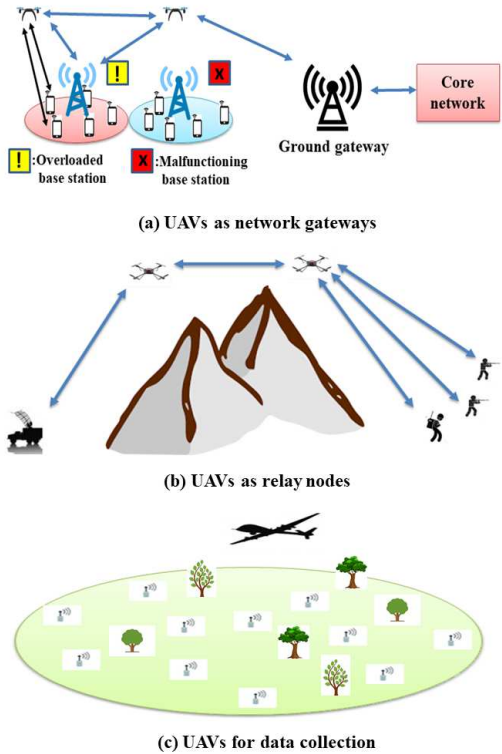}
	\caption{Typical Use Cases of UAV-Aided Wireless Communications.}
	\label{haz9}
\end{figure}

\begin{figure*}[!h]
	\centering
	\begin{forest}
		for tree={
			align=center,
			parent anchor=south,
			child anchor=north,
			font=\footnotesize,
			edge={thick, -{Stealth[]}},
			l sep+=10pt,
			edge path={
				\noexpand\path [draw, \forestoption{edge}] (!u.parent anchor) -- +(0,-10pt) -| (.child anchor)\forestoption{edge label};
			},
			if level=0{
				inner xsep=0pt,
				tikz={\draw [thick] (.south east) -- (.south west);}
			}{}
		}
		[  UAV links
		[Control links
		[Command and\\control from\\ GCS to UAVs		
		]
		[UAV status\\ report from\\ UAVs to GCS
		]
		[Sense and avoid\\ information\\
		among UAVs		
		]
		]
		[Data links
		[Direct ground\\ mobile-UAV\\ communication
		]
		[UAV-base station\\ and UAV-gateway\\ wireless backhaul
		]
		[UAV-UAV \\wireless backhaul
		]
		]
		]
	\end{forest}
	\caption{Applications of UAV Links.}
	\label{haz6}
\end{figure*}
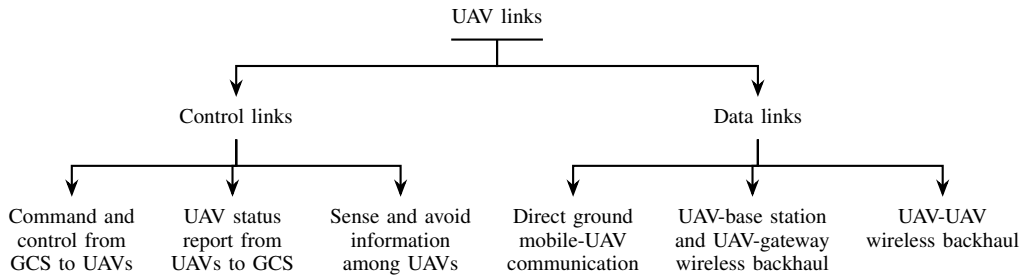

\begin{table*}[!t]
	\scriptsize
	\renewcommand{\arraystretch}{1.45}
	\caption{\uppercase{A comparison between data and control links of UAV\lowercase{s}}}
	\label{tablehaz3}
	\centering
	\begin{tabular}{|c|c|c|c|c|}
		\hline
		Type of link&Security requirement&Latency requirement&Capacity requirement&Frequency bands\\
		\hline
		Control link&High&High&Low&L-band (960-977MHz) \\
		&&&&C-band (5030-5091MHz)\\
		\hline
		Data link&Low&Low&High&450 MHz to mmWave \\
		\hline 
	\end{tabular}
\end{table*}
\begin{figure}[!t]
	\centering
	\includegraphics[scale=0.355]{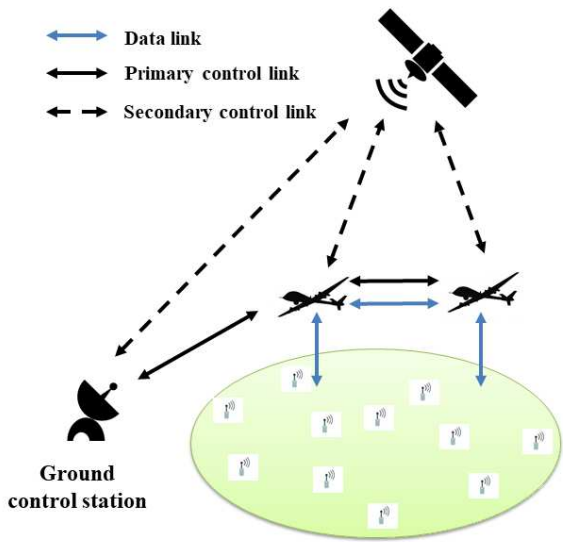}
	\caption{Basic Networking Architecture of UAV-Aided Wireless Communications.}
	\label{haz7}
\end{figure}

\begin{figure*}[h]
	\centering
	\begin{forest}
		for tree={
			align=center,
			parent anchor=south,
			child anchor=north,
			font=\footnotesize,
			edge={thick, -{Stealth[]}},
			l sep+=10pt,
			edge path={
				\noexpand\path [draw, \forestoption{edge}] (!u.parent anchor) -- +(0,-10pt) -| (.child anchor)\forestoption{edge label};
			},
			if level=0{
				inner xsep=0pt,
				tikz={\draw [thick] (.south east) -- (.south west);}
			}{}
		}
		[  Classification of Path loss Models for UAVs
		[Altitude of UAV
		[LAP		
		]
		[HAP
		]
		]
		[Environment
		[Outdoor-ground users
		]
		[Indoor users
		]
		[UAV-UAV
		]
		]
		[Telecommunication link
		[Downlink
		]
		[Uplink
		]
		]
		]
	\end{forest}
	\caption{Classification of Path Loss Models for UAVs.}
	\label{haz1}
\end{figure*}
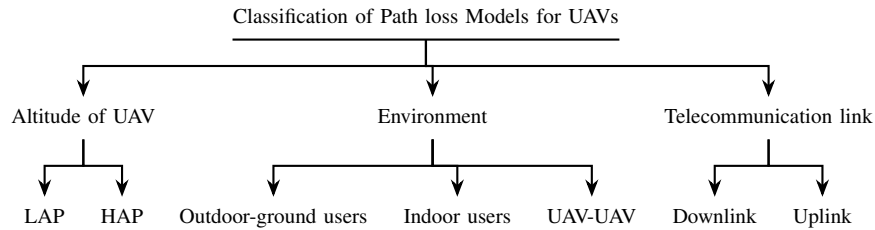

\subsection{Aerial Wireless Base Stations Use Cases}
The authors in~\cite{zeng2016wireless,jawhar2017communication,zhao2017beam} present the typical use cases of aerial wireless base stations which are discussed in the following:
\begin{itemize}
	\item {UAVs for ubiquitous coverage:} UAVs are utilized to assist wireless network in providing seamless wireless coverage within the serving area. Two example scenarios are rapid service recovery after disaster situations, and when the cellular network service is not available or it is unable to serve all ground users as shown in Figure~\ref{haz9}.a.
	\item {UAVs as network gateways: }In remote geographic or disaster stricken areas, UAVs can be used as gateway nodes to provide connectivity to backbone networks, communication infrastructure, or the Internet.
	\item {UAVs as relay nodes:} UAVs can be utilized as relay nodes to provide wireless connectivity between two or more distant wireless devices without reliable direct communication links as shown in Figure~\ref{haz9}.b.
	\item {UAVs for data collection:} UAVs are utilized to gather delay-tolerant information from a large number of distributed wireless devices. An example is wireless sensors in precision agriculture applications as shown in Figure~\ref{haz9}.c.
	\item {UAVs for worldwide coverage:} UAV-satellite communication is an essential component for building the integrated space-air-ground network to provide high data rates anywhere, anytime, and towards the seamless wide area coverage.
\end{itemize}
\subsection{UAV Links and Channel Characteristics }
\subsubsection{{Control Links}}
The control links are fundamental to guarantee the safe operation of all UAVs.  These links must be low-latency, highly reliable, and secure two-way communications, usually with low data rate requirement for safety-critical information exchange among UAVs, and additionally between the UAV and ground control stations. The control links information flow can be classified into three types: 1) command and control from ground control stations to UAVs; 2) UAV status report from UAVs to ground; 3) sense-and-avoid information among UAVs. Also for autonomous UAV, which can fulfill missions depending on intelligent devices without real-time human control, the control links are important in case of emergency when human intervention is needed~\cite{zeng2016wireless}. There are two frequency bands allocated for the control links, namely the L-band (960-977MHz) and the C-band (5030-5091MHz)~\cite{matolak2015unmanned}. For delay reasons, we always prefer the primary control links between ground control stations to UAVs, but the secondary control links via satellite could also be utilized as a backup to enhance reliability and robustness. Another key necessity for the control links is the high security. Specifically, efficient security mechanisms should be utilized to avoid the so-called ghost control scenario, a possibly disastrous circumstance in which the UAVs are controlled by unapproved operators via spoofed control or navigation signals. Accordingly, practical authentication techniques, perhaps supplemented by the emerging physical layer security techniques, should be applied for control links. Compared to data links, the control links usually have lower tolerance in terms of security and latency requirements~\cite{zeng2016wireless}.
\subsubsection{{Data Links}}
The purpose of using data links is to support task-related communications for the ground terminals which include ground base stations, mobile terminals, gateway nodes, wireless sensors, etc. The data links of UAVs need to provide the following communication services: 1) Direct mobile-UAV communication; 2) UAV-base station and UAV-gateway wireless backhaul; 3) UAV-UAV wireless backhaul. The capacity requirement of these data links depends on the communication services, ranging from low capacity (kbps) in UAV-sensor links to high speed (Gbps) in UAV-gateway wireless backhaul. The UAV data links could utilize the existing band assigned for the particular communication services to enhance performance, e.g., using millimeter wave band~\cite{rappaport2014millimeter} and free space optics~\cite{alzenad2016fso} for high capacity UAV-UAV wireless backhaul~\cite{zeng2016wireless}. In Figure~\ref{haz6}, we show the applications of UAV links. In Table~\ref{tablehaz3}, we make a comparison between control and data links.
\subsubsection{{UAV Channel Characteristics}}
Both control and data channels in UAV communication networks consist of two primary types of channels, UAV-ground and UAV-UAV channels as shown in Figure~\ref{haz7}. These channels have several unique characteristics compared with the characteristics of terrestrial communication channels. While line of sight links are expected for UAV-ground channels in most cases, they could also be occasionally blocked by obstacles such as buildings. For low-altitude UAVs, the UAV-ground channels may also suffer a number of multipath components due to reflection, scattering, and diffraction by buildings, ground surface, etc. The UAV-UAV wireless channels are line of sight dominated and thus the effect of multipath is minimal compared to that experienced in UAV-ground or ground-ground channels.  Due to the continuous movements of UAVs with different velocities, the UAV-to-UAV wireless channels will have high Doppler frequencies, especially the fixed wing UAVs.  On one hand, we can utilize the dominance of the line of sight channels to achieve high-capacity for emerging mmWave communications. On the other hand, due to the continuous movements of UAVs with different velocities coupled with the higher carrier frequency in the mmWave band, the doppler shift will increase~\cite{zeng2016wireless}.
\subsection{Path Loss Models}
Path loss is an essential in the design and analysis of wireless communication channels and represents the amount of reduction in power density of a transmitted signal. The characteristics of aerial wireless channels are different than the terrestrial wireless channels due to the variations in the propagation environments and hence the path loss models for UAVs are also different than the traditional path loss models for terrestrial wireless channels. We classify the UAV path loss models based on environment, altitude and telecommunication link as shown in Figure~\ref{haz1}
\subsubsection{{Air-to-Ground Path Loss for Low Altitude Platforms}}
\begin{figure*}[t]
	\begin{minipage}[b]{0.35\linewidth}
		\centering
		\includegraphics[width=\textwidth]{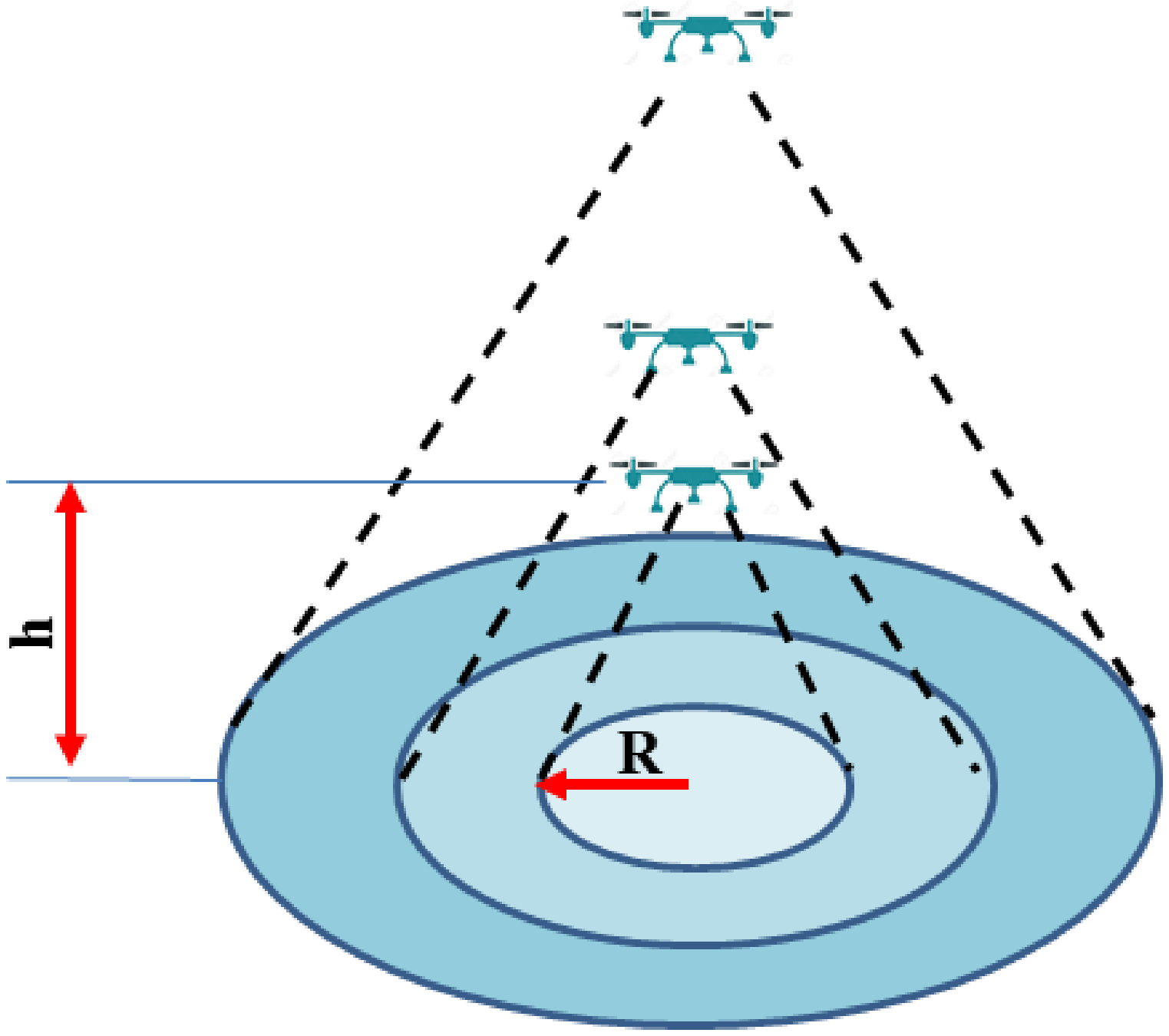}
		\caption{Coverage Zone By a Low Altitude UAV.}
		\label{haz2}
	\end{minipage}
	\hspace{3.6 cm}
	\begin{minipage}[b]{0.32\linewidth}
		\centering
		\includegraphics[width=\textwidth]{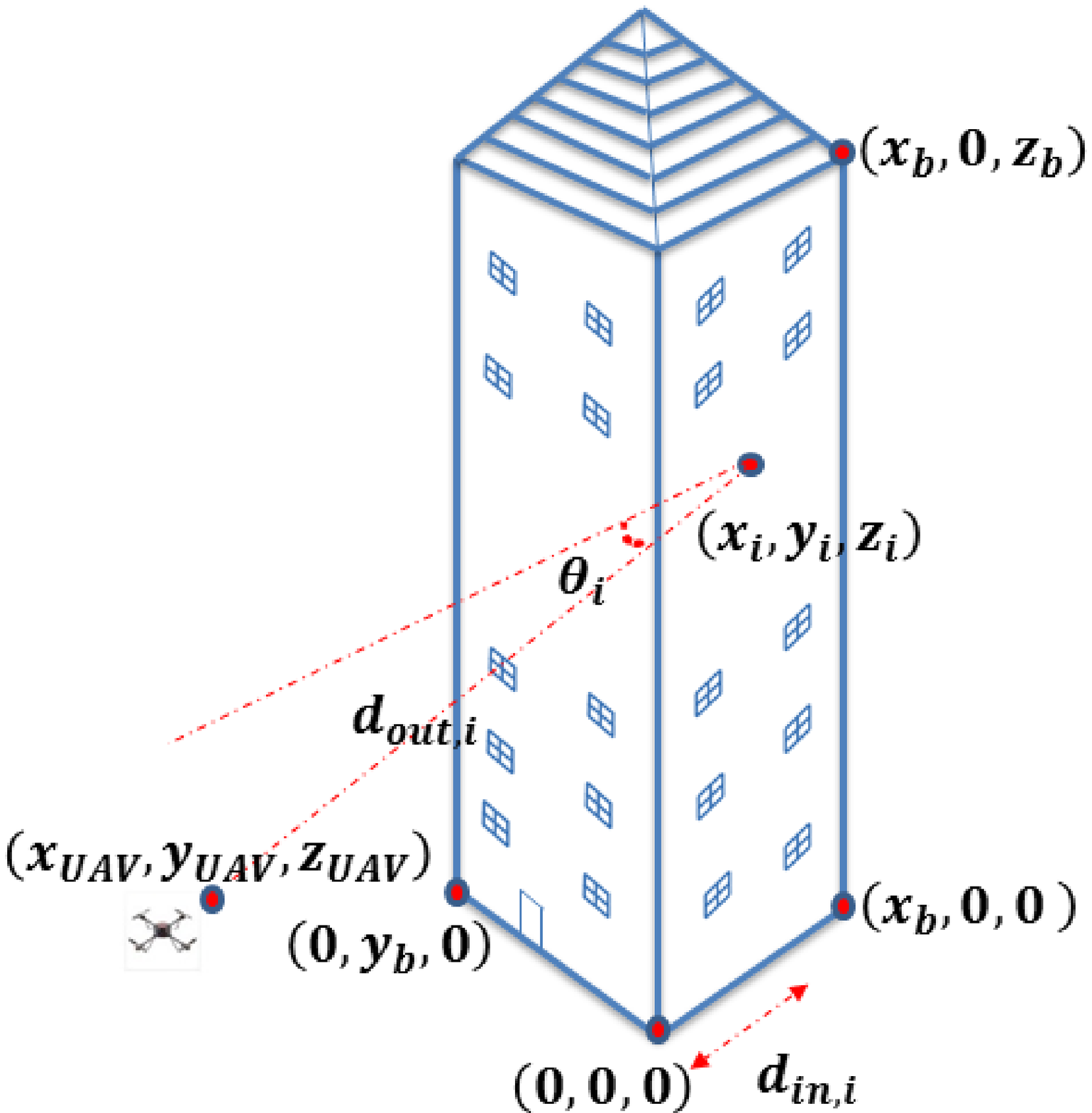}
		\caption{Providing Wireless Coverage for Indoor Users.}
		\label{haz3}
	\end{minipage}
\end{figure*}

The authors in~\cite{al2014modeling} present a statistical propagation model for predicting the path loss between a low altitude UAV and a Ground terminal. In this model, the authors assume that a UAV transmits data to more than 37000 ground receivers and use the Wireless InSite ray tracing software to model three types of rays (Direct, Reflected and Diffracted). Based on the simulation results, they divide the receivers to three groups. The first group corresponds to receivers that have Line-of-Sight or near-Line-of-Sight conditions. The second group corresponds to receivers with non Line-of-Sight condition, but still receiving coverage via strong reflection and refraction. The third group corresponds to receivers that have deep fading conditions resulting from consecutive reflections and diffractions, the third group only represents 3\% of receivers. Therefore, based on the first and second groups, the authors present the path loss model as a function of the altitude $h$ and the coverage radius $R$ as shown in Figure~\ref{haz2} and it is given as follows:

\begin{equation}
\begin{split}
L(h,R)=P(LOS)\times L_{LOS}+P(NLOS)\times L_{NLOS}
\end{split}
\end{equation}
\begin{equation}
\begin{split}
P(LOS)=\dfrac{1}{1+\alpha.exp(-\beta[\frac{180}{\pi}\theta -\alpha] )}
\end{split}
\end{equation}
\begin{equation}
\begin{split}
L_{LOS}(dB)=20log(\dfrac{4\pi f_cd}{c})+\zeta_{LOS}
\end{split}
\end{equation}
\begin{equation}
\begin{split}
L_{NLOS}(dB)=20log(\dfrac{4\pi f_cd}{c})+\zeta_{NLOS}
\end{split}
\end{equation}

where $P(LOS)$ is the probability of having line of sight (LOS) connection at an elevation angle of $\theta$, $P(NLOS)$ is the probability of having non LOS connection and it equals (1- $P(LOS)$), $L_{LOS}$ and $L_{NLOS}$ are the average path loss for LOS and NLOS paths. In equations (2), (3) and (4), $\alpha$ and $\beta$ are constant values which depend on the environment, $f_c$ is the carrier frequency, $d$ is the distance between the UAV and the ground user, $c$ is the speed of the light, $\zeta_{LOS}$ and $\zeta_{NLOS}$ are the average additional loss which depends on the environment. 

In this path loss model, the authors assume that all users are outdoor and the location of each user can be represented by an outdoor 2D point. These assumptions limit the applicability of this model when one needs to consider indoor users. The authors of~\cite{mozaffari2015drone} describe the tradeoff in this model. At a low altitude, the path loss between the UAV and the ground user decreases, while the probability of line of sight links also decreases. On the other hand, at a high altitude line of sight connections exist with a high probability, while the path loss increases.
\subsubsection{{Outdoor-to-Indoor Path Loss Model}}
The Air-to-Ground path loss model presented in~\cite{al2014modeling} is not appropriate when we consider wireless coverage for indoor users, because this model assumes that all users are outdoor and located at 2D points. In~\cite{haz2017efficient}, the authors adopt the Outdoor-Indoor path loss model, certified by the International Telecommunication Union (ITU)~\cite{series2009guidelines} to provide wireless coverage for indoor users using UAV as shown in Figure~\ref{haz3}. The path loss is given as follows: 
\begin{equation}
\begin{split}
L=L_F+L_B+L_I=(wlog_{10}d_{out}+wlog_{10}f+g_1)\\
+(g_2+g_3(1-cos\theta_i)^2)+(g_4d_{in})       
\end{split}
\end{equation}
where $L_F$ is the free space path loss, $L_B$ is the building penetration loss, and $L_I$ is the indoor loss. Also, $d_{out}$ is the distance between the UAV and indoor user, $\theta_i$ is the incident angle, and $d_{in}$ is the indoor distance of the user inside the building. In this model, we also have $w=20$, $g_1$=32.4, $g_2$=14, $g_3$=15, $g_4$=0.5 and $f$ is the carrier frequency. The authors of~\cite{haz2017efficient} describe the tradeoff in the above model when the horizontal distance between the UAV and a user changes. When this horizontal distance increases, the free space path loss (i.e.,$L_F$) increases as $d_{out}$ increases, while the building penetration loss (i.e., $L_B$) decreases as the incident angle (i.e., $\theta_i$) decreases. 
\subsubsection{{Cellular-to-UAV Path Loss Model}}
\begin{figure}[!t]
	\centering
	\includegraphics[scale=0.27]{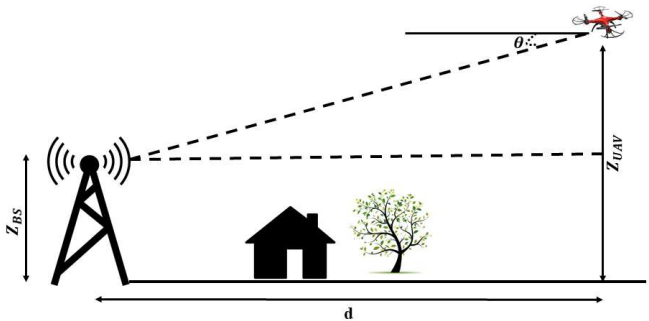}
	\caption{Cellular-to-UAV Path Loss Parameters.}
	\label{haz4}
\end{figure}

In~\cite{al2017modeling}, the authors model the statistical behavior of the path loss from a cellular base station towards a flying UAV. They present the path loss model based on extensive field experiments that involve collecting both terrestrial and aerial coverage samples. They report the value of the path loss as a function of the depression angle and the terrestrial coverage beneath the UAV as shown in Figure~\ref{haz4}. The path loss is given as follows:
\begin{equation}
\begin{split}
L(d,\theta)=L_{ter}(d)+\eta(\theta)+X_{uav}(\theta)=10\alpha log(d)+\\A(\theta -\theta_o)exp(-\dfrac{\theta -\theta_o}{B})+\eta_o+N(0,a\theta +\sigma_o)
\end{split}
\end{equation}
where $L_{ter}(d)$ is the mean terrestrial path-loss at a given terrestrial distance $d$ from the cellular base station, $\eta (\theta)$ is the excess aerial path-loss, $X_{uav}(\theta)$ is a Gaussian random variable with an angle-dependent standard deviation $\sigma_{uav}(\theta)$ representing the shadowing component, $\theta$ is depression angle between the UAV and the cellular base station, $\alpha$ is the path-loss exponent. Also, $A, B, \theta_o$  and $\eta_o$ are fitting parameters.
\subsubsection{{Air-to-Ground Path Loss for High Altitude Platforms}}
The authors in~\cite{holis2008elevation} present an empirical propagation prediction model for mobile communications from high altitude platforms in built-up areas, where the frequency band is $2\textendash6$ GHz. The probability of LOS paths between the UAV and a ground user and the additional shadowing path loss for NLOS paths are presented as a function of the elevation angle $(\theta)$. The path loss model is defined for four different types of built-up area (Suburban area, Urban area, Dense urban area and Urban high-rise area). The path loss is given as follows:
\begin{equation}
\begin{split}
L=P(LOS)\times L_{LOS}+P(NLOS)\times L_{NLOS}
\end{split}
\end{equation}
\begin{equation}
\begin{split}
P(LOS)=a-\dfrac{a-b}{1+(\frac{\theta-c}{d})^e}
\end{split}
\end{equation}
\begin{equation}
\begin{split}
L_{LOS}(dB)=20log(d_{km})+20log(f_{GHz})+92.4+\zeta_{LOS}
\end{split}
\end{equation}
\begin{equation}
\begin{split}
L_{NLOS}(dB)=20log(d_{km})+20log(f_{GHz})+92.4+L_s\\+\zeta_{NLOS}
\end{split}
\end{equation}
In equation (7), $P(LOS)$ is the probability of having line of sight (LOS) connection at an evaluation angle of $\theta$, $P(NLOS)$ is the probability of having non LOS connection and it equals (1- $P(LOS)$), $L_{LOS}$ and $L_{NLOS}$ are the average path loss for LOS and NLOS paths. In equation (8), $a, b, c, d$ and $e$ are the empirical parameters. In equations (9) and (10),  $d_{km}$ is the distance between the UAV and the ground user in km, $f_{GHz}$ is the frequency in GHz, $L_s$ is a random shadowing in dB as a function of the elevation angle $(\theta)$, $\zeta_{LOS}$ and $\zeta_{NLOS}$ are the average additional losses which depend on the environment.
\subsubsection{{UAV to UAV path Loss Model}}
In general, the UAV-to-UAV wireless channels are line of sight dominated, so that the free space path loss can be adopted for the aerial channels. Due to the continuous movements of UAVs with different velocities, the UAV-to-UAV wireless channels will have high Doppler frequencies (especially the fixed wing UAVs). 

In Table~\ref{tablehaz1}, we make a comparison among the path loss models based on operating frequency, altitude, environment, type of link, type of experiments and challenges.
\begin{table*}[!h]
	\scriptsize
	\renewcommand{\arraystretch}{1.3}
	\caption{\uppercase{comparison among path loss models}}
	\label{tablehaz1}
	\centering
	\begin{tabular}{|c|c|c|c|c|c|c|}
		\hline
		Path loss model& Frequency band&Altitude&Environment&Type of link&Type of experiments&Challenges\\
		\hline 
		~\cite{al2014modeling}	&2 GHz&LAP&Outdoor-ground users&Downlink&Simulations&1) The authors didn’t consider the indoor users.\\ &&&&&& 2) The authors  didn't consider different 5G {~~~} \\&&&&&&frequency bands.{~~~~~~~~~~~~~~~~~~~~~~~~~~~~~~}\\
		&&&&&&3) The locations of outdoor users are 2D.{~~~~~~}\\
		\hline
		~\cite{series2009guidelines}	&2 GHz to 6 GHz&LAP&Indoor users&Downlink&Real experiments&1) The authors didn’t consider the different {~~~~}\\
		&&&&&& types of the building structures.{~~~~~~~~~~}\\
		&&&&&& 2) The authors  didn't consider different 5G {~~} \\&&&&&&frequency bands.{~~~~~~~~~~~~~~~~~~~~~~~~~~~~}\\
		\hline
		~\cite{al2017modeling}	&850 MHz&LAP&Outdoor cellular&Uplink&Real experiments&1) The authors didn’t investigate denser urban\\ 
		&&&base station&&& environments.{~~~~~~~~~~~~~~~~~~~~~~~~~~~~~} \\ &&&&&&
		2) The authors  didn't consider different 5G {~} \\&&&&&&frequency bands such as mmWave.{~~~}\\
		\hline
		~\cite{holis2008elevation}	&2 GHz to 6 GHz&HAP&Outdoor-ground users&Downlink&Simulations&1) The authors didn’t consider the indoor users{~}\\ &&&&&& 2) The authors  didn't consider different 5G {~~} \\&&&&&&frequency bands.{~~~~~~~~~~~~~~~~~~~~~~~~~~~~~}\\
		&&&&&&3) The locations of outdoor users are 2D.{~~~~~~}\\
		\hline
		Free space	&All frequency&LAP,&UAV to UAV &Uplink, &Real experiments& 1) High Doppler shift.{~~~~~~~~~~~~~~~~~~~~~~~~~~~~} \\
		&bands&HAP&channels&downlink&&\\
		\hline
	\end{tabular}
\end{table*}

\subsection{UAV Deployment Strategies}
\begin{figure*}[h] 
	\centering
	\begin{tikzpicture}[font=\scriptsize]
	\tikzset{every node/.style=
		{align=center, minimum height=46pt, text width=100pt}}
	\node[,draw=black] (b1) {Deployment strategies 
		for minimizing the transmit power of UAVs
	};
	\node[right=5pt,draw=black] (b2) at (b1.east) {Deployment strategies for maximizing the wireless coverage of UAVs
	};
	\node[right=5pt,draw=black] (b3) at (b2.east) {Deployment strategies 
		for minimizing the
		number of UAVs required to perform task
		
	};  
	\node[right=5pt,draw=black] (b4) at (b3.east) {Deployment strategies 
		to collect data using 
		UAVs
		
	};
	\node[above=10pt, text width=160pt,draw=black] (top) at ($(b2.north)!.5!(b3.north)$) {\small{Classification of Deployment strategies for UAVs
	}};
	\coordinate (atop) at ($(top.south) + (0,-5pt)$);
	\coordinate (btop) at ($(b3.south) + (0,-5pt)$);
	\draw[thick] (top.south) -- (atop)
	(b1.north) |- (atop) -| (b4.north)
	(b2.north) |- (atop) -| (b3.north);
	
	
	
	
	\end{tikzpicture}
	\caption{Classification of Deployment Strategies for UAVs.
	}
	\label{haz5}
\end{figure*}
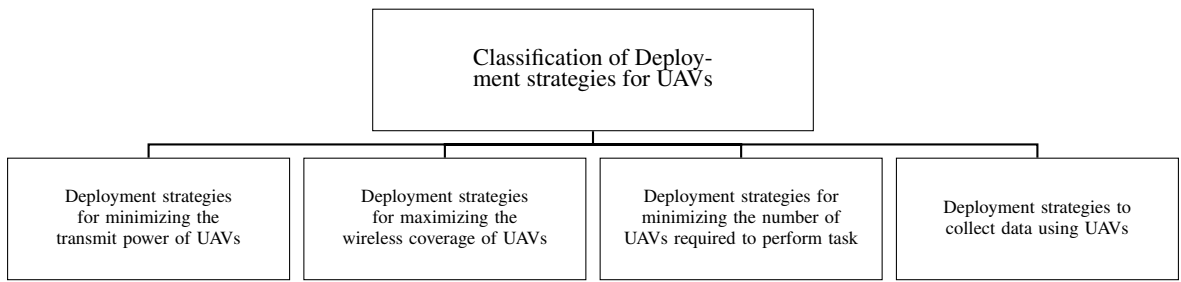

UAVs deployment problem is gaining significant importance in UAV-based wireless communications where the performance of the aerial wireless network depends on the deployment strategy and the 3D placements of UAVs. In this section, we classify the UAV deployment strategies based on the objective functions as shown in Figure~\ref{haz5}.
\subsubsection{{Deployment Strategies for Minimizing the Transmit Power of UAVs}}
The authors in~\cite{mozaffari2016optimal} propose an efficient deployment framework for deploying the aerial base stations, where the goal is to minimize the total required transmit power of UAVs while satisfying the users rate requirements. They apply the optimal transport theory to obtain the optimal cell association and derive the optimal UAV's locations using the facility location framework. The authors in~\cite{mozaffari2015drone} investigate the downlink coverage performance of a UAV, where the objective is to find the optimal UAV altitude which leads to the maximum ground coverage and the minimum transmit power. The authors in~\cite{haz2017efficient} propose using a single UAV to provide wireless coverage for indoor users inside a high-rise building under disaster situations. They study the problem of efficient UAV placement, where the objective is to minimize the total transmit power required to cover the entire high-rise building. In~\cite{shakhatreh2017efficient}, the authors propose a particle swarm optimization algorithm to find an efficient 3D placement of a UAV that minimizes the total transmit power required to cover the indoor users. The authors in~\cite{MICC17}
utilize UAVs to provide wireless coverage for indoor and outdoor users in massively crowded events, where the objective is to find the optimal UAV placement which lead to the minimum transmit power. In~\cite{alzenad20173d}, the authors propose an optimal placement algorithm for UAV that maximizes the number of covered users using the minimum transmit power. The algorithm decouple the UAV deployment problem in the vertical and horizontal dimensions without any loss of optimality. The authors in~\cite{mozaffari2016unmanned} consider two types of
users in the network: the downlink users served by the UAV and device-to-device users that communicate directly with one another. In the mobile UAV scenario, using the disk covering problem, the entire target geographical area can be completely covered by the UAV in a shortest time with a minimum required
transmit power.  They also derive the overall outage probability for device-to-device users, and show that the outage probability increases as the number of stop points that the UAV needs to completely cover the area increases.
\subsubsection{{Deployment Strategies for Maximizing the Wireless Coverage of UAVs}}
In~\cite{bor2016efficient}, the authors highlight the properties of the UAV placement problem, and formulate it as a 3D placement problem with the objective of maximizing the revenue of the network, which is proportional to the number of users covered by a UAV. They formulate an equivalent problem which can be solved efficiently to find the size of the coverage region and the altitude of a UAV. The authors in~\cite{mozaffari2016efficient} study the optimal
deployment of UAVs equipped with directional antennas, using
circle packing theory. The 3D locations of the UAVs are
determined in a way that the total coverage area is maximized. In~\cite{kalantari2017backhaul}, the authors introduce network-centric and user-centric approaches, the optimal 3D backhaul-aware placement of a UAV is found for each approach. In the network-centric approach, the network tries to serve as many users as possible, regardless of their rate requirements. In the user-centric approach, the users are determined based on the priority. The total number of served users and sum-rates are maximized in the network-centric and user-centric frameworks. The authors in~\cite{alzenad20173da} study an efficient 3D UAV placement that maximizes the number of covered users with different Quality-of-Service requirements. They model the placement problem as a multiple circles placement problem and propose an optimal placement algorithm that utilizes an exhaustive search over a one-dimensional parameter in a closed region. They propose a low-complexity algorithm, maximal weighted area algorithm, to tackle the placement problem. In~\cite{shakhatreh2017micc}, the authors aim to maximize the indoor wireless coverage using UAVs equipped with directional antennas. They present two methods to place the UAVs; providing wireless coverage from one building side and from two building sides. The authors in~\cite{shah2017distributed} utilize UAVs-hubs to provide connectivity to small-cell base stations with the core network. The goal is to find the best possible association of the small cell base stations with the UAVs-hubs such that the sum-rate of the overall system is maximized depending on a number of factors including  maximum backhaul data rate of the link between the core network and mother-UAV-hub, maximum bandwidth of each UAV-hub available for small-cell base stations, maximum number of links that every UAV-hub can support and minimum signal-to-interference-plus-noise ratio. They present an efficient and distributed solution of the designed problem, which performs a greedy search to maximize the sum rate of the overall network. In~\cite{mozaffari2017wireless}, the authors propose an efficient framework for optimizing the performance of UAV-based wireless systems in terms of the average number of bits transmitted to users and UAVs's hovering duration. They investigate two scenarios: UAV communication under hover time constraints and UAV communication under load constraints. In the first scenario, given the maximum possible hover time of each UAV, the total data service under user fairness considerations is maximized. They utilize the framework of optimal transport theory and prove that the cell partitioning problem is equivalent to a convex optimization problem. Then, they propose a gradient-based algorithm for optimally partitioning the geographical area based on the users's distribution, hover times, and locations of the UAVs. In the second scenario, given the load requirement of each user at a given location, they minimize the average hover time needed for completely serving the ground users by exploiting the optimal transport theory.

 \begin{table*}[!h]
 	\scriptsize
 	\renewcommand{\arraystretch}{1.5}
 	\caption{\uppercase{comparison among research papers related to UAV deployment strategies based on objective functions}}
 	\label{tablehaz2}
 	\centering
 	\begin{tabular}{|c|c|c|c|c|c|c|}
 		\hline
 		Reference&Number of UAVs&Type of environment&Type of antenna&Operating frequency&Objective function&Deployment strategy\\
 		\hline 
 		~\cite{mozaffari2016optimal}&Multiple UAVs&Outdoor&Omni-directional&2 GHz&Minimizing the transmit power&Optimal transport theory\\
 		&&&&&of UAVs&\\
 		\hline
 		~\cite{mozaffari2015drone}&Single UAV&Outdoor&Directional&2 GHz&Minimizing the transmit power&Closed-form expression \\
 		&&&&&of UAV& for the UAV placement\\
 		\hline
 		~\cite{haz2017efficient}&Single UAV&Indoor&Omni-directional&2 GHz&Minimizing the transmit power& Gradient descent \\
 		&&&&&of UAV&  algorithm\\
 		\hline
 		~\cite{shakhatreh2017efficient}	&Single UAV&Indoor&Omni-directional&2 GHz&Minimizing the transmit power& Particle swarm \\
 		&&&&&of UAV&  optimization\\
 		\hline
 		~\cite{MICC17}&Single UAV&Outdoor-Indoor&Omni-directional&2 GHz&Minimizing the transmit power& Particle swarm\\
 		&&&&&of UAV&optimization, K-means  \\
 		&&&&&&with ternary search \\
 		&&&&&&algorithms\\
 		\hline
 		~\cite{alzenad20173d}&Single UAV&Outdoor&Directional&2 GHz&Minimizing the transmit power& Optimal 3D placement\\
 		&&&&&of UAV&  algorithm\\
 		\hline
 		~\cite{mozaffari2016unmanned}&Single UAV&Outdoor&Directional&2 GHz&Minimizing the transmit power&Disk covering problem\\
 		&&&&&of UAV& \\
 		\hline
 		~\cite{bor2016efficient}&Single UAV&Outdoor&Directional&2.5 GHz&Maximizing the wireless coverage&Bisection search \\
 		&&&&&of UAV&algorithm\\
 		\hline
 		~\cite{mozaffari2016efficient}&Multiple UAVs&Outdoor&Directional&2 GHz&Maximizing the wireless coverage&Circle packing\\
 		&&&&&of UAVs&theory\\
 		\hline
 		~\cite{kalantari2017backhaul}&Single UAV&Outdoor&Directional&2 GHz&Maximizing the wireless coverage&branch and 
 		\\
 		&&&&&of UAV&bound algorithm\\
 		\hline
 		~\cite{alzenad20173da} &Single UAV&Outdoor&Directional&2 GHz&Maximizing the wireless coverage&exhaustive search
 		\\
 		&&&&&of UAV&algorithm\\
 		\hline
 		~\cite{shakhatreh2017micc}&Multiple UAVs&Indoor&Directional&2 GHz&Maximizing the wireless coverage&efficient algorithm\\
 		&&&&&of UAVs&\\
 		\hline
 		~\cite{shah2017distributed}&Multiple UAVs&Outdoor&Directional&2 GHz&Maximizing the wireless coverage&branch and\\
 		&&&&&of UAVs&bound algorithm\\
 		\hline
 		~\cite{mozaffari2017wireless}&Multiple UAVs&Outdoor&Directional&2 GHz&Maximizing the wireless coverage&Optimal transport \\
 		&&&&&of UAVs&theory, Gradient\\
 		&&&&&&algorithm\\
 		\hline
 		~\cite{kalantari2016number}&Multiple UAVs&Outdoor&Directional&2 GHz&Minimizing the number &Particle swarm\\
 		&&&&&of UAVs&optimization\\
 		\hline
 		~\cite{shakhatreh2016continuous}	&Multiple UAVs&Outdoor&Directional&2 GHz&Minimizing the number & The cycles with limited
 		\\
 		&&&&&of UAVs& energy algorithm\\
 		\hline
 		~\cite{shakhatreh2017indoor}	&Multiple UAVs&Indoor&Omni-directional&2 GHz&Minimizing the number &Particle swarm\\
 		&&&&&of UAVs&optimization, K-means\\
 		&&&&&&algorithms\\
 		\hline
 		~\cite{zhu2014using}	&Multiple UAVs&Outdoor&Directional&2 GHz&Minimizing the number &polynomial time \\
 		&&&&&of UAVs&approximate
 		algorithm\\
 		\hline
 		~\cite{lyu2017placement}	&Multiple UAVs&Outdoor&Directional&2 GHz&Minimizing the number &Spiral UAVs 
 		\\
 		&&&&&of UAVs&placement algorithm\\
 		\hline
 		~\cite{mozaffari2016mobile}	&Multiple UAVs&Outdoor&Directional&2 GHz&Collecting data using&Optimal transport\\
 		&&&&& UAVs&theory\\
 		\hline
 		~\cite{yang2017energy}	&Single UAV&Outdoor&Directional&2 GHz&Collecting data using&Two practical \\
 		&&&&& UAV& UAV trajectories:\\
 		&&&&&&circular and \\
 		&&&&&&straight flights\\
 		\hline
 		~\cite{alejo2015efficient}	&Multiple UAVs&Outdoor&Directional&2 GHz&Collecting data using&Genetics\\
 		&&&&& UAVs&algorithm\\
 		\hline
 		~\cite{wang2015efficient}&Multiple UAVs&Outdoor&directional&2 GHz&Collecting data using& efficient algorithm\\
 		&&&&& UAVs&for path planning\\
 		\hline
 		~\cite{zhan2017energy}&Single UAV&Outdoor&Directional&2 GHz&Collecting data using&efficient iterative  \\
 		&&&&& UAVs&algorithm\\
 		\hline
 	\end{tabular}
 \end{table*}

\subsubsection{{Deployment Strategies for Minimizing the Number of UAVs Required to Perform Task}}
The authors in~\cite{kalantari2016number} propose a
method to find the placements of UAVs in an area with different
user densities using the particle swarm optimization. The goal is to find the minimum number of UAVs and their 3D placements so that all the users are served. In~\cite{shakhatreh2016continuous}, the authors study the problem of minimizing the number of UAVs required for a continuous coverage of a given area, given the recharging requirement. They prove that this problem is NP-complete. Due to its intractability, they study partitioning the coverage graph into cycles that start at the charging station. Based on this analysis, they then develop an efficient algorithm, the cycles with limited energy algorithm, that minimizes the number of UAVs required for a continuous coverage. The authors in~\cite{shakhatreh2017indoor} study the problem of minimizing the number of UAVs required to provide wireless coverage to indoor users and prove that this problem is NP-complete.
Due to the intractability of the problem, they use clustering to minimize the number of UAVs required to cover the indoor users. They assume that each cluster will be covered by only one UAV and apply the particle swarm optimization to find the UAV 3D location and UAV transmit power needed to cover each cluster. In~\cite{zhu2014using}, the authors study the problem of deploying minimum number of UAVs to maintain the connectivity of ground MANETs under the condition that some UAVs have already been deployed in the field. They formulate this problem as a minimum steiner tree problem with existing mobile steiner points under edge length bound constraints and prove that the problem is NP-Complete. They propose an existing UAVs aware polynomial time approximate algorithm to solve the problem that uses a maximum match heuristic to compute new positions for existing UAVs. The authors in~\cite{lyu2017placement} aim to minimize the number of UAVs required to provide wireless coverage for a group of distributed ground terminals, ensuring that each ground terminal is within the communication range of at least one UAV. They propose a polynomial-time algorithm with successive UAV placement, where the UAVs are placed sequentially starting from the area perimeter of the uncovered ground terminals along a spiral path towards the center, until all ground terminals are covered.
\begin{figure}
	\centering
	\includegraphics[scale=0.28]{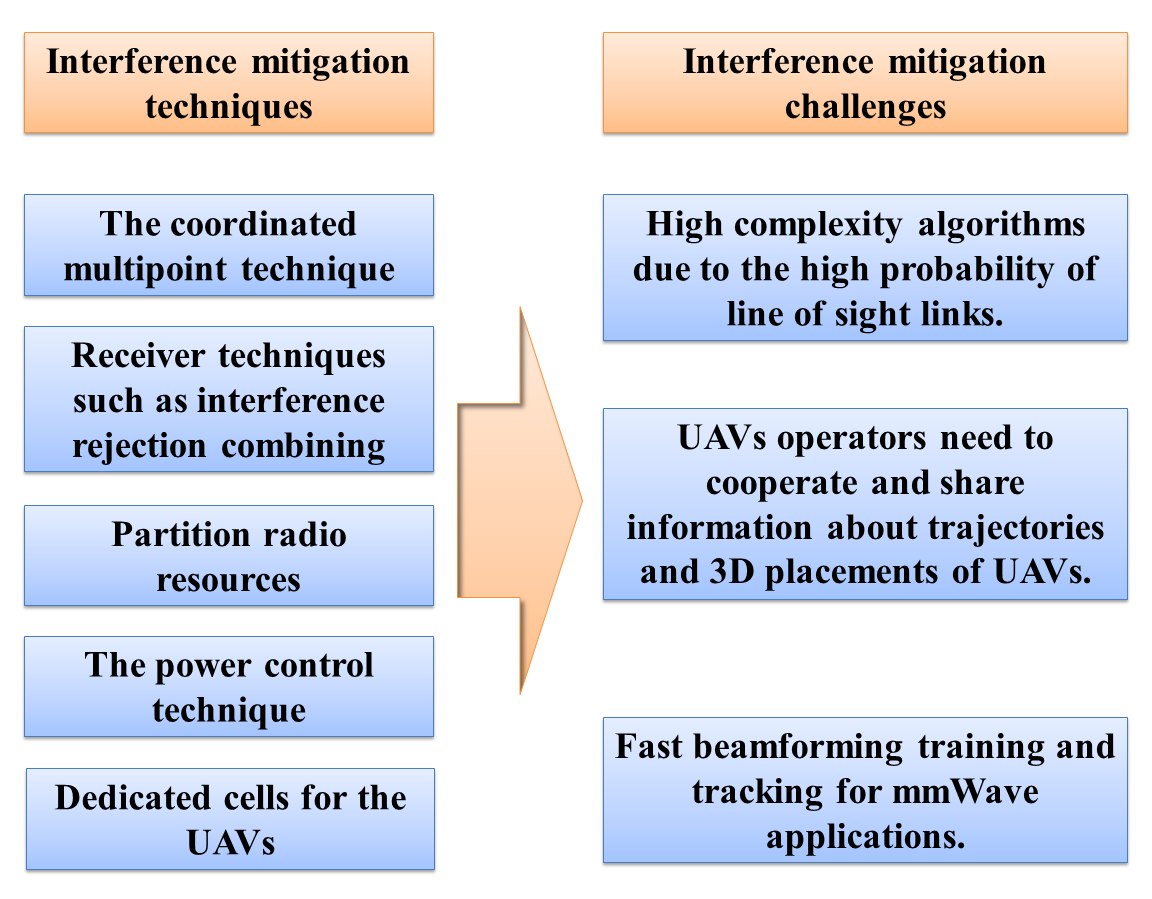}
	\caption{Interference Mitigation Techniques and Challenges.}
	\label{haz8}
\end{figure}
\subsubsection{{Deployment Strategies to Collect Data Using UAVs}}
The authors in~\cite{mozaffari2016mobile} propose an efficient framework for deploying and moving UAVs to collect data from
ground Internet of Things devices. They minimize the total
transmit power of the devices by properly clustering
the devices where each cluster being served by one UAV. The optimal trajectories of the UAVs are determined by exploiting the framework of optimal transport theory. In~\cite{yang2017energy}, the authors present a UAV enabled data collection system, where a UAV is dispatched to collect a given amount of data from ground terminals at fixed location. They aim to find the optimal ground terminal transmit power and UAV trajectory that achieve different Pareto optimal energy trade-offs between the ground terminal and the UAV. The authors in~\cite{alejo2015efficient} study the problem of trajectory planning for wireless sensor network data collecting deployed in remote areas with a cooperative system of UAVs. The missions are given by a set of ground points which define wireless sensor network gathering zones and each UAV should pass through them to gather the data while avoiding passing over forbidden areas and collisions between UAVs. The proposed UAV trajectory planners are based on Genetics Algorithm, Rapidly-exploring Random Trees and Optimal Rapidly-exploring Random Trees. The authors in~\cite{wang2015efficient} design a basic framework for aerial data collection, which includes the following five components: deployment of networks, nodes positioning, anchor points searching, fast path planning for UAV, and data collection from network. They  identify the key challenges in each component and propose efficient solutions. They propose a Fast Path Planning with Rules algorithm based on grid division, to increase the efficiency of path planning, while guaranteeing the length of the path to be relatively short. In~\cite{zhan2017energy}, the authors jointly optimize the sensor nodes wake-up schedule and UAV’s trajectory to minimize the maximum energy consumption of all sensor nodes, while ensuring that the required amount of data is collected reliably from each sensor node. They formulate a mixed-integer non-convex optimization problem and apply the successive convex optimization technique, an efficient iterative algorithm is proposed to find a sub-optimal solution. 

UAVs can be classified into two types: fixed wing and rotary wing, each with its own strengths and weaknesses. Fixed-wing UAV usually has high speed and payload, but they need to maintain a continuous forward motion to remain aloft, thus are not suitable for stationary uses. In contrast, rotary-wing UAV such as quadcopter, usually has limited mobility and payload, but they are able to move in any direction as well as to stay stationary in the air. Thus, the choice of UAVs critically depends on the uses~\cite{zeng2016wireless}. In Table~\ref{tablehaz2}, we make a comparison among the research papers related to UAV deployment strategies based on the objective functions.
\subsection{Interference Mitigation}
One of the techniques to mitigate interference is the coordinated multipoint technique~\cite{maattanen2012system}. In downlink coordinated multipoint technique, the transmission aerial base stations cooperate in scheduling and transmission in order to strength the received signal and mitigate inter-cell interference~\cite{maattanen2012system}. On the other hand, the physical uplink shared channel (PUSCH) is received at aerial base stations in uplink coordinated multipoint technique. The scheduling decision is based on the coordination among UAVs~\cite{sawahashi2010coordinated}. The new challenge here is that UAVs receive interfering signals from more ground terminals in the downlink and their uplink transmitted signals are visible to more ground terminals due to the high probability of line of sight links. Therefore, the coordinated multipoint techniques must be applied across a larger set of cells to mitigate the interference and hence the coordination complexity will increase~\cite{lin2017sky}.

We can also mitigate interference by utilizing receiver techniques such as interference rejection combining and network-assisted interference cancellation and suppression. Compared to smart phones, we can equip UAVs with more antennas, which can be used to mitigate interference from more ground base stations. With MIMO antennas, UAVs can use beamforming to enables directional signal transmission or reception to achieve spatial selectivity which is also an efficient interference mitigation technique~\cite{lin2017sky}.

Another interference mitigation technique is to partition radio resources so that ground traffic and aerial traffic are served with orthogonal radio resources. This simple technique may not be efficient since the reserved radio resources for aerial traffic may be not fully utilized. Therefore, UAV operators need to provide more information about the trajectories and 3D placements of UAVs to construct more dynamic radio resource management~\cite{lin2017sky}.

A powerful interference mitigation technique is the power control technique in which an optimized setting of power control parameters can be applied to reduce the interference generated by UAVs. This technique can minimize interference, increase spectral efficiency, and benefit UAVs as well as ground terminals~\cite{lin2017sky}.

Dedicated cells for the UAVs is another option to mitigate interference, where the directional antennas are pointed towards the sky instead of down-tilted. These dedicated cells will be a practical solution especially in UAV hotspots where frequent and dense UAV takeoffs and landings occur~\cite{lin2017sky}. In Figure~\ref{haz8}, we summarize the interference mitigation techniques and challenges.

\begin{table*}[!h]
	\scriptsize
	\renewcommand{\arraystretch}{1.2}
	\caption{\uppercase{UAV Machine learning algorithms}}
	\label{table4}
	\centering
	\begin{tabular}{|c|c|c|c|}
		\hline 
		Category&Learning techniques&Key characteristics& Application\\ 
		\hline
		&Regression models&- Estimate the variables’ relationships~~~~~&- Energy learning~\cite{donohoo2014context}~~~~~~~~~~~~~~~ \\
		&&- Linear and logistics regression~~~~~~~~~~~&\\
		\hhline{|~|-|-|-|} 
		&K-nearest neighbor&- Majority vote of neighbors~~~~~~~~~~~~~~~&- Energy learning~\cite{donohoo2014context}~~~~~~~~~~~~~~~~\\
		\hhline{|~|-|-|-|} 
		Supervised	&Support vector machines&- Non-linear mapping to high dimension&- MIMO channel learning~\cite{feng2012determination}~~~~~~\\
		learning	&&- Separate hyperplane classification~~~~~~~&\\
		\hhline{|~|-|-|-|} 
		&	Bayesian learning	&- A posteriori distribution calculation~~~~~&-  Massive MIMO learning~\cite{wen2015channel}~~~~~\\	
		&&- Gaussians mixture, expectation max~~~~&- Cognitive spectrum learning~~~~~~\\
		&& 
		and hidden Markov models~~~~~~~~~~~~&~\cite{choi2013estimation,assra2016approach,yu2010cognitive}\\			
		\hline
		Unsupervised	&K-means clustering&- K partition clustering~~~~~~~~~~~~~~~~~~~~~&- Heterogeneous
		networks~\cite{xia2012optical}~~~~~\\
		&&- Iterative updating algorithm~~~~~~~~~~~~~~&\\
		\hhline{|~|-|-|-|} 
		learning	& Independent
		component analysis&- Reveal hidden independent factors~~~~~&- Spectrum learning in~\cite{nguyen2013binary}~~~~~~~	\\
		&&&cognitive radio~~~~~~~~~~~~~~~~~~~\\
		\hline
		&Markov decision processes/A partially&-  Bellman equation
		maximization~~~~~~~~&- Energy harvesting~\cite{aprem2013transmit}~~~~~~~~~~~~\\
		&
		observable Markov decision process&- Value iteration algorithm~~~~~~~~~~~~~~~~&\\
		\hhline{|~|-|-|-|}
		Reinforcement&Q-learning&- Unknown system
		transition model~~~~~&- Femto and small cells~~~~~~~~~~~~ \\
		learning&&- Q-function maximization~~~~~~~~~~~~~~~&~\cite{alnwaimi2015dynamic}-\cite{onireti2016cell}\\
		
		\hhline{|~|-|-|-|}
		&Multi-armed bandit&- Exploration vs. exploitation~~~~~~~~~~~~& - Device-to-device networks~\cite{maghsudi2015channel} ~\\	
		&&- Multi-armed bandit game~~~~~~~~~~~~~~&\\														
		\hline 
	\end{tabular}
\end{table*}

\subsection{Research Trends and Future Insights}
\subsubsection{{Cloud and Big Data}}
A cloud for UAVs contains data storage and high-performance computing, combined with big data analysis tools~\cite{bor2016new}. It can provide an economic and efficient use of centralized resources for decision making and network-wide monitoring~\cite{bor2016new,bradai2015cellular,zhou2014toward}. If UAVs are utilized by a traditional cellular network operator (CNO), the cloud is just the data center of the CNO (similar to a private cloud), where the CNO can choose to share its knowledge with some other CNOs or utilize it for its own business uses. On the other hand, if the UAVs are utilized by an infrastructure provider, the infrastructure provider can utilize the cloud to gather information from mobile virtual network operators and service providers. Under such scenario, it is important to guarantee security, privacy, and latency. To better exploit the benefit of the cloud, we can use a programmable network allowing dynamic updates based on big data processing, for which network functions virtualization and software defined networking can be research trends ~\cite{bor2016new}.
\subsubsection{{Machine Learning}}
Next-generation aerial base stations are expected to learn the diverse characteristics of users’ behavior, in order to autonomously find the optimal system settings. These intelligent UAVs have to use sophisticated learning and decision-making, one promising solution is to utilize machine learning. Machine learning algorithms can be simply classified as supervised, unsupervised and reinforcement learning as shown in Table~\ref{table4}. The family of supervised learning algorithms utilizes known models and labels to enable the estimation of unknown parameters. They can be used for spectrum sensing and white space detection in cognitive radio, massive MIMO channel estimation and data detection, as well as for adaptive filtering in signal processing for 5G communications. They can also be utilized in higher-layer applications, such as estimating the mobile users’ locations and behaviors, which can help the UAV operators to enhance the quality of their services. The family of unsupervised learning algorithms utilizes the input data itself in a heuristic manner. They can be used for heterogeneous base station clustering in wireless networks, for access point association in ubiquitous WiFi networks, for cell clustering in cooperative ultra-dense small-cell networks, and for load-balancing in heterogeneous networks. They can also be utilized in fault/intrusion detections and for the users’ behavior-classification. The family of reinforcement learning algorithms utilizes dynamic iterative learning and decision-making process. They can be used for estimating the mobile users’ decision making under unknown scenarios, such as channel access under unknown channel availability conditions in spectrum sharing, base station association under the unknown energy status of the base stations in energy harvesting networks, and distributed resource allocation under unknown resource quality conditions in femto/small-cell networks~\cite{jiang2017machine}.
\subsubsection{{Network Functions Virtualization}}
Network functions virtualization (NFV) reduces the need of deploying specific network devices for the integration of UAVs~\cite{bor2016new,bradai2015cellular}. NFV allows a programmable network structure by virtualizing the network functions on storage devices, servers, and switches, which is practical for UAVs requiring seamless integration to the existing network. Moreover, virtualization of UAVs as shared resources among cellular virtual network operators can decrease OPEX for each party~\cite{liang2015wireless}. Here, the SDN can be useful for the complicated control and interconnection of virtual network functions (VNFs)~\cite{bor2016new,bradai2015cellular}.
\subsubsection{{Software Defined Networking}} 
For mobile networks, a centralized SDN controller can make a more efficient allocation of radio resources, which is particularly important to exploit UAVs~\cite{bor2016new,bradai2015cellular}. For instance, SDN-based load balancing can be useful for multi-tier UAV networks, such that the load of each aerial base station and terrestrial base station is optimized precisely. A SDN controller can also update routing such that part of traffic from the UAVs is carried through the network without any network congestions~\cite{bor2016new,bradai2015cellular,zhou2014toward}. For further exploitation of the new degree of freedom provided by the mobility of UAVs, the 3D placements of UAVs can be adjusted to optimize paging and polling, and location management parameters can be updated dynamically via the unified protocols of SDN~\cite{bor2016new}.
\subsubsection{{Millimeter-Wave}} 
Millimeter-wave (mmWave) technology can be utilized to provide high data rate wireless communications for UAV networks~\cite{xiao2016enabling}. The main difference between utilizing mmWave in UAV aerial networks and utilizing mmWave in terrestrial cellular networks is that a UAV aerial base station may move around. Hence, the challenges of mmWave in terrestrial cellular networks apply to the mmWave UAV cellular network as well, including rapid channel variation, blockage, range and directional communications, multi-user access, and others~\cite{xiao2016enabling,rangan2014millimeter}. Compared to mmWave communications for static stations, the time constraint for beamforming training is more critical due to UAV mobility. For fast beamforming training and tracking in mmWave UAV cellular networks, the hierarchical beamforming codebook is able to create highly directional beam patterns, and achieves excellent beam detection performance~\cite{xiao2016enabling}. Although the UAV wireless channels have high Doppler frequencies due to the continuous movements of UAVs with different velocities, the major multipath components are only affected by slow variations due to high gain directional transmissions. In beam division multiple access (BDMA), multiple users with different beams may access the channel at the same time due to the highly directional transmissions of mmWave~\cite{xiao2016enabling,rangan2014millimeter,sun2015beam}. This technique improves the capacity significantly, due to the large bandwidth of mmWave technology and the use of BDMA in the spatial domain. The blockage problem can be mitigated by utilizing intelligent cruising algorithms that enable UAVs to fly out of a blockage zone and enhance the probability of line of sight links~\cite{xiao2016enabling}. 
\subsubsection{{Free Space Optical}} 
Free space optical (FSO) technology can be used to provide wireless connectivity to remote places by utilizing UAVs, where physical access to 3G or 4G network is either minimal or never present~\cite{kaushal2017optical}. It can be involved in the integration of ground and aerial networks with the help of UAVs by providing last mile wireless coverage to sensitive areas (e.g., battlefields, disaster relief, etc.) where high bandwidth and accessibility are required. For instance, Facebook will provide wireless connectivity via FSO links to remote areas by utilizing solar-powered high altitude UAVs~\cite{kaushal2017optical,adweek,mobileeurope}. For areas where deployment of UAVs is impractical or uneconomical, geostationary earth orbit and low earth orbit satellites can be utilized to provide wireless connectivity to the ground users using the FSO links. The main challenge facing UAV-free space optics technology is the high blockage probability of the vertical FSO link due to weather conditions. The authors in~\cite{alzenad2016fso} propose some methods to tackle this problem. The first method is to use an adaptive algorithm that controls the transmit power
according to weather conditions and it may also adjust other system parameters such as incident angle of the FSO transceiver to compensate the link degradation, e.g., under bad weather conditions, high power vertical FSO beams should be used while low power beams could be used under good weather conditions. The second method is to use a system optimization algorithm that can optimize the UAV placement, e.g., hovering below clouds over negligible turbulence geographical area.

\subsubsection{Future Insights} 
Some of the future possible directions for this research are: 

\begin{itemize}
	\item The majority of literature focuses on providing the wireless coverage for outdoor ground users, although 90\% of the time  people are indoor and 80\% of the mobile Internet access traffic
	also happens indoors~\cite{ericsson,alcatel,cisco}. Therefore, it is important to study the indoor wireless coverage problems by utilizing UAVs.
	
	\item The majority of literature does not consider the limited energy capacity of UAV as a constraint when they study the wireless coverage of UAVs,  where the energy consumption during data transmission and reception is much smaller than the energy consumption during the UAV hovering. It only constitutes 10\%-20\% of the UAV energy capacity~\cite{gupta2016survey}.
	
	\item Some path loss models are based on simulations softwares such as Air-to-Ground path loss for low altitude platforms and Air-to-Ground path loss for high altitude platforms, therefore it is necessary to do real experiments to model the statistical behavior of the path loss.
	\item To the best of our knowledge, no studies present the path loss models for uplink scenario and mmWave bands.
	\item There are challenges facing the new technologies such as mmwave. The challenges facing UAV-mmWave technology are fast beamforming training and tracking requirement, rapid channel variation, directional and range communications, blockage and multi-user access~\cite{xiao2016enabling}. 
	\item We need more studies about the topology of UAV wireless networks where this topology remains fluid during: a) Changing the number of UAVs; b) Changing the number of channels; c) The relative placements of the UAVs altering~\cite{gupta2016survey}.
	\item There is a need to study UAV routing protocols where it is difficult to construct a simple implementation for proactive or reactive schemes~\cite{gupta2016survey}.
	\item There is a need for a seamless mechanism to transfer the users during the handoff process between UAVs, when out of service UAVs are replaced by active ones~\cite{gupta2016survey}.
	\item Further research is required to determine which technology is right for UAV applications, where the common technologies utilized in UAV applications are the IEEE 802.11 due to wide availability in current wireless devices and their appropriateness for small-scale UAVs~\cite{hayat2016survey}.
	\item D2D communications is an efficient technique to improve the capacity in ground communications systems~\cite{asadi2014survey}. An important problem for future research is the joint optimization of the UAV path planning, coding, node clustering, as well as D2D file sharing~\cite{zeng2016wireless}.
	\item Utilizing UAVs in public safety communications needs further research, where topology formation, cooperation between UAVs in a multi-UAV network, energy constraints, and  mobility model are the challenges that are facing UAVs in public safety communications~\cite{kumbhar2017survey}. 
	\item In UAVs-based IoT services, it is difficult 
	to control and manage a high number of UAVs. The reason
	is that each UAV may host more than one IoT device, such as different types of cameras and aerial sensors. Moreover, conflict of interest between different devices likely to happen many times, e.g., taking two videos
	from two different angles from one fixed placement. For future research, we need to propose efficient algorithms to
	solve the conflict of interest among Internet of things devices on-board~\cite{motlagh2016low}.
	\item For future research, it is important to propose efficient techniques that manage and control the power consumption of Internet of things devices on-board~\cite{motlagh2016low}. 
	\item The security is one of the most critical issues to think about in UAVs-based Internet of things services, methods to avoid the aerial jammer on the communications are needed~\cite{motlagh2016low}.
\end{itemize}

\section*{PART III: Key Challenges and Conclusion}

\section{Key Challenges}

\subsection{Charging Challenges}
UAV missions necessitate an effective energy management for battery-powered UAVs. Reliable, continuous, and smart management can help UAVs to achieve their missions and prevent loss and damage. The UAV's battery capacity is a key factor for enabling persistent missions. But as the battery capacity increases, its weight increases, which cause the UAV to consume more energy for a given mission. The main directions in the literature to mitigate the limitations in UAV's batteries are: (1) UAV battery management, (2) Wireless charging for UAVs, (3) Solar powered UAVs, and (4) Machine learning and communications techniques.  

\subsubsection{Battery Management}
\begin{figure}[h]
	\centering
	
	\includegraphics[width=6cm, height=7cm]{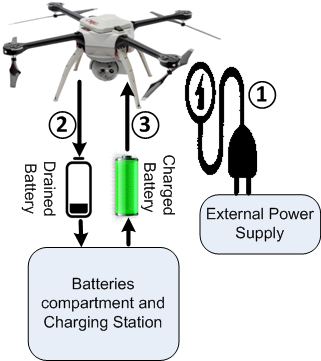}
	\caption{Illustration of Battery Hot Swapping System.}
	\label{eyad4}
\end{figure}

Battery management research in UAVs includes planning, scheduling, and replacement of battery so UAVs can accomplish their flight missions.  This has been studied in the literature from different perspectives. Saha et al. in \cite{Saha2011} build a model to predict the end of battery charge for UAVs based on particle filter algorithm. They utilize a discharge curve for UAV Li-Po battery to tune the particle filter. They have shown that the depletion of the battery is not only related to the initial Start of Charge (SOC) but also, load profile and battery health conditions can be crucial factors. Park et al. in \cite{Park2017}  investigate the battery assignment and scheduling for UAVs used in delivery business. Their objective is to minimize the deterioration in battery health. After splitting the problem into two parts; one for battery assignment and the other for battery scheduling. Heuristic algorithm and integer linear programming are used to solve the assignment and scheduling problems, respectively. The idle time between two successive charging cycles and the depth of the discharge are the main factors that control these algorithms. The use of UAVs in long time and enduring missions makes UAV battery swapping solutions necessary to accomplish these missions. An autonomous battery swapping system for UAVs is first introduced in \cite{Swieringa2010}. The battery swapping consists of landing platform, battery charger, battery storage compartment, and micro-controller. Similar concept to this autonomous battery swapping system was adopted and improved by different researchers in \cite{Michini2011,Suzuki2012,Lee2015,Ure2015}. The hot swap term is adopted to refer to the continuous powering for the UAV during battery swapping. Figure~\ref{eyad4} shows an illustration of hot swapping systems. First, the UAV is connected to an external power supply during the swapping process. This will prevent data loss during swapping as it usually happens in cold swapping. Second, the drained battery is removed and stored in multi-battery compartment and charging station to be recharged. Third, a charged battery is installed in the UAV, and finally, the external power supply is disconnected. Another important aspect of the improvements presented in \cite{Michini2011,Suzuki2012,Lee2015,Ure2015} is the several designs for the landing platform, which is an important part of the swapping system because it can compensate the error in the UAV positioning on the landing point.  Table~\ref{eyadt3} shows the classification of autonomous battery swapping systems. It can be seen from this table that the swapping time for most of these systems is around one minute. This is a short period comparing to the battery charging average time, which is between 45-60 minutes.

	\begin{table*}[!t]
	\footnotesize
	\renewcommand{\arraystretch}{1.3}
	\caption{\uppercase{classification of autonomous battery swapping systems}}
	\label{eyadt3}
	\centering
	\begin{tabular}{|l|c|l|c|l|c|}
		\hline
			\hline
		\textbf{ Ref \# /Year} & \textbf{UAV Type} & \textbf{Battery Type} & \textbf{{Positioning System}}& \textbf{Swapping time}&\textbf{Hot vs Cold Swapping}\\
		\hline
			\hline
		 {\cite{Swieringa2010}/2010} &{Quadrotor}&	{Li-Po}&	{V-Shape Basket}&	{2 min.}	& {Cold swap}\\ 		\hline
	{\cite{Michini2011}/2011} &{Quadrotor}&	{Li-Po}& {Sloped landing pad}&	{11.8±3 sec.}&	{Hot swaps}\\
			\hline 
		 {\cite{Suzuki2012}/2012} &{Quadrotor}&	{Li-Po}	& {Dynamic forced positioning (Arm Actuation)}&	{47-60 sec.}&	{Cold swap} \\	
			\hline 
			{\cite{Ure2015}/2015} &{Quadrotor}&	{Li-Po} &	{Sloped landing pad}&	{60-70 sec.}&	{Hot swaps}\\
				\hline
				{\cite{Lee2015}/2015} &{Quadrotor}&	{Li-Po} &	{Rack with landing guide}&	{60 sec.}&	{Hot swaps}\\
					\hline
						\hline
			
	\end{tabular}
\end{table*}

\subsubsection{Wireless Charging For UAV}
Simic et. Al in \cite{Simic2015} study the feasibility of recharging the UAV from power lines while inspecting these power lines. Experimental tests show the possibility of energy harvesting from power lines. The authors design a circular antenna to harvest the required energy from power lines. Wang et al. in \cite{Wang2016} propose an automatic charging system for the UAV. The system uses charging stations allocated along the path of UAV mission. Each charging station consists of wireless charging pad, solar panel, battery, and power converter. All these components are mounted on a pole. The UAV employs GPS module and wireless network module to navigate to the autonomous charging station. When the power level in the UAV's battery drops to a predefined level. The UAV will communicate with the central control room, which will direct the UAV to the closest charging station. The GPS module will help the UAV to navigate to the assigned station. Different wireless power transfer technologies were used in literature to implement the automatic charging stations. In \cite{Junaid2016,Choi2017,Aldhaher2017}, the authors utilize the magnetic resonance coupling technology to implement an automatic drone charging station. Dunbar et al. in \cite{Dunbar2015} use RF Far field Wireless Power Transfer (WPT) to recharge micro UAV after landing. A flexible rectenna was mounted on the body of the UAV to receive RF signals from power transmitter. Mostafa et al. in \cite{Mostafa2017}  develop a WPT  charging system based on the capacitive coupling technology. One of the major challenges that face the deployment of autonomous wireless charging stations is the precise landing of the UAV on the charging pad. The imprecise landing will lead to imperfect alignment between the power transmitter and the power receiver, which means less efficiency and long charging time. Table~\ref{eyadt4} shows precise methodologies adopted in the literature. In \cite{Wang2016}, the authors use GPS to land the UAV on the charging pad. In \cite{Junaid2017}, the authors corporate the GPS system with camera and image processing system to detect the charging station. The authors in \cite{Choi2017} allow for imprecise landing on wide frame landing pad. Then a wireless charging transmitting coil is stationed perfectly in the bottom of the UAV by using a stepper motor and two laser distance sensors.
\begin{table}[h!]
	\centering
	\caption{\uppercase{classification of Precise Landing Methodologies}}
	\label{eyadt4}
	\begin{tabular}{ |c|c|} 
		\hline
		{Ref \# /Year} &   {Precise Landing Methodology} \\
		\hline
		{\cite{Wang2016}/2016} &{GPS system}\\ 
		{\cite{Choi2017}/2017} &{Landing frame with moving transmitting coil}\\ 
		{\cite{Junaid2017}/2017} &{GPS system and vision-based target detection} \\ 
		\hline
	\end{tabular}
\end{table}

 \subsubsection{Solar Powered UAVs}   
For long endurance and high altitude flights, solar-powered UAVs (SUAVs) can be a great choice. These UAVs use solar power as a primary source for propulsion and battery as a secondary source to be used during night and sun absence conditions. The literature in flight endurance and persistent missions for SUAVs can be classified into two main directions: path planning and optimization  and hybrid Models.  Path planning and optimization includes: (i) gravitational potential energy planning. This scheme optimizes the path of the UAV over three stages, ascending and battery charging stage during the day using solar power, descending stage during the night using gravitational potential energy, and level flight stage using battery \cite{Hosseini2016,Lee2017,Gao2013}, (ii) optimal path planing while tracking ground object \cite{Huang2016,Spangelo2013}, (iii) optimal path planning based on UAV kinematics and its energy loss model \cite{Klesh2007}, and (iv) optimal path planning utilizing wind and metrological data \cite{Chakrabarty2011,Oettershagen2017}.  On the other hand, Hybrid models include: (i) UAVs with hybrid power sources such as solar, battery, and fuel cells \cite{Lee2014} and (ii) UAVs with a hybrid propulsion system that can be transformed from fixed wing to quadcopter UAV \cite{DSa2016,DSa2017}.   

\subsubsection{Machine Learning and Communications Techniques}
Communication equipment and their status (transmit, receive, sleep, or idle) can affect drone flight time. Utilizing the state-of-the-art machine learning techniques can result in smarter energy management. These technologies and their role in UAV persistent  missions are covered in the literature under the following themes: 
\begin{itemize}
	\item Energy efficient UAV networks: This area covers maximizing energy efficiency in the network layer, data link layer, physical layer, and cross-layer protocols in a bid to help UAVs to perform long missions. Gupta et al. in \cite{Gupta2016} list and compare several algorithms used in each layer.   
	\item Energy-based UAV fleet management: This includes modifying the mobility model of UAV fleet to incorporate energy as decision criterion as in \cite{Messous2016}  to determine the next move of each UAV inside a fleet.  
	\item Machine learning: This area covers research that utilizes machine learning techniques for path planning and optimization applications, taking into consideration energy limitations. Zhang et al. in  \cite{Zhang2017} use deep reinforcement learning to determine the fastest path to a charging station. While Choi et al. in  \cite{Choi2017a} use a density-based spatial clustering algorithm to build a two-layer obstacle collision avoidance system.
\end{itemize}


Several challenges still exist in the area of battery recharging and need to be appropriately addressed, including:
\begin{itemize}
	\item Advancements in wireless charging techniques such as magnetic resonant coupling and inductive coupling techniques are popular and suitable to be used because both of them have acceptable efficiency for small to mid-range distances. The advancements in this technology are expected to have a positive impact on the use of UAVs. 
	\item Development of battery technologies will allow UAVs to extend flight ranges. While Lithium-Ion batteries still dominant, PEM Fuel cells can be more convenient in UAVs because of their higher power density.
	\item Motion planning of UAVs is an extremely important factor since it affects the wireless charging efficiency and the average power delivered .
	\item The vast majority of the research, concerning UAV power management, focuses on multi-rotor UAVs. Other types of UAVs, such as fixed-wing UAVs, need more attention in future research. 
	\item Artificial intelligence, especially deep learning, can promote more advances in the field of UAV power management. Many research frontiers still need to be explored in this field such as; Deep reinforcement learning in the path planning and battery scheduling, convolutional neural networks in identifying charge stations and precise landing in a charging station, and recurrent neural networks in developing discharge models and precisely predicting the end of the charge. 
	\item Image processing techniques and smart sensors can also play an important role in identifying charging stations and facilitating precise positioning of UAVs on them.
\end{itemize}

\subsection{Collision Avoidance and Swarming Challenges}

One of the challenges facing UAVs is collision avoidance. UAV can collide with obstacles, which can either be moving or stationary objects in either indoor or outdoor environment. During UAV flights, it is important to avoid accidents with these obstacles. Therefore, the development of UAV collision avoidance techniques has gained research interest \cite{pham2015survey,lacher2007unmanned}. In this section, we first present the major categories of collision avoidance methods. Then, the challenges associated with UAVs collision avoidance are presented. Moreover, we provide research trends and some future insights.

\subsubsection{Collision Avoidance Approaches}	
Many collision avoidance approaches have been proposed in order to avoid potential collisions by UAVs.
The authors in \cite{pham2015survey} presented major categories of collision avoidance methods. 
These methods can be summarized as geometric approaches \cite{park2008uav}, path planning approaches \cite{alexopoulos2013comparative}, potential field approaches \cite{mujumdar2009nonlinear,khatib1986real}, and vision-based approaches \cite{saunders2009obstacle}. Several collision avoidance techniques can be utilized for an indoor environment such as vision based methods (using cameras and optical sensors), on-board sensors based methods (using IR, ultrasonic, laser scanner, etc.) and vision based combined with sensor based methods~\cite{ luo2013uav}. The collision avoidance approaches are discussed in the following:

\begin{itemize}
	\item Geometric Approach is a method that utilizes 
	the geometric analysis to avoid the collision. In~\cite{park2008uav}, the geometric approach was utilized to ensure that the predefined minimum separation distance was not violated. This was done by computing the distance between two UAVs and the time required for the collision to occur. Moreover, it was used for path planning to avoid UAV collisions with obstacles. 
	Several different approaches utilize this method to avoid collision such as Point of Closest Approach \cite{park2008uav}, Collision Cone Approach \cite{chakravarthy1998obstacle,mujumdar2009nonlinear}, and Dubins Paths Approach \cite{dubins1957curves,shanmugavel2010co}.
	
	\item Path Planning Approach, which is also referred to as the optimized trajectory approach \cite{jun2003path}, is a grid based method that utilizes the path re-planning algorithm with graph search algorithm to find a collision free trajectory during the flight. It uses geometric techniques to find an efficient and collision free trajectory, so it shares some similarities with the geometric approach. This method divides the map into a grid and represents the grid as a weighted graph \cite{alexopoulos2013comparative}. The grid with graph search algorithm are used to find collision free trajectory towards the desired target.
	

	\item Potential Field Approach is proposed by \cite{khatib1986real} to be used as a collision avoidance method for ground robots. It has also been utilized for collision avoidance among UAVs and obstacles. This method uses the repulsive force fields which cause the UAV to be repelled by obstacles. The potential function is divided into attracting force field which pulls the UAV towards the goal, and repulsive force field which is assigned with the obstacle\cite{mujumdar2009nonlinear}. 
	
	\item Vision-based obstacle detection approaches utilize images from small cameras mounted on UAVs to tackle collision challenges. Advances in integrated circuits technology have enabled the design of small and low power sensors and cameras. Combined with advances in computer vision methods, such camera can be used in effective obstacle detection and collision avoidance \cite{saunders2009obstacle}. Moreover, this approach can be used efficiently for collision avoidance in the indoor environment. Such cameras provide real-time information about walls and obstacles in this environment. Many researchers utilized this method to avoid a collision and to provide fully autonomous UAV flights \cite{ schmid2013stereo, alvarez2016collision, schmid2014autonomous, mustafah2012indoor }. The researchers in \cite{alvarez2016collision} use a monocular camera with forward facing to generate collision-free trajectory. In this method, all computations were performed off-board the UAV. To solve the off-board image processing problem, the authors in \cite{roelofsen2015reciprocal} proposed a collision avoidance system for UAVs with visual-based detection. The image processing operation was performed using two cameras and a small on-board computation unit.
	
\end{itemize}

\subsubsection{Challenges}
There are several challenges in the area of collision avoidance approaches and need to be appropriately addressed, including:

\begin{itemize}
	\item The geometric approach utilizes the information such as location and velocity, which can be obtained using Automatic Dependent Surveillance Broadcast (ADS-B) sensing method. Thus, it is not applicable to non-aircraft obstacles. Furthermore, input data from ADS-B is sensitive to noise which hinders the exact calculation requirement of this approach. Moreover, ADS-B requires cooperation from another aircraft, which can be an intruder, which is referred to as cooperative sensing. 
	
	\item In non-cooperative sensing, the geometric approach requires UAVs to sense and extract information about the environment and obstacles such as position, speed and size of the obstacles. One of the possible solutions is to combine with a vision-based approach that uses a passive device to detect obstacles. However, this approach requires significant data processing and only can be used when the objects are close enough. Therefore, hardware limitation for on-board processing should be taken into consideration.
	
	\item The diverse robotic control problem that allows UAV to perform complex maneuvers without collision can be solved using the standard control theory. However, each solution is limited to a specific case and not able to adapt to changes in the environments. This limitation can be overcome by learning from experience. This approach can be achieved using deep learning technique. More specifically, deep learning allows inferring complex behaviors from raw observation data. However, this approach has issues with samples efficiency. For real time application, deep learning requires the usage of on-board GPUs. In \cite{carrio2017review}, a review of deep learning methods and applications for UAVs were presented.
	
	\item In the context of multi-UAV systems, the collision avoidance among UAVs is a very complicated task, a technique known as formation control can be used for obstacle avoidance. More specifically, cooperative formation control algorithms were developed as a collision-avoidance strategy for the multiple UAVs \cite{6858777,7331006}. A recent advancement in this area employs Model Predictive Controllers (MPC) in order to reduce computation time for the optimization of UAV's trajectory.
	
	\item The limited available power and payload of UAVs are challenging issues, which restrict on-board sensors requirements such as sensor weight, size and  required power. These sensors  such  as  IR, Ultrasonic, laser scanner, LADAR  and  RADAR  are  typically  heavy and large for sUAV \cite{griffiths2007obstacle}.

	\item The high speed of UAVs is another challenge. With speed ranges between 35 and 70 Km/h, obstacle avoidance approach must be executed quickly to avoid the collision \cite{griffiths2007obstacle}.
	
	

	\item The use of UAVs in an indoor environment is a challenging task and it needs higher requirements than outdoor environment. It is very difficult to use GPS for avoiding collisions in an indoor environment, \textit{usually indoor is a GPS denied environment}. Moreover, RF signals cannot be used in this environment, RF signals could be reflected, and degraded by the indoor obstacles and walls. 
	
	\item Vision-based collision avoidance methods using cameras suffer from heavy computational operations for image processing. Moreover, these cameras and optical sensors are light sensitive 
	(need sufficient lighting to work properly), so steam and smoke could cause the collision avoidance system to fail. Therefore, sensor based collision avoidance methods can be used to tackle these problems \cite{chee2013control,gageik2012obstacle}. 
	
	\item Some of the vision based collision avoidance approaches use a monocular camera with forward facing to generate collision free trajectory. In this method, all computations are performed off-board, with 30 frames per second which are sent in real time over Wi-Fi network to perform image processing at a remote base station. This makes UAV sensing and avoiding obstacles a challenging task \cite{alvarez2016collision}

\end{itemize}

\subsubsection{{Future Insights}}
Based on the reviewed literature of collision avoidance challenges, we suggest these possible future directions:
\begin{itemize}
	
	\item  UAVs could be integrated with vision sensors, laser range finder sensors, IR, and/or ultrasound sensors to avoid collisions in all directions \cite{chee2013control}.  
	\item  UAV control algorithms can be developed for autonomous hovering without collision, and to ensure the completion of their mission successfully.
	\item Standardization rules and regulations for UAVs are highly needed  around the world to regulate their operations, to reduce the likelihood of collision among UAVs, and to guarantee safe hovering \cite{clarke2014regulation,archick2005european}.
	
	\item Under the deep learning techniques with Model Predictive Controllers (MPC), further work can be done on developing methods to generate guiding samples for more superior obstacle avoidance. The main concern in the deploying deep learning models is the processing requirement. Therefore, hardware limitation for on-board processing should be taken into consideration.  
	
	\item More studies are required to improve indoor and outdoor collision avoidance algorithms in order to compute smoother collision free paths, and to evaluate optimized trajectory in terms of aspects such as energy consumption \cite{roelofsen2015reciprocal}.
	\item On-board processing is required for many UAV operations, such as dynamic sense and avoid algorithms, path re-planning algorithms and image processing. Design on-board powerful processor devices with low power consumption is an active area for future researches \cite{carrio2017review}. 
	
\end{itemize}

\subsection{Networking Challenges}
\label{FANET}

Fluid topology and rapid changes of links among UAVs are the main characteristics of multi-UAV networks or FANET \cite{gupta2016survey}. A UAV in FANET is a node flying in the sky with 3D mobility and speed between 30 to 460 km/h. This high speed causes a rapid change in the link quality between a UAV and other nodes, which introduces different design challenges to the communications protocols. Therefore, FANET needs new communication protocols to fulfill the communication requirements of Multi-UAV systems \cite{tareque2015routing}.

\subsubsection{FANET Challenges}
\begin{itemize}

	\item One of the challenging issues in FANET is to provide wireless communications for UAVs, (wireless communications between UAV-GCS, UAV-satellite, and UAV-UAV). More specifically, coordination, cooperation, routing and communication protocols for the UAVs are challenging tasks due to the frequent connections interruption, fluid network topology, and limited energy resources of UAVs. Therefore, FANET requires some special hardware and a new networking model to address these challenges
	\cite{sahingoz2014networking,motlagh2016low}.
	
	
	\item The UAV power constraints limit the computation, communication, and endurance capabilities of UAVs. Energy-aware deployment of UAVs with Low power and Lossy Networks (LLT) approach can be efficiently used to handle this challenge \cite{zeng2016wireless,winter2012rpl}.  
	
	\item FANET is a challenging environment for resources management in view of the special characteristics of this network. The challenge is the complexity of network management, such as the configuration difficulties of the hardware components of this network \cite{kirichek2016model}. Software-Defined Networking (SDN) and Network Function Virtualization (NFV) are useful approaches to tackle this challenge \cite{white2017programmable}.

	\item Setting up an ad-hoc network among nodes of UAVs is a challenging task for FANET, due to the high node mobility, large distance between nodes, fluid network topology, link delays and high channel error rates. Therefore, transmitted data across these channels could be lost or delayed. Here the loss is usually caused by disconnections and rapid changes in links among UAVs. Delay-Tolerant Networking (DTN) architecture was introduced to be used in FANET to address this challenge \cite{burleigh2003delay, loo2016mobile }.

\end{itemize}


\subsubsection{UAV New Networking Trends}

\subsubsection*{$a$) Delay-Tolerant Networking (DTN)}
The DTN architecture was designed to handle challenges facing the dynamic environments as in FANET. 
\cite{warthman2012delay}.
In FANET, DTN approach based on store-carry-forward model can be utilized to tackle the long delay for packets delivery. In this model, a UAV can store, carry and forward messages from source to destination with long term data storage and forwarding functions in order to compensate intermittent connectivity of links \cite{fall2007delay,le2006uav}.

DTN can be used with a set of protocols operating at MAC, transport and application layers to provide reliable data transport functions, such as Bundle Protocol (BP) \cite{scott2007bundle}, Licklider Transmission Protocol (LTP) \cite{burleigh2008licklider} and Consultative Committee for Space Data Systems (CCSD) File Delivery Protocol (CFDP) \cite{protocol2007part}. These protocols use store, carry and forward model, so UAV node keeps a copy for each sent packet until it receives acknowledgment from the next node to confirm that the packet has been received successfully. 

BP and LTP are protocols developed to cope with FANET challenges and solve the performance problems for FANET. 
The BP, forms a store, carry and forward overlay networks, to handle message transmissions, receptions and re-transmissions using a convergence layer protocol (CLP) services \cite{scott2007bundle, yang2014analytical} with the underlying transport protocols such as TCP-based \cite{demmer2014delay}, UDP-based \cite{kruseudp} or LTP-based \cite{burleigh2008licklider}.
The CFDP was designed for file transfer from source to destination based on store, carry and forward approach for DTN paradigm. It can be run over either reliable or unreliable service mode using transport layer protocols (TCP or UDP) \cite{wang2009protocols}. 

A routing strategy that combines DTN routing protocols in the sky and the existing Ad-hoc On-demand Distance Vector (AODV) on the ground for FANETs was proposed in \cite{le2006uav}. In this work, they implement a DTN routing protocol on the top of traditional and unmodified AODV. They also use UAV to store, carry and forward the messages from source to destination.

\subsubsection*{$b$) Network Function Virtualization (VFV)}
NFV is a new networking architecture concept for network softwarization and it is used to enable the networking infrastructure to be virtualized. More specifically, an important change in the network functions provisioning approach has been introduced using NFV, by leveraging existing IT virtualization technologies, therefore, network devices, hardware and underlying functions can be replaced with virtual appliances. NFV can provide programming capabilities in FANETs and reduces the network management complexity challenge  \cite{han2015network,jain2013network}.

The research in  \cite{bor2016new}, proposed drone-cell  management  framework (DMF) using UAVs act as aerial base stations with a multi-tier drone-cell network to complement the terrestrial cellular network based on NFV and SDN paradigms.
In \cite{rametta2017designing}, the authors proposed a video monitoring platform as a service (VMPaaS) using swarm of UAVs that form a FANET in rural areas. This platform utilizes the recent NFV and SDN paradigms.

Due to the complexity of the interconnections and controls in NFV, SDN can be consolidated with NFV as a useful approach to address this challenge.

\subsubsection*{$c$) Software-Defined Networking (SDN)}

SDN is a promising network architecture, which provides a separation between control plane and data plane. It can also provide a centralized network programmability with global view to control network. Benefiting from the centralized controller in SDN, the original network switches could be replaced by uniform SDN switches \cite{zhao2014leveraging}. Therefore, the deployment and management of new applications and services become much easier. Moreover, the network management, programmability and reconfiguration can be greatly simplified \cite{zhang2017software}.

FANET can utilize SDN to address its environment's challenges and performance issues, such as dynamic and rapid topology changes, link intermittent between nodes; when UAV goes out of service due to coverage problems or for battery recharging. SDN also can help to address the complexity of network management\cite{gupta2016survey}.

OpenFlow is one of the most common SDN protocols, used to implement SDN architecture and it separates the network control and data planes functionalities.
OpenFlow switch consists of flow Table, OpenFlow Protocol and a Secure Channel \cite{mckeown2008openflow}. UAVs in FANETs can carry OpenFlow Switches. The SDN control plane could be centralized (one centralized SDN controller), decentralized (the SDN controller is distributed over all UAV nodes), or hybrid in which the processing control of the forwarded packets can be performed locally on each UAV node and control traffic also exists between the centralized SDN controller and all other SDN elements \cite{gupta2016survey}.

The controller collects network statistics from OpenFlow switches, and it also needs to know the latest topology of the UAVs network. Therefore, it is important to maintain the connectivity of the SDN controller with the UAV nodes.
The centralized SDN controller has a global view of the network. 
OpenFlow switches contain software flow tables and protocols to provide a communication between network and control planes, then the controller determines the path and tells network elements where to forward the packets \cite{mckeown2008openflow}.

\begin{figure}[!h]
	\centering
	\includegraphics[scale=0.335]{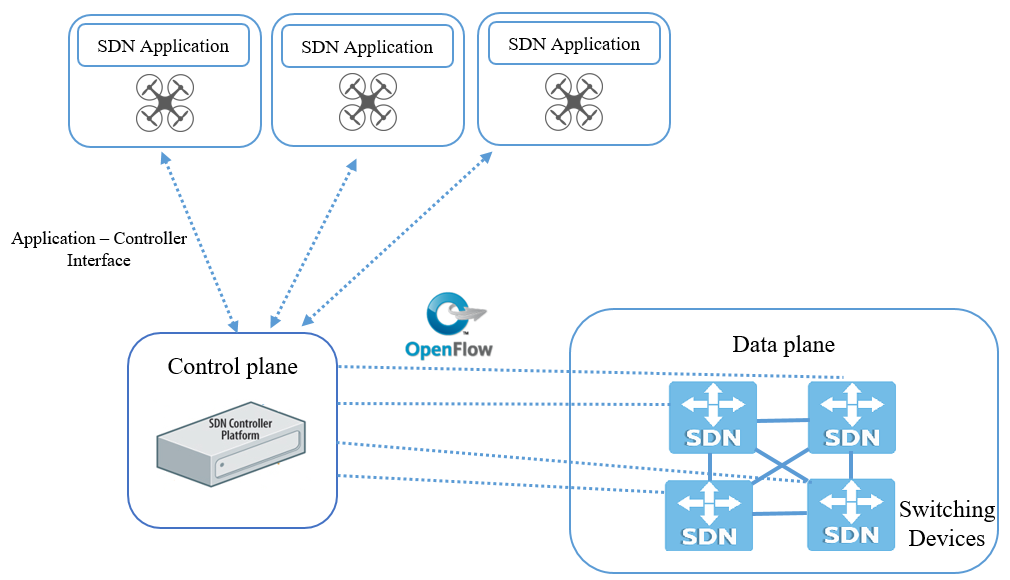}
	\caption{Control and Data Planes in SDN Platform.  }
	\label{fig:ahmad14}
\end{figure}

Figure~\ref{fig:ahmad14} shows separation of control and data planes in SDN platform with OpenFlow interface.

\subsubsection*{$d$) Low power and Lossy Networks (LLT)}
In FANETs, UAV nodes are typically characterized by limited resources, such as battery, memory and computational resources. LLN composed of many embedded devices such as UAVs with limited power, and it can be considered as a promising approach to be used in FANET to handle power challenge in UAVs \cite{winter2012rpl, rault2013multi}. 

The Internet Engineering Task Force (IETF) has defined routing protocol for LLN known as routing for low-power and lossy network (RPL) \cite{winter2012rpl}, this protocol can utilize different low power communication technologies, such as low power WiFi, IEEE 802.15.4 \cite{domingo2012overview}, which is a standard for low-power and low rate wireless communications and can be used for UAV-UAV communications in FANET \cite{zafar2017reliable}. The IPv6 over Low Power Wireless Personal Area Networks (6LoWPAN) defines a set of protocols that can be used to integrate nodes with IPv6 datagrams over constrained networks such as FANET \cite{atzori2010internet}.

\subsubsection{{Future Insights}}
Based on the reviewed literature of the networking challenges and new networking trends of FANET, we suggest these future insights:

\begin{itemize}
	
	\item SDN services can be developed to support FANET functions, such as surveillance, safety and security services. 
	
	\item  Further studies are needed to study the SDN deployment in FANET with high reliability, reachability and fast mobility of UAV nodes \cite{xia2015survey}
	
	\item More research is needed to develope specific security protocols for FANETs based on the SDN paradigm.
	
	\item More studies are needed to replace the conventional network devices by fully decoupled SDN switching devices, in which the control plane is completely decoupled from data plane and the routing protocols can be  
	executed on-board from SDN switching devices \cite{xia2015survey}.
	
	\item For NFV, more research is needed on optimal virtual network topology, customized end-to-end protocols and dynamic network embedding \cite{zhang2017software}. 
	
	\item New routing protocols need to be tailored to FANETs to conserve energy, satisfy bandwidth requirement, and ensure the quality of service \cite{sahingoz2014networking}. 
	
	
	\item Most of the existing MANET routing protocols partially fail to provide reliable communications among UAVs. So, there is a need to design and implement new routing protocols and networking models for FANETs \cite{bekmezci2013flying}.
	
	\item FANETs share the same wireless communications bands with other applications such as satellite communications and GSM networks. This leads to frequency congestion problems.  Therefore, there is a need to standardize FANET communication bands to mitigate this problem \cite{gupta2016survey}.

	\item FANETs lack security support. Each UAV in FANET is required to exchange messages and routing information through wireless links. Therefore, FANETs are vulnerable to attacks. As a result, it is important to design and implement secure FANET routing protocols.

	\item The design of congestion control algorithms become an important issue for FANETs. Current research efforts focus on modifying and improving  protocols instead of developing new transport protocols that better suite  FANETs.  \cite{ivancic2012evaluation}.
	
\end{itemize}

\subsection{Security Challenges}

Figure \ref{fig1} illustrates the general architecture of UAV systems \cite{javaid2012cyber}, which includes: (1) UAV units; (2) GCSs; (3) satellite (if necessary); and (4) communications links. Various components of the UAV systems provide a large attack surface for malicious intruders which brings huge cyber security challenges to the UAV systems. In this section, to present a comprehensive view of the cyber security challenges of UAV systems, we first summarize the attack vectors of general UAV systems. Then, based on the attack vectors and the potential capabilities of attackers, we classify the cyber attacks/challenges against UAV systems into different categories. More importantly, we present a comprehensive literature review of the state-of-the-art countermeasures for these security challenges. Finally, based on the reviewed literature, we summarize the security challenges and provide high level insights on how to approach them.

\subsubsection{Attack Vectors in UAV Systems}

\begin{figure}
	\centering
	\includegraphics[scale=0.4]{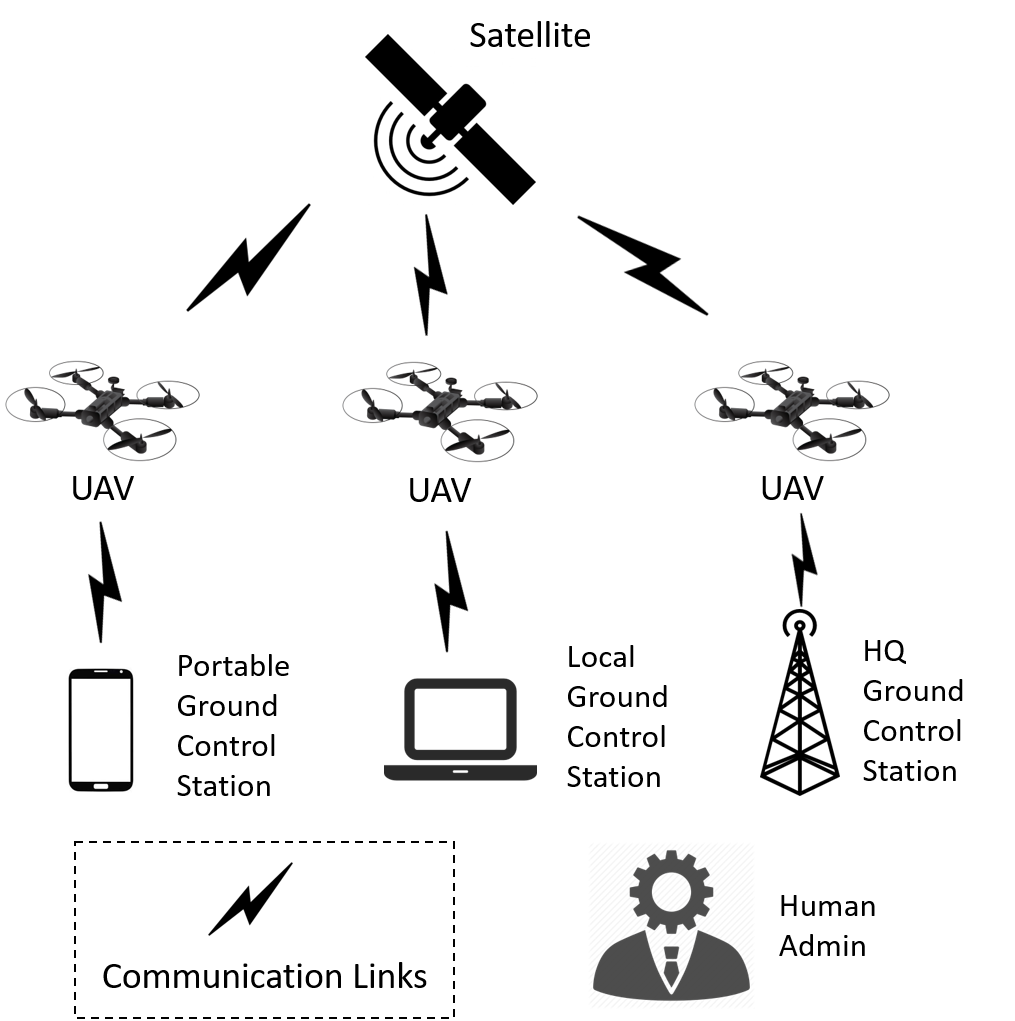}
	\caption{General Architecture of UAV Systems. Reprinted from \cite{javaid2012cyber}.}
	\label{fig1}
\end{figure}

Based on the architecture illustrated in Figure \ref{fig1}, we identify four attack vectors:
\begin{itemize}
	\item Communications links: various attacks could be applied to the communications links between different entities of UAV systems, such as eavesdropping and session hijacking.
	\item UAVs themselves: direct attacks on UAVs can cause serious damage to the system. Examples of  such attacks are signal spoofing and identity hacking. 
	\item GCSs: attacks on GCSs are more fatal than others because they issue commands to actual devices and collect all the data from the UAVs they control. Such attacks normally involve malwares, viruses and/or key loggers.
	\item Humans: indirect attacks on human operators can force wrong/malicious operating commands to be issued to the system. This type of attacks is usually enabled with social engineering techniques.
\end{itemize}

\subsubsection{Taxonomy of Cyber Security Attacks/Challenges Against UAV Systems}

\begin{figure*}[!h]
	\centering
	\begin{forest}
		for tree={
			align=center,
			parent anchor=south,
			child anchor=north,
			font=\footnotesize,
			edge={thick, -{Stealth[]}},
			l sep+=10pt,
			edge path={
				\noexpand\path [draw, \forestoption{edge}] (!u.parent anchor) -- +(0,-10pt) -| (.child anchor)\forestoption{edge label};
			},
			if level=0{
				inner xsep=0pt,
				tikz={\draw [thick] (.south east) -- (.south west);}
			}{}
		}
		[ {Cyber security challenges of UAV applications}  
		[Confidentiality
		[UAVs
		[{Key-logger, \\malware, etc.}]
		]
		[GCSs
		[ { Hijacking,\\ etc.}]
		]
		[Humans
		[{Social\\ engineering}]
		]
		[{Comms. links}
		[{Eavesdrop,\\ Hijacking, etc.}]
		]
		]
		[Availability
		[DOS/DDOS
		[{Buffer overflow,\\ flooding, etc.}]
		]
		]
		[Integrity
		[Fabrication Modification
		[{Signal spoofing,\\ MitM, etc.}]
		]
		]
		]
		]
	\end{forest}
	\caption{Cyber Attack/Challenges Taxonomy in UAV Systems.}\label{fig2}
\end{figure*}
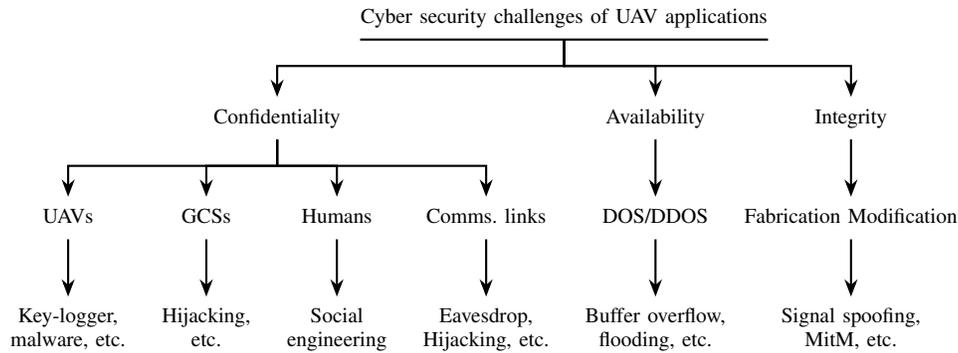

Based on the attack vectors and capabilities of attackers, as well as the work in \cite{javaid2012cyber}, we classify the possible attacks against UAV systems into different categories. Figure \ref{fig2} presents a graph that summarizes the proposed attack taxonomy. This attack model defines three general cyber security challenges for the UAV systems, namely, confidentiality challenges, integrity challenges, and availability challenges.

Confidentiality refers to protecting information from being accessed by unauthorized parties. In other words, only the people who are authorized to do so can gain access to data. Attackers could compromise the confidentiality of UAV systems by various approaches (e.g., malware, hijacking, social engineering, etc.) utilizing different attack vector.

Integrity ensures the authenticity of information. Attackers could modify or fabricate information of UAV systems (e.g., data collected, commands issued, etc.) through communications links, GCSs or compromised UAVs. For example, GPS signal spoofing attack. 

Availability ensures that the services (and the relevant data) that UAV systems carry are running as expected and are accessible to authorized users. Attackers could perform DoS (Deny of Service) attacks on UAV systems by, for example, flooding the communications links, overloading the processing units, or depleting the batteries.

\subsubsection{Literature Review of the State-of-the-art Security Attacks/Challenges and Countermeasures}\label{review_security}

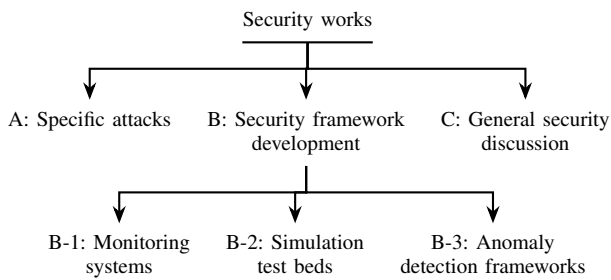
\begin{figure}[!h]
	\centering
	\begin{forest}
		for tree={
			align=center,
			parent anchor=south,
			child anchor=north,
			font=\footnotesize,
			edge={thick, -{Stealth[]}},
			l sep+=10pt,
			edge path={
				\noexpand\path [draw, \forestoption{edge}] (!u.parent anchor) -- +(0,-10pt) -| (.child anchor)\forestoption{edge label};
			},
			if level=0{
				inner xsep=0pt,
				tikz={\draw [thick] (.south east) -- (.south west);}
			}{}
		}
		[  Security works 		
		[A: Specific attacks 
		]
		[B: Security framework\\ development 
		[{B-1: Monitoring\\ systems}
		]
		[{B-2: Simulation\\ test beds}		
		]
		[{B-3:  Anomaly\\ detection frameworks}		
		]
		]
		[C: General security\\ discussion 
		]
		]
	\end{forest}
	\caption{Summary of the Literature of the State-of-the-Art Attacks/Challenges and Countermeasures in UAV systems.}
	\label{fig3}
\end{figure}

The UAV related cyber security research can be classified into three main categories as shown in Figure \ref{fig3}, namely:
\begin{itemize}
	\item A: Specific attack discussion: This line of research focuses on one specific type of attack (e.g., signal spoofing) and proposes corresponding analysis or countermeasures.
	\item B: General security analysis: This line of research presents the high level analysis, discussion or modeling of various attacks that exist in current UAV systems.
	\item C: Security framework development: This line of research introduces new monitoring systems, simulation test beds or anomaly detection frameworks for the state-of-the-art UAV applications.
\end{itemize} 
Table \ref{t3} summarizes the 15 relevant papers that we have reviewed including their attack vectors and categories. In addition, for the papers that discuss specific attacks/challenges, we summarize the proposed countermeasures, their limitations and propose high level countermeasures as guidelines for future enhancements. 

\begin{sidewaystable*}[]
	\footnotesize
	\centering
	\caption{\uppercase{The summary of the literature of state-of-the-art attacks and countermeasures for UAV}}
	\label{t3}
	\begin{tabular}{|l|l|l|l|l|l|l|}
		\hline
		\multicolumn{1}{|c|}{\textbf{}}   & \multicolumn{1}{c|}{\textbf{Year}} & \multicolumn{1}{c|}{\textbf{Attack Vector}} & \multicolumn{1}{c|}{\textbf{Category}}                                                      & \multicolumn{1}{c|}{\textbf{\begin{tabular}[c]{@{}c@{}}Proposed\\ Countermeasure\end{tabular}}}                                                          & \multicolumn{1}{c|}{\textbf{Limitations}}                                                                                                                          & \multicolumn{1}{c|}{\textbf{\begin{tabular}[c]{@{}c@{}}Suggested \\ Countermeasures\end{tabular}}}                                                                                                                                                \\ \hline
		\cite{javaid2012cyber}          & 2012                               & Communication Links, UAV, GCSs              & C                                                                                           & N/A                                                                                                                                                      & N/A                                                                                                                                                                & N/A                                                                                                                                                                                                                                               \\ \hline
		\cite{giray2013anatomy}         & 2013                               & Communication Links                         & \begin{tabular}[c]{@{}l@{}}A: Signal integrity\\  (GPS spoofing)\end{tabular}               & \begin{tabular}[c]{@{}l@{}}1. Cryptography based\\ signal authentication;\\ 2. Multiple receiver design.\end{tabular}                                    & \begin{tabular}[c]{@{}l@{}}1. Lack of details;\\ 2. Require hardware changes \\ (i.e., complexity and cost issue).\end{tabular}                                    & \begin{tabular}[c]{@{}l@{}}1. Strong authentication \\ (e.g., trust platform module,\\  Kerberos, etc.);\\ 2. Signal distortion detection;\\ 3. Direction-of-arrival sensing \\ (i.e.,  transmitter antenna \\ direction detection).\end{tabular} \\ \hline
		\cite{javaid2013uavsim}         & 2013                               & Communication Links, UAV, GCSs              & B-1                                                                                         & N/A                                                                                                                                                      & N/A                                                                                                                                                                & N/A                                                                                                                                                                                                                                               \\ \hline
		\cite{goppert2014software}      & 2014                               & Communication Links, UAV, GCSs              & C                                                                                           & N/A                                                                                                                                                      & N/A                                                                                                                                                                & N/A                                                                                                                                                                                                                                               \\ \hline
		\cite{maxa2015secure}           & 2015                               & Communication Links, UAVs                   & \begin{tabular}[c]{@{}l@{}}A: Routing attack\\ (eavesdropping, DOS, etc.)\end{tabular}      & \begin{tabular}[c]{@{}l@{}}A secure routing protocol\\ and its modeling process.\end{tabular}                                                            & \begin{tabular}[c]{@{}l@{}}Lack of evaluations \\ or security proof.\end{tabular}                                                                                  & \begin{tabular}[c]{@{}l@{}}Perform evaluations under \\ various attack scenarios \\ or provide theoretical \\ security proof.\end{tabular}                                                                                                        \\ \hline
		\cite{schumann2015r2u2}         & 2015                               & UAVs                                        & B-2                                                                                         & N/A                                                                                                                                                      & N/A                                                                                                                                                                & N/A                                                                                                                                                                                                                                               \\ \hline
		\cite{birnbaum2015unmanned}     & 2015                               & UAVs                                        & B-1                                                                                         & N/A                                                                                                                                                      &                                                                                                                                                                    &                                                                                                                                                                                                                                                   \\ \hline
		\cite{muzzi2015using}           & 2015                               & Communication Links, UAVs                   & A: DDoS                                                                                     & \begin{tabular}[c]{@{}l@{}}Botnet platform\\  (as a future work).\end{tabular}                                                                           & \begin{tabular}[c]{@{}l@{}}There is no concrete \\ solution provided.\end{tabular}                                                                                 & \begin{tabular}[c]{@{}l@{}}1. Delivers behavior-based \\ anomaly detection\\  (e.g., packet analysis about \\ its type, rate, volume, \\ source IP, etc.);\\ 2. Enables immediate \\ response to DDoS attacks;\end{tabular}                       \\ \hline
		\cite{vattapparamban2016drones} & 2016                               & Communication Links, UAV, GCSs              & C                                                                                           & N/A                                                                                                                                                      & N/A                                                                                                                                                                & N/A                                                                                                                                                                                                                                               \\ \hline
		\cite{sedjelmaci2016detect}     & 2016                               & Communication Links, UAV, GCSs              & B-3                                                                                         & N/A                                                                                                                                                      & N/A                                                                                                                                                                & N/A                                                                                                                                                                                                                                               \\ \hline
		\cite{davidson2016controlling}  & 2016                               & Communication Links, UAVs                   & \begin{tabular}[c]{@{}l@{}}A: Signal integrity\\  (sensor input spoofing)\end{tabular}      & \begin{tabular}[c]{@{}l@{}}Optical flow analysis\\ (RANSAC algorithm\\ \cite{fischler1981random})\end{tabular}                                         & \begin{tabular}[c]{@{}l@{}}The attack model is too \\ demanding (i.e., it is difficult \\ to perform this attack. ).\end{tabular}                                  & \begin{tabular}[c]{@{}l@{}}1. Strong firewall/authentication \\ schemes to prevent hijacking attacks;\\ 2. Delivers behavior-based \\ anomaly detection.\end{tabular}                                                                             \\ \hline
		\cite{mcneely2016detection}     & 2016                               & UAVs                                        & A: UAV hijacking                                                                            & \begin{tabular}[c]{@{}l@{}}Profiling of in-flight\\ statistics.\end{tabular}                                                                             & \begin{tabular}[c]{@{}l@{}}Temporary control instability\\  (e.g., short time decreasing of\\ amplitude) caused by\\ the attacker cannot be detected.\end{tabular} & \begin{tabular}[c]{@{}l@{}}1. Decreasing the detection \\ interval;\\ 2. Strong firewall/authentication \\ schemes to prevent hijacking attacks;\\ 3. Packets analysis of malicious \\ injected commands.\end{tabular}                            \\ \hline
		\cite{hagerman2016security}     & 2016                               & N/A                                         & B-2                                                                                         & N/A                                                                                                                                                      & N/A                                                                                                                                                                & N/A                                                                                                                                                                                                                                               \\ \hline
		\cite{rodday2016exploring}      & 2016                               & Communication Links                         & \begin{tabular}[c]{@{}l@{}}A: Hijacking \\ (MitM attack)\end{tabular}                       & \begin{tabular}[c]{@{}l@{}}1. XBee 868LP on-board\\ encryption;\\ 2. Dedicated hardware\\ encryption;\\ 3. Application layer \\ encryption.\end{tabular} & No details provided.                                                                                                                                               & \begin{tabular}[c]{@{}l@{}}1. Integrate detection and immediate \\ response mechanisms.\\ 2. Strong authentication \\ (e.g., trust platform module, Kerberos, etc.);\end{tabular}                                                                 \\ \hline
		\cite{wang2016authentication}   & 2017                               & Communication Links                         & \begin{tabular}[c]{@{}l@{}}A: Authentication/key\\ management targeted attacks\end{tabular} & \begin{tabular}[c]{@{}l@{}}A set of modified \\ authentication and key\\ management protocols.\end{tabular}                                              & \begin{tabular}[c]{@{}l@{}}Lack of evaluations \\ or security proof.\end{tabular}                                                                                  & \begin{tabular}[c]{@{}l@{}}Perform evaluations under \\ various attack scenarios \\ or provide theoretical \\ security proof.\end{tabular}                                                                                                        \\ \hline
		\cite{podhradsky2017improving}  & 2017                               & Communication Links, GCS, UAV               & C                                                                                           & N/A                                                                                                                                                      & N/A                                                                                                                                                                & N/A                                                                                                                                                                                                                                               \\ \hline
	\end{tabular}
\end{sidewaystable*}

\begin{itemize}
	\item Challenge of specific attack - GPS spoofing: GPS spoofing attack refers to the malicious attempt to manipulate the GPS signals in order to achieve the benefits of the attackers.
	
	In \cite{giray2013anatomy}, the authors provide an analysis of UAV hijacking attack (e.g., GPS signal spoofing) with an anatomical approach and outline a model to illustrate: (1) that such attacks can be observed to reveal their vulnerabilities; (2) how to exploit such attacks; (3) how to provide countermeasures and risk mitigation; and (4) details of the impact of such attacks. The use of anatomical investigation of a UAV hijacking aims to provide insights that help all the practitioners of the UAV systems. The article mainly focuses on the analysis of GPS signal spoofing attacks. However, the proposed countermeasures (i.e., cryptography based signal authentication and multiple receiver design) are not novel and are only described at high level without sufficient details.
	
	\item Challenge of specific attack - DDoS (Distributed Deny of Service) attack: DDoS attack is an attempt to make the UAV systems unavailable/unreachable by overwhelming it with traffic from multiple sources. 
	
	In \cite{muzzi2015using}, the authors present a platform to conduct a DDoS attack against UAV systems and layout the general mitigation method. Firstly, the authors present a DDoS attack modeling and specification. Then, they analyze the effects caused by this kind of attack. However, the proposed attack scenario is too simple (i.e., UDP flooding) and only a high level solution is said to be included in the future work. 
	
	\item Challenge of specific attack - sensor input spoofing attack: In \cite{davidson2016controlling}, the authors introduce a new attack against UAV systems, named sensor input spoofing attack. If the attacker knows exactly how the sensor algorithms work, he can manipulate the victim's environment to create a new control channel such that the entire UAV system is in dangerous. In addition, they provide suitable mitigation for the introduced attack. To perform the proposed sensor input spoofing attack, the attacker must meet three requirements: (1) environment influence requirement: the attacker must be able to modify the physical phenomenon that the sensor measures; (2) plausible input requirement: the attacker must be able to generate valid input that the sensor system can read; and (3) meaningful response requirement: the attacker must be able to generate meaningful UAV behavior corresponding to the spoofed input. This attack model is too demanding to be practical for many UAV applications and  no convincing justifications were provided.  
	
	\item Challenge of specific attack - hijacking attacks: hijacking attack is a type of attack in which the attacker takes control of a communication between two parties and masquerades as one of them (e.g., inject extra commands).
	
	In \cite{rodday2016exploring},  the authors demonstrate the capabilities of the attackers who target UAV systems to perform Man-in-the-Middle attacks (MitM) and control command injection attacks. They also propose corresponding countermeasures to help address these vulnerabilities. However, the introduced MitM attack requires fully compromised UAV system via reverse-engineering techniques which largely limits the feasibility/practicability of this type of attacks. In addition, the proposed encryption scheme lacks the necessary details that warrant credible evaluation of robustness and security. 
	
	The article of \cite{mcneely2016detection} proposes an anomaly detection scheme based on flight patterns fingerprint via statistical measurement of flight data. Firstly, a baseline flight profile is generated. Then, simulated hijacking scenarios are compared to the baseline profile to determine the detection result. The proposed scheme is able to detect all direct hijacking scenarios (i.e., assume a fully compromised UAV and the flight plan has been altered to some random places.). However, temporary control  instability (e.g., short time decreasing of amplitude) caused by the attacker cannot be detected.

	\item Challenge of specific attack - Authentication/key management targeted attacks: The authors in \cite{wang2016authentication}  first propose a communication architecture to integrate LTE technology into integrated CNPC (Control and Non-Payload Communication) networks. Then, they define several security requirements of the proposed architecture. Also, they modify the authentication, the key agreement and the handover key management protocols to make them suitable to the integrated architecture. The proposed modified protocol is proved to outperform the LTE counterpart protocols via a comparative analysis. In addition, it introduces almost the same amount of communications overhead.
	
	\item Challenge of security framework development: In \cite{javaid2013uavsim}, the authors introduce UAVSim, a simulation testbed for Unmanned Aerial Vehicle Networks cyber security analysis. The proposed test bed can be used to perform various experiments by adjusting different parameters of the networks, hosts and attacks. The UAVSim consists of five modules: (1) Attack library: generating various attacks (i.e., jamming attacks and DoS attacks); (2) UAV network module: forming different UAV networks; (3) UAV model library: providing different host functions (e.g., attack hosts or UAV hosts); (4) Graphical user interface; (5) Result analysis module; and (6) UAV Module browser. The proposed test bed is helpful in analyzing existing attacks against UAV systems by adjusting different parameters. However, the proposed attack library only contains two types of attacks for now (i.e., jamming attacks and DoS attacks). Therefore, UAVSim must be enriched with other types of attacks to improve its applicability and usability.
	
	The authors in \cite{schumann2015r2u2} present R2U2, a novel framework for run-time monitoring of security threats and system behaviors for Unmanned Aerial Systems (UAS). Specifically, they extend previous R2U2 from only monitoring of hardware components to both hardware and software configuration monitoring to achieve security threats diagnostic. Also, the extended version of R2U2 provides detection of attack patterns (i.e., ill-formatted and illegal commands, dangerous commands, nonsensical or repeated navigation commands and transients in GPS signals) rather than component failures. In addition, the FPGA implementation ensures independent monitoring and achieves software re-configurable feature. The extended version of R2U2 is more advanced than the original prototype in terms of attack pattern recognition and secure \& independent monitoring. However, the independent monitoring system introduces a new security hole (i.e., the monitoring system itself) without introducing appropriate security mechanism to mitigate it. 
	
	In the work of \cite{birnbaum2015unmanned}, the authors present a UAV monitoring system that captures flight data to perform real-time abnormal behavior detection. If an abnormal behavior is detected, the system will raise an alert. The proposed monitoring system includes the following features:
	\begin{itemize}
		\item A specification language for UAV in-flight behavior modeling;
		\item An algorithm to covert a given flight plan into a behavioral flight profile;
		\item A decision algorithm to determine if an actual UAV flight behavior is normal or not based on the behavioral flight profile;
		\item A visualization system for the operator to monitor multiple UAVs. 
	\end{itemize}
	The proposed monitoring system can detect abnormal behaviors based on the in flight data. However, attacks which do not alter in-flight data are not detectable (e.g., attacks that only collect flight data).
	
	The authors of \cite{sedjelmaci2016detect} propose and implement a cyber security system to protect UAVs from several dangerous attacks, such as attacks that target data integrity and network availability. The detection of data integrity attacks uses Mahalanobis distance method to recognize the malicious UAV that forwards erroneous data to the base station. For the availability attacks (i.e., wormhole attack in this paper), the detection process is summarized as follows:
	\begin{itemize}
		\item The UAV relays a packet when it becomes within the range of the base station. The packet includes: (1) node's type (source, relay or destination); and (2) its next (and previous) hops.
		\item The base station collects the forwarded packets from the UAVs, verifies whether the relay node forwards a packet or not and computes the Message Dropping Rate (MDR).
		\item The base station will raise an alert if the MDR is higher than a false MDR assigned to a normal UAV.
	\end{itemize}
	The proposed detection scheme for data integrity attacks and network availability attacks is proved to achieve high detection accuracy via simulations. The attack model assumes that every UAV node has the detection module and the node which is determined as malicious node would lose the ability to invoke the detection module. However, there is no co-operation detection algorithm provided and only single node detection procedure is introduced.
	
	In the work of \cite{hagerman2016security}, the authors present an improved mechanism for UAV security testing. The proposed approach uses a behavioral model, an attack model, and a mitigation model to build a security test suite. The proposed security testing goes as follows: 
	\begin{itemize}
		\item  Model system behavior: associates behavior criteria (BC) with a behavioral model (BM) and build a behavioral test (BT).
		\item  Attack type definition: defines attack types (A) with attack criteria (AC) and determine attack applicability matrix (AM).
		\item Determine security test requirements: determines at which point during the operation of a behavioral test an attack will occur.
		\item Test generation: a required mitigation process is injected.
	\end{itemize}
	This work provides a systematic approach to identify vulnerabilities in UAV systems and provides corresponding mitigation. However, the proposed mitigation methods are limited to state roll back or re-executing. In other words, other prevention or detection methods (e.g., encryption or anomaly detection) cannot be included in the proposed scheme.
	
	\item Challenges of security analysis works: In the work of \cite{javaid2012cyber}, the authors perform an overall security threat analysis of a UAV system as described in Figure \ref{fig1}.  A cyber security threat model is proposed and analyzed to show existing or possible attacks against UAV systems. This model helps designers and users of UAV systems across different aspects, including: (1) understanding threat profiles of different UAV systems; (2) addressing various system vulnerabilities; (3) identifying high priority threats; and (4) selecting appropriate mitigation techniques. In addition, a risk evaluation mechanism (i.e., a standard risk evaluation grid included in the ETSI threat assessment methodology ) is used to assess the risk of different threats in UAV systems. Even though this paper provides a comprehensive profile of existing or potential attacks for UAV systems, the proposed attacks are general/high level threats that did not take into consideration hardware and software differences among different UAV systems.
	
	The survey paper of \cite{vattapparamban2016drones} reviews various aspects of drones (UAVs) in future smart cities, relating to cyber security, privacy, and public safety. In addition, it also provides representative results on cyber attacks using UAVs. The studied cyber attacks include de-authentication attack and GPS spoofing attack. However, no countermeasures are discussed. 
	
	The authors in \cite{podhradsky2017improving} introduce an implementation of an encrypted Radio Control (RC) link that can be used with a number of popular RC transmitters. The proposed design uses Galois Embedded Crypto library together with openLRSng open-source radio project. The key exchange algorithm is described below:
	\begin{itemize}
		\item $TX$ first generates the ephemeral key $K_e$ and , then encrypts $K_e$ using permanent key $K_p$ and $IV_{rand}$. The encrypted result is $m_1^{TX}$ and sent to $RX$.
		\item $RX$ decrypts $m_1^{TX}$ with $K_p$ to obtain $K_e$, then generate $K_e^{'}$ and encrypts $K_e^{'}$ with $K_e$ and $IV_0=0$. The encrypted result is $m_2^{RX}$ and sent to $TX$.
		\item $TX$ decrypts $m_2^{RX}$ with $K_e$ and $IV_0=0$ to verify that $RX$ has a copy of $K_e$. Then $TX$ sends $ACK$ message encrypted with $K_e^{'}$ and $IV_0$ to $RX$ as a confirmation (denoted as $m_3^{TX}$ ).
		\item $RX$ receives $m_3^{TX}$it to verify that $TX$ has a copy of $K_e^{'}$.  Key exchange is then successful.
	\end{itemize}
	The proposed scheme achieves secure communication link for open source UAV systems. However, the symmetric key is assumed to be generated in a third-party trusted computer and hard-coded in the source code, which degrades the security guarantees and limits the feasibility of the scheme.
	
\end{itemize}

\subsubsection{Summarizing of Cyber Security Challenges for UAV Systems}
Based on the reviewed literature, we summarize four key findings of the cyber security challenges on UAV applications:
\begin{itemize}
	\item DoS attacks and hijacking attacks (i.e., for signal spoofing) are the most prevailing threats in the UAV systems: Various DoS attacks have been found and proved to cause serious availability issues in UAV systems. In addition, signal spoofing via hijacking attacks is able to severely damage the behaviors of certain UAV systems.
	
	\textit{Possible solution}: (1) Strong authentication (e.g., trust platform module, Kerberos, etc.); (2) Signal distortion detection; and (3) Direction-of-arrival sensing (i.e., transmitter antenna direction detection). 
	\item Existing countermeasure algorithms are limited to single UAV systems: most of the existing schemes are designed only for single UAV systems. A few of them discuss multiple UAV scenarios without providing any concrete solutions.
	
	\textit{Possible solution}: develop co-operation countermeasure algorithms for multiple UAVs. For example, modify and adopt existing distributed security frameworks (e.g., Kerberos) to multiple UAVs systems; (2) 
	\item Most of the current security analysis of UAV systems overlook the hardware/software differences (e.g., different hardware platform, different communication protocols, etc.) among various UAV systems: On one hand, some of the attacks only exist in specific hardware or software configuration. On the other hand, most of the proposed countermeasures are well studied solutions in other communication systems and there could be many deployment difficulties when applying them on different UAV systems.
	
	\textit{Possible solution}: design unified/standard deployment interface or language for the diverse UAV systems.
	\item Current UAV simulation test beds are still far from mature: Existing emulators for UAV security analysis are limited to few attack scenarios and specific hardware/software configurations.
	
	\textit{Possible solution}: leveraging powerful simulation tools (e.g., Labview) or design customized simulation environments. 
\end{itemize}

\section{Conclusion}
The use of UAVs has become ubiquitous in many civil applications. From rush hour delivery services to scanning inaccessible areas, UAVs are proving to be critical in situations where humans are unable to reach or cannot perform dangerous/risky tasks in a timely and efficient manner. In this survey, we review UAV civil applications and their challenges. We also discuss current research trends and provide future insights for potential UAV uses.

In SAR operations, UAVs can provide timely disaster warnings and assist in speeding up rescue and recovery operations. They can also carry medical supplies to areas that are classified as inaccessible. Moreover, they can quickly provide coverage of a large area without ever risking the security or safety of the personnel involved. Using UAVs in SAR operations reduces costs and human lives.

In remote sensing, UAVs equipped with sensors can be used as an aerial sensor network for environmental monitoring and disaster management. They can provide numerous datasets to support research teams, serving a broad range of applications such as drought monitoring, water quality monitoring, tree species, disease detection, etc. In risk management, insurance companies can utilize UAVs to generate NDVI maps in order to have an overview of the hail damage to crops, for instance.

Civil infrastructure is expected to dominate the addressable market value of UAV that is forecast to reach $\$45$ Billion in the next few years. In construction and infrastructure inspection applications, UAVs can be used to monitor real-time  construction project sites. They can also be utilized in power line and gas pipeline inspections. Using UAVs in civil infrastructure applications can reduce work-injuries, high inspection costs and time involved with conventional inspection methods.

In agriculture, UAVs can be efficiently used in irrigation scheduling, plant disease detection, soil texture mapping, residue cover and tillage mapping, field tile mapping, crop maturity mapping and crop yield mapping. The next generation of UAV sensors can provide on-board image processing and in-field analytic capabilities, which can give farmers instant insights in the field, without the need for cellular connectivity and cloud connection.

With the rapid demise of snail mail and the massive growth of e-Commerce, postal companies have been forced to find new methods to expand beyond their traditional mail delivery business models. Different postal companies have undertaken various UAV trials to test the feasibility and profitability of UAV delivery services. To make UAV delivery practical, more research is required on UAVs design. UAVs design should cover creating aerial vehicles that can be used in a wide range of conditions and whose capability rivals that of commercial airliners.

UAVs have been considered as a novel traffic monitoring technology to collect information about real-time road traffic conditions. Compared to the traditional monitoring devices, UAVs are cost-effective and can monitor large continuous road segments or focus on a specific road segment. However, UAVs have slower speeds compared to vehicles driving on highways. A possible solution might entail changing the regulations to allow UAVs to fly at higher altitudes. Such regulations would allow UAVs to benefit from high views to compensate the limitation in their speed.

One potential benefit of UAVs is the capability to fill the gaps in current border surveillance by improving coverage along remote sections of borders. Multi-UAV cooperation brings more benefits than single UAV surveillance, such as wider surveillance scope, higher error tolerance, and faster task completion time. However, multi-UAV surveillance requires more advanced data collection, sharing, and processing algorithms. To develop more efficient and accurate multi-UAV cooperation algorithms, advanced machine learning algorithms could be utilized to achieve better performance and faster response.

The use of UAVs is rapidly growing in a wide range of wireless networking applications. UAVs can be used to provide wireless coverage during emergency cases where each UAV serves as an aerial wireless base station when the cellular network goes down. They can also be used to supplement the ground base station in order to provide better coverage and higher data rates. Utilizing UAVs in wireless networks needs further research, where topology formation, cooperation between UAVs in a multi-UAV network, energy constraints, and mobility modelس are challenges facing UAVs in wireless networks.

Throughout the survey, we discuss the new technology trends in UAV applications such as mmWave, SDN, NFV, cloud computing and image processing. We thoroughly identify the key challenges for UAV civil applications such as charging challenges, collision avoidance and swarming challenges, and networking and security related challenges. In conclusion, a complete legal framework and institutions regulating the civil uses of UAVs are needed to spread UAV services globally. We hope that the key research challenges and opportunities described in this survey will help pave the way for researchers to improve UAV civil applications in the future.

\section{ACRONYMS}
The acronyms and abbreviations used throughout the survey and their
definitions are listed bellow:
\scriptsize
\begin{abbreviations}
	\item [6LoWPAN]	IPv6 over Low Power Wireless Personal Area Networks. 
	\item       [AC]	Attack Criteria.
	\item 	    [AODV]	Ad-hoc On-demand Distance Vector.
	\item       [ATG]	Air-to-Ground. 
	\item       [BC]	Behavior Criteria. 
	\item 	    [BDMA]	Beam Division Multiple Access. 
	\item 	    [BM]	Behavioral Model. 
	\item 	    [BP]	Bundle Protocol. 
	\item 	    [BT]	Behavioral Test.
	\item		[CC]	Capacitive Coupling. 
	\item		[CCSD]	Consultative Committee for Space Data Systems. 
	\item		[CFDP]	CCSD File Delivery Protocol. 
	\item		[CLP]	Convergence Layer Protocol. 
	\item		[CNN]	Convolutional Neural Network.
	\item		[CNO]	Cellular Network Operator. 
	\item		[CNPC]	Control and Non-Payload Communication.
	\item		[CoA]	Certificate of Authorization.
	\item		[DDoS]	Distributed Deny of Service.
	\item		[DMF]	Drone-cell  management  frame-work. 
	\item		[DoS]	 Deny of Service.
	\item		[DTN]	Disruption Tolerant Networking. 
	\item       [EVI]   Enhanced Vegetation Index.
	\item		[FAA]	Federal Aviation Administration.
	\item       [FANETs]  Flying Ad-Hoc Networks.
	\item		[FLIR]	Forward Looking Infrared. 
	\item		[FMI]	Fourier Mellin Invariant. 
	\item		[FSO]	Free Space Optical. 
	\item		[GCS]	Ground Control Station.
	\item		[GIS]	Geographic Information System.
	\item		[GPS]	Global Position System.
	\item       [GNDVI]   Green Normalized Difference Vegetation index.
	\item       [GVI]     Green Vegetation Index.
	\item		[HAP]	High Altitude Platform.
	\item		[IETF]	Internet Engineering Task Force.
	\item		[IMU]	Inertial Sensor. 
	\item		[IoT]	Internet of Things.
	\item		[IR]	Infra Red.
	\item		[ITU]	International Telecommunication Union. 
	\item		[KDOT]	Kansas Department of Transportation.
	\item	    [LAP]	Low Altitude Platform.
	\item		[LLT]	Low power and Lossy Networks. 
	\item		[LOS]	Line of Sight. 
	\item		[LRF]	Laser Range Finder. 
	\item		[LTP]	Licklider Transmission Protocol.
	\item		[MAC]	Medium Access Control. 
	\item		[MANETs]	Mobile Ad-hoc Networks. 
	\item		[MB-LBP]	Multi-scale Block Local Binary Patterns. 
	\item		[MDR]	Message Dropping Rate.
	\item		[MEC]	Mobile Edge Computing.
	\item		[MitM]	Man-in-the-Middle attacks. 
	\item		[mmWave]	Millimeter-Wave. 
	\item		[MPC]	Model Predictive Controller.
	\item		[MRC]	Magnetic Resonance Coupling. 
	\item		[NDVI]	Normalized difference vegetation index.
	\item		[NLOS]	Non-Line of Sight. 
	\item       [OBIA]      Object-Based Image Analysis.
	\item		[OD]	Origin-Destination. 
	\item		[PA]	Precision Agriculture.
	\item		[PBR]	Performance-Based Regulations. 
	\item		[PCA]	Point of closest Approach. 
	\item		[PG\&E]	Pacific Gas and Electric Company. \item		[PUSCH]	Physical Uplink Shared Channel. 
	\item       [PVI]   Perpendicular Vegetation Index.
	\item		[RC]	Radio Control. 
	\item		[RPL]	Routing for low-power and Lossy network.  
	\item		[RSU]	Roadside Unit.
	\item		[SAC]	Special Airworthiness Certificate. 
	\item		[SAR]	Search and Rescue.
	\item       [SAVI]  Soil Adjusted Vegetation Index.
	\item		[SDN]	Software-Defined Networking. 
	\item		[SHM]   Structural Health Monitoring. 
	\item		[SOC]	Start of Charge. 
	\item		[SUAVs]	Solar-Powered UAVs.
	\item		[sUAV]	Small Unmanned Aerial Vehicle. 
	\item		[SVM]	Support Vector Machine. 
	\item       [TIR]    Thermal Infrared.
	\item		[UAS]	Unmanned Aircraft Systems.
	\item	    [UAV]	Unmanned Aerial Vehicle.
	\item		[VANETs]	Vehicular Ad-hoc Networks. 
	\item		[NFV]	Network Function Virtualization. 
	\item		[VIs]	Vegetation Indices. 
	\item		[VMPaaS]	Video Monitoring Platform as a Service. 
	\item		[VTOL]	Vertical Take-Off and Landing.
	\item		[WPT]	Wireless Power Transfer.
	\item		[WSN]	Wireless Sensor Networks. 
\end{abbreviations}

\renewcommand{\baselinestretch}{0.94} 

\bibliographystyle{IEEEtran}
\bibliography{Ref5}

\begin{IEEEbiography}[{\includegraphics[width=1in,height=1.25in,clip,keepaspectratio]{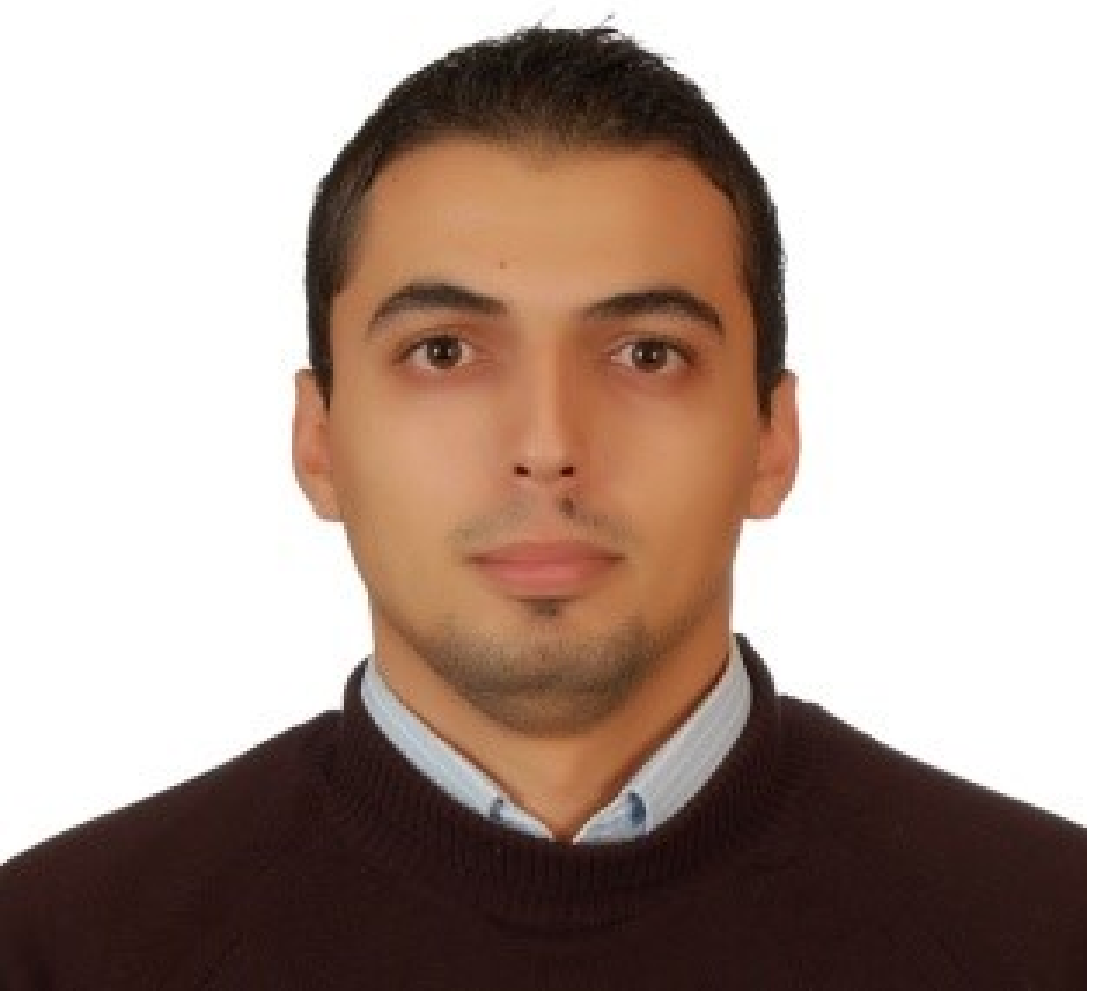}}]{Hazim Shakhatreh}
	is a Ph.D. student at the ECE department of New Jersey Institute of Technology. He received the B.S. degree and M.S. degree 
	in wireless communications engineering from Yarmouk University, Jordan, in 2008 and 2012, respectively. His research interests include wireless communications
	and emerging technologies with focus on Unmanned Aerial Vehicle (UAV) networks.
\end{IEEEbiography}
\begin{IEEEbiography}[{\includegraphics[width=1in,height=1.25in,clip,keepaspectratio]{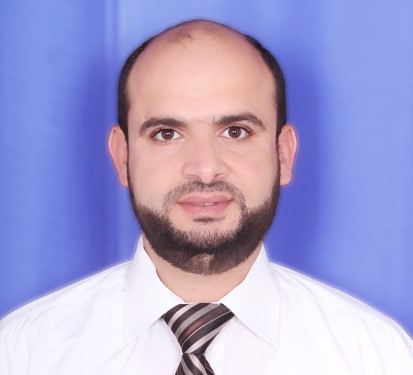}}]{Ahmad H. Sawalmeh}
	is a PhD student at the Engineering College of Universiti Tenaga Nasional (UNITEN), Malaysia. He received his M.S. degree in Computer Engineering from Jordan University of Science and Technology in 2005. He also received his B.S. degree from Jordan University of Science and Technology in 2003. He worked as a lecturer in the Computer Department at the Technical Vocational Training Corporation (TVTC), Kingdom of Saudi Arabia from 2006 – 2016. His research interests include wireless communications, UAVs networks and Flying Ad-Hoc Networks (FANETs).	
\end{IEEEbiography}
\begin{IEEEbiography}[{\includegraphics[width=1in,height=1.25in,clip,keepaspectratio]{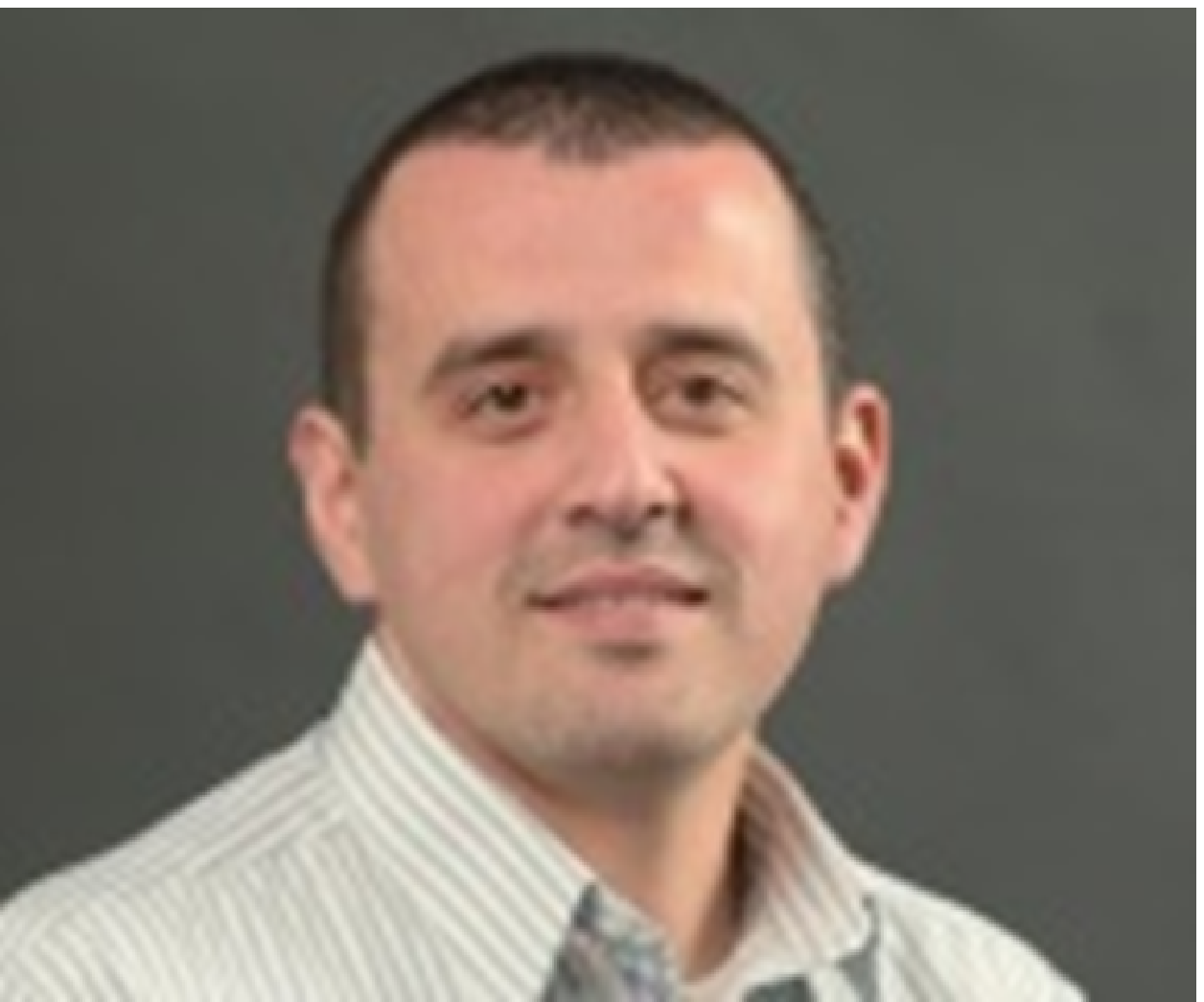}}]{Ala Al-Fuqaha} (S'00-M'04-SM'09) received his M.S. and Ph.D. degrees in Electrical and Computer Engineering from the University of Missouri-Columbia and the University of Missouri-Kansas City, in 1999 and 2004, respectively. Currently, he is Professor and director of NEST Research Lab at the Computer Science Department of Western Michigan University. His research interests include Wireless Vehicular Networks (VANETs), cooperation and spectrum access etiquettes in cognitive radio networks, smart services in support of the Internet of Things, management and planning of software defined networks (SDN) and performance analysis and evaluation of high-speed computer and telecommunications networks. In 2014, he was the recipient of the outstanding researcher award at the college of Engineering and Applied Sciences of Western Michigan University. He is currently serving on the editorial board for John Wiley’s Security and Communication Networks Journal, John Wiley’s Wireless Communications and Mobile Computing Journal, EAI Transactions on Industrial Networks and Intelligent Systems, and International Journal of Computing and Digital Systems. He is a senior member of the IEEE and has served as Technical Program Committee member and reviewer of many international conferences and journals.
\end{IEEEbiography}
\begin{IEEEbiography}
	[{\includegraphics[width=1in,height=1.25in,clip,keepaspectratio]{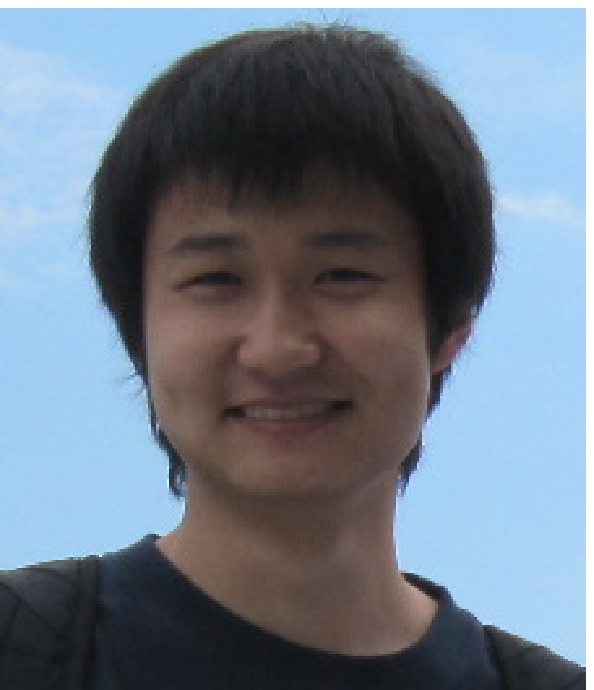}}]{Zuochao Dou}
	received his B.S. degree in Electronics in 2009 at Beijing University of Technology. From 2009 to 2011, he studied at University of Southern Denmark, concentrating on embedded control systems for his M.S degree. Then, he received his second M.S. degree at University of Rochester in 2013 majoring in communications and signal processing. He is currently working towards his Ph.D. degree in the area of cloud computing security and network security with the guidance of Dr. Abdallah Khreishah and Dr. Issa Khalil.
\end{IEEEbiography}
\begin{IEEEbiography}
	[{\includegraphics[width=1in,height=1.25in,clip,keepaspectratio]{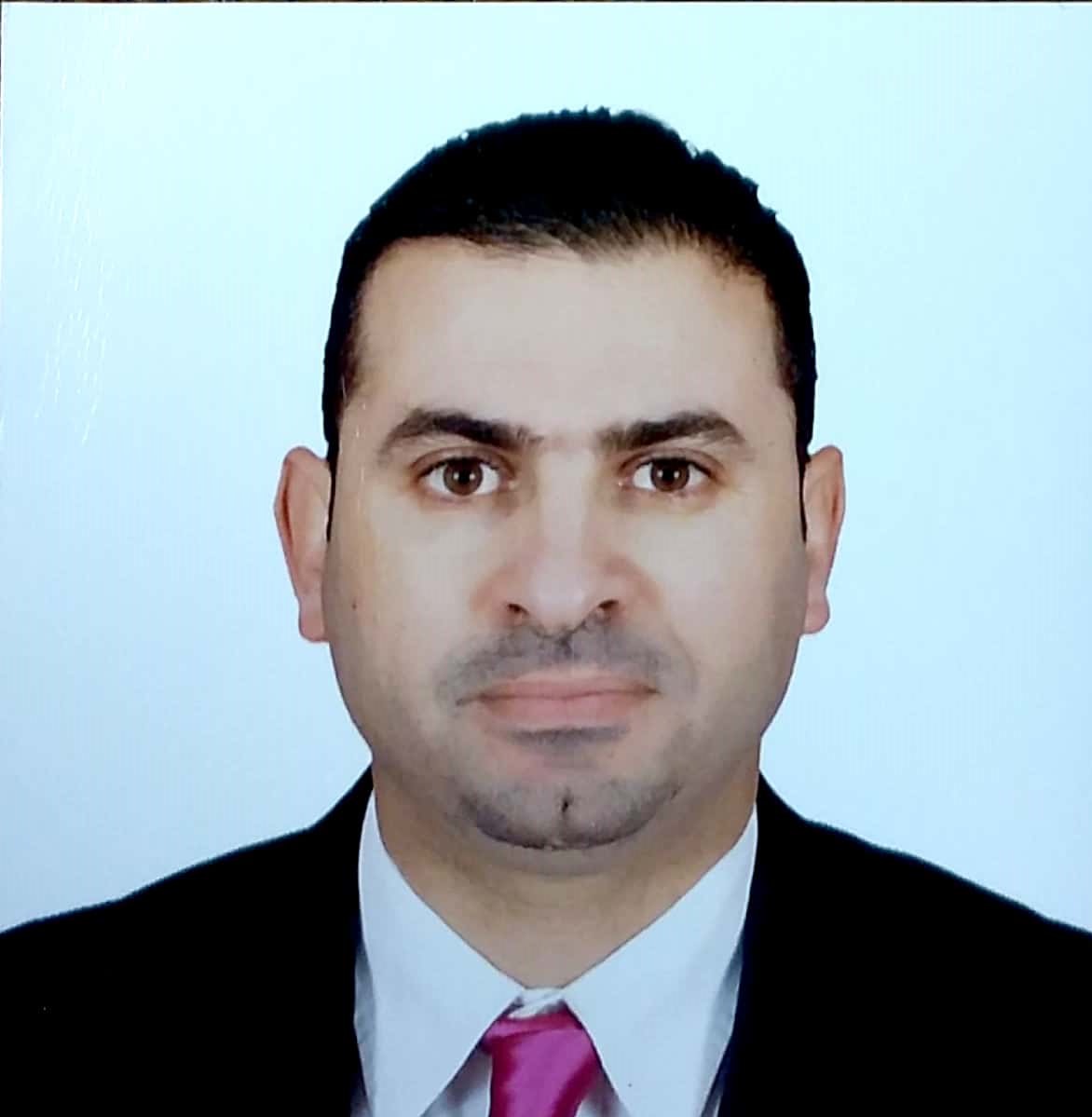}}]{Eyad Almaita}
	received the B.Sc. from the Al-Balqa Applied University, Jordan in 2000, M.Sc. from Al-Yarmouk University, Jordan in 2006, Ph.D. from Western Michigan University, USA in 2012. He is now serving as associate professor at Mechatronics and power engineering dept. in Tafila Technical University, Jordan. His research interests include energy efficient systems, smart systems, power quality, and artificial intelligence.
\end{IEEEbiography}
\begin{IEEEbiography}
	[{\includegraphics[width=1in,height=1.25in,clip,keepaspectratio]{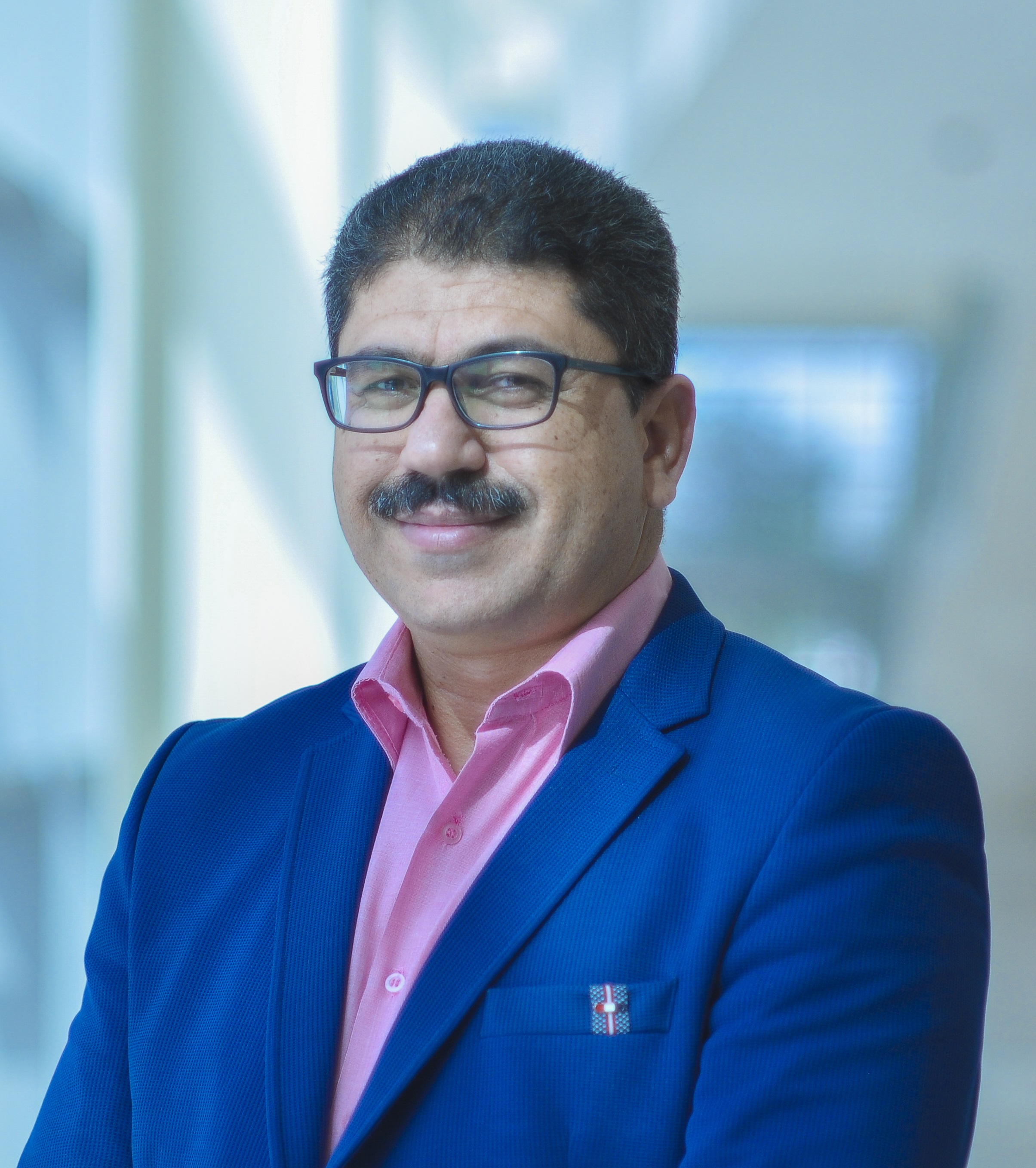}}]{Issa Khalil}
	received PhD degree in Computer Engineering from Purdue University, USA in 2007. Immediately thereafter he joined the College of Information Technology (CIT) of the United Arab Emirates University (UAEU) where he served as an associate professor and department head of the Information Security Department. In 2013, Khalil joined the Cyber Security Group in the Qatar Computing Research Institute (QCRI), a member of Qatar Foundation, as a Senior Scientist, and a Principal Scientist since 2016. Khalil’s research interests span the areas of wireless and wireline network security and privacy. He is especially interested in security data analytics, network security, and private data sharing. His novel technique to discover malicious domains following the guilt-by-association social principle attracts the attention of local media and stakeholders, and received the best paper award in CODASPY 2018. Dr. Khalil served as organizer, technical program committee member and reviewer for many international conferences and journals. He is a senior member of IEEE and member of ACM and delivers invited talks and keynotes in many local and international forums. In June 2011, Khalil was granted the CIT outstanding professor award for outstanding performance in research, teaching, and service. 
\end{IEEEbiography}
\begin{IEEEbiography}
	[{\includegraphics[width=1in,height=1.25in,clip,keepaspectratio]{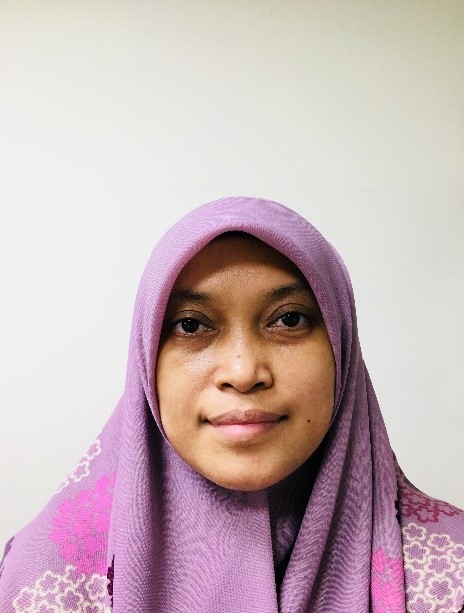}}]{Noor Shamsiah Othman}
	received the B.Eng. degree in electronic and electrical engineering and the M.Sc. degree in microwave and optoelectronics from the University College London, London, U.K., in 1998 and 2000, respectively, and and the Ph.D. degree in wireless communications from the University of Southampton, U.K., in 2008. She is currently with Universiti Tenaga Nasional, Malaysia, as Senior Lecturer. Her research interest include audio and speech coding, joint source/channel coding, iterative decoding, unmanned aerial vehicle communications and SNR estimator. 	
\end{IEEEbiography}
\begin{IEEEbiography}[{\includegraphics[width=1in,height=1.25in,clip,keepaspectratio]{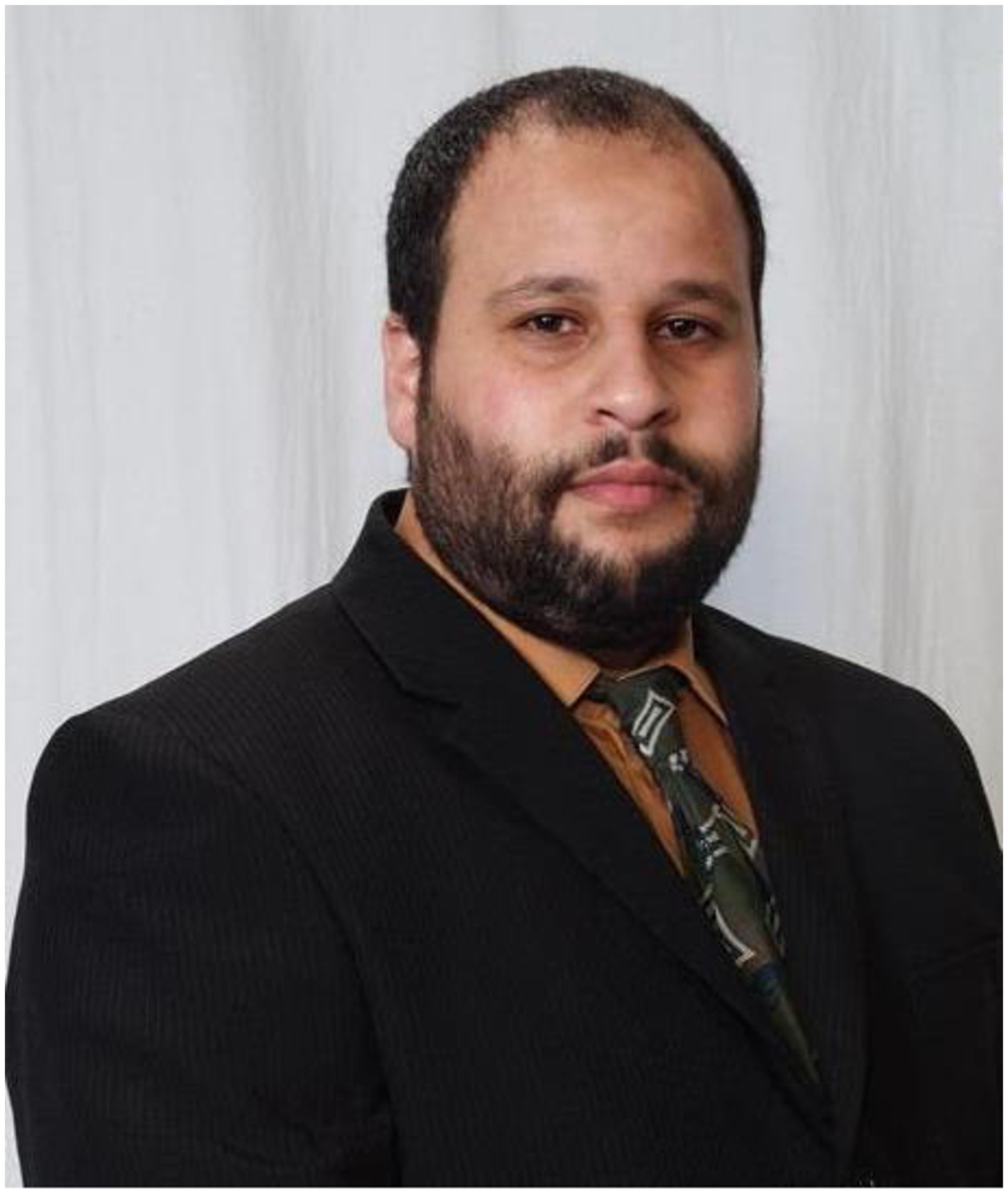}}]{Abdallah Khreishah} is an associate professor in the Department of Electrical and Computer Engineering at New Jersey Institute of Technology. His research interests fall in the areas of visible-light communication, green networking, network coding, wireless networks, and network security. Dr. Khreishah received his BS degree in computer engineering from Jordan University of Science and Technology in 2004, and his MS and PhD degrees in electrical \& computer engineering from Purdue University in 2006 and 2010. While pursuing his PhD studies, he worked with NEESCOM. He is a senior member of the IEEE and the chair of North Jersey IEEE EMBS chapter.
\end{IEEEbiography}
\begin{IEEEbiography}[{\includegraphics[width=1in,height=1.25in,clip,keepaspectratio]{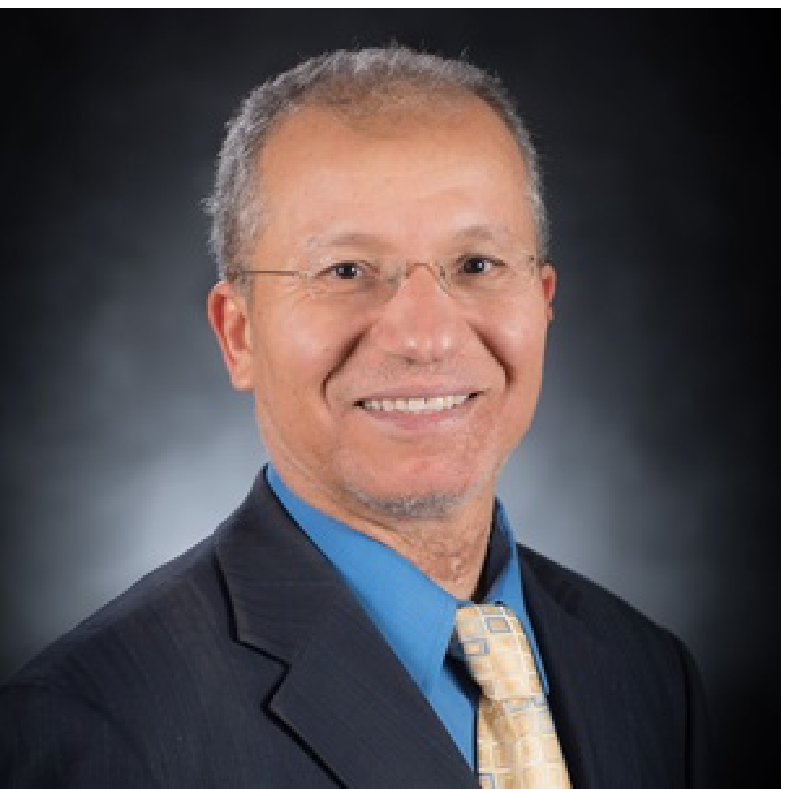}}] {Mohsen Guizani} (S'85, M'89, SM'99, F'09) received his B.S. (with distinction) and M.S. degrees in electrical engineering, and M.S. and Ph.D. degrees in computer engineering from Syracuse University, New York, in 1984, 1986, 1987, and 1990, respectively. He is currently a professor and ECE Department Chair at the University of Idaho. Previously, he served as the associate vice president of Graduate Studies and Research, Qatar University, chair of the Computer Science Department, Western Michigan University, and chair of the Computer Science Department, University of West Florida. He also served in academic positions at the University of Missouri-Kansas City, University of Colorado-Boulder, Syracuse University, and Kuwait University. His research interests include wireless communications and mobile computing, computer networks, mobile cloud computing, security, and smart grid. He currently serves on the Editorial Boards of several international technical journals, and is the Founder and Editor-in-Chief of Wireless Communications and Mobile Computing (Wiley). He is the author of nine books and more than 400 publications in refereed journals and conferences. He has guest edited a number of special issues in IEEE journals and magazines. He has also served as a member, Chair, and the General Chair of a number of international conferences. He was selected as the Best Teaching Assistant for two consecutive years at Syracuse University. He was Chair of the IEEE Communications Society Wireless Technical Committee and TAOS Technical Committee. He served as an IEEE Computer Society Distinguished Speaker from 2003 to 2005.
\end{IEEEbiography}
\end{document}